%% file: manuscript.tex
\documentclass[]{article}

\usepackage{authblk}
\usepackage{graphicx}
\graphicspath {{figures/}}
\usepackage{amsmath}
\usepackage{amssymb}
\usepackage{color}
\usepackage{caption}
\usepackage{subcaption}
\usepackage{xspace} 
\usepackage{multirow}
\usepackage{tabu}
\usepackage{booktabs}
\usepackage[table,xcdraw]{xcolor}
\usepackage{longtable}
\usepackage{rotating}
\usepackage{pdflscape}
\usepackage[colorlinks=true]{hyperref}
\usepackage[nameinlink]{cleveref}

\usepackage{geometry}
\newcommand{\figwidthfull}{0.98\textwidth}
\newcommand{\figwidth}{0.8\textwidth}
\newcommand{\figsemiwidth}{0.48\textwidth}

\newcommand{\beginsupplement}{%
	\setcounter{section}{0}
	\renewcommand{\thesection}{S\arabic{section}}%
	\setcounter{table}{0}
	\renewcommand{\thetable}{S\arabic{table}}%
	\setcounter{figure}{0}
	\renewcommand{\thefigure}{S\arabic{figure}}%
	\setcounter{equation}{0}
	\renewcommand{\theequation}{S\arabic{equation}}%
}

\providecommand{\keywords}[1]
{
	\small	
	\textbf{\textit{Keywords---}} #1
}

\begin{document}

\title{Peak Alignment of Gas Chromatography-Mass Spectrometry Data with Deep Learning}

\author[1]{Mike Li\thanks{mili7522@uni.sydney.edu.au}}
\author[2]{X. Rosalind Wang\thanks{Rosalind.Wang@csiro.au}}
\affil[1]{Centre for Complex Systems,
	The University of Sydney, Sydney, Australia }
\affil[2]{CSIRO Data61,	PO Box 76, Epping, NSW 1710, Australia }

\date{}

\maketitle

\begin{abstract}
	We present ChromAlignNet, a deep learning model for alignment of peaks in Gas Chromatography-Mass Spectrometry (GC-MS) data. In GC-MS data, a compound's retention time (RT) may not stay fixed across multiple chromatograms. To use GC-MS data for biomarker discovery requires alignment of identical analyte's RT from different samples. Current methods of alignment are all based on a set of formal, mathematical rules. We present a solution to GC-MS alignment using deep learning neural networks, which are more adept at complex, fuzzy data sets. We tested our model on several GC-MS data sets of various complexities and analysed the alignment results quantitatively. We show the model has very good performance (AUC $\sim 1$ for simple data sets and AUC $\sim 0.85$ for very complex data sets). Further, our model easily outperforms existing algorithms on complex data sets. Compared with existing methods, ChromAlignNet is very easy to use as it requires no user input of reference chromatograms and parameters. This method can easily be adapted to other similar data such as those from liquid chromatography. The source code is written in Python and available online. 
\end{abstract}


\section{Introduction}
\label{sec:intro}

Gas chromatography when coupled with mass spectrometry (GC-MS) provides a vast amount of information about chemical compositions in samples, thus is regarded as a gold standard in analysis of compounds in metabolomics research. Metabolomics data are influenced by environmental factors~\cite{LNS_ER2013}, individuals' life styles and diet~\cite{Ellis_BM2012}, and diseases~\cite{BernaJID2015,DragonieriJACI2007,DragonieriLung2017}. Metabolomics is therefore seen as an effective tool for health monitoring, disease diagnosis and personalised health care.
One field in which metabolomic analysis is rapidly gaining traction is in human breath diagnostics. Gaseous compounds expired in breath can be sampled and detected by GC-MS to provide insights into the metabolic status of the source,  presenting a readily accessible and non-invasive means of acquiring metabolic data.
Recent advances in data analytics, such as information theory~\cite{BernaJBR2018} and machine learning~\cite{GrissaFMB2016}, have provided some powerful tools for uncovering new information from metabolomics data. 

The wealth of information from GC-MS data allows researchers the opportunity for biomarker discovery by analysing the variations in samples from different cohorts.
Typically, variation among samples is interpreted based on quantitative differences in chemicals, in which analyte identity is recognised using RT.
Performing data analysis on GC-MS samples thus requires that the RT of a given substance remain the same between samples. 

However, across multiple GC-MS analyses the retention time for a given substance may shift due to environmental conditions, gas flow rate, the age of the column, as well as sample overloading and interactions between different components of the mixture being analysed.  
Such factors can influence the RT in a non-linear manner across the duration of the chromatogram. Consequently, alignment of chromatographic peaks corresponding to an identical analyte in different samples is a critical step in data pre-processing. 

Many alignment algorithms have been proposed. For a comprehensive list, we direct the readers to the excellent review article by Vu and Laukens~\cite{VuMetabolites2013}. Moreover, Koh \emph{et al.}~\cite{KohJCA2010} provided a comprehensive evaluation of a number of alignment software that are publicly and commercially available. These authors concluded that due to the complexity of the metabolome, all existing software will require further improvement and researchers are recommended to perform manual checks on the alignment of important biomarkers. 
More alignment algorithms have been developed since the review papers above~\cite{ZhengJCA2013, DomingoAC2016, FuJCA2017, CouprieJCA2017, YangJCA2018, Ottensmann2018}, many of these utilising the mass spectral information for alignment. 

Current alignment algorithms are all based on traditional, symbolic artificial intelligence (AI) techniques, that is, the use of a set of formal, mathematical rules. Symbolic AI have been shown to be excellent at solving well-defined, logical problems, such as chess, however they run into difficulties when turned towards complex, fuzzy problems such as images and speech~\cite{Chollet2017,Goodfellow2016}. RT alignment of metabolomic data belongs to the latter problem. When manually identifying compounds in the chromatogram, experts rely on their knowledge of the instruments, the properties of the chemical compounds, and the profiles of the chemicals in the chromatogram. Much of this knowledge is subjective and intuitive, and can not be captured by formal mathematics. Machine learning techniques tackle these complex problems by allowing AI systems to acquire their own knowledge of the world through extracting patterns from the data. 
We present here ChromAlignNet, a novel method for chromatogram RT alignment through the use of deep learning, a type of machine learning (see Section~\ref{suppsec:deeplearning} in Supplementary Materials for background). We demonstrate the validity of this method through the alignment results of several data sets of various complexities and compare the results with existing algorithms.  The method presented can easily be adapted to other similar data sets such as liquid chromatography, and nuclear magnetic resonance (NMR) spectroscopy.

\section{Proposed Methodology}
\label{sec:method}

\subsection{Network Architecture}
\label{subsec:method_network}

The objective of aligning two chromatograms is to compare the compounds within them and match those ones that are identical in different samples. This type of problem belong to a field of machine learning called One-Shot Learning~\cite{fei2006}. 
We therefore choose to build a deep learning network to compare individual peaks, basing the network architecture on the Siamese network~\cite{Bromley1994}, which contains two or more branches that are identical in their structure, parameters and weights. These branches are connected by a function which computes some metric as a distance between their highest level representations. Each branch can also be called an `encoder', as their top level layer represents features that best encode identifying information about the input. Siamese networks have been applied to problems such as signature verification~\cite{Bromley1994}, face comparison~\cite{Chopra2005} and assessing sentence similarity \cite{Mueller2016}, all classic one-shot learning problems. 

\begin{figure}
	\centering
		\includegraphics[width = \figwidthfull]{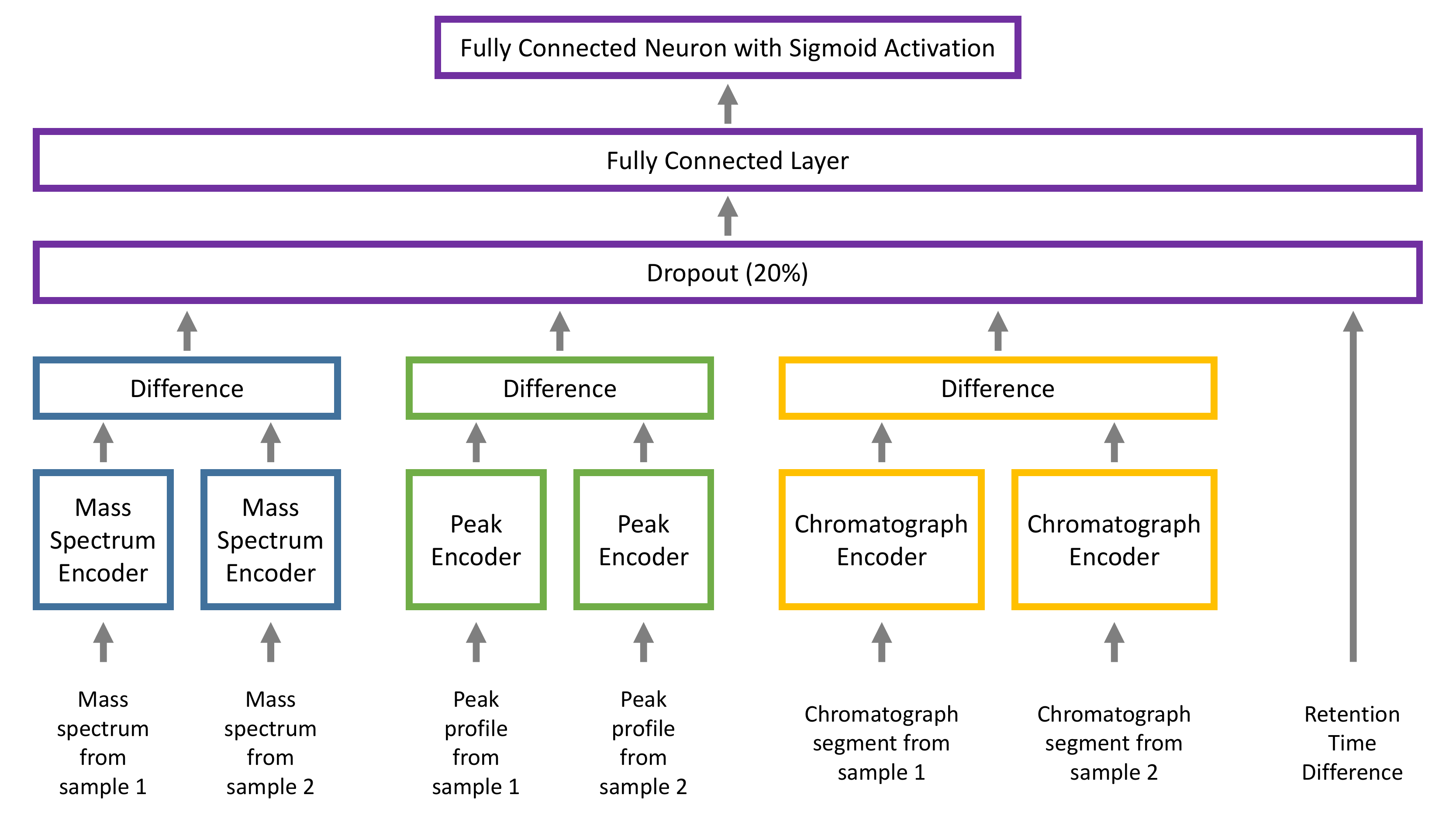}
	\caption{Overall network architecture, featuring a Siamese design for three components - the `Mass Spectrum Encoder', `Peak Encoder' and `Chromatogram Encoder'. The retention time difference is concatenated with the outputs from each Siamese sub-network. A 20\% dropout is applied before passing to a fully connected layer of 64 neurons. Another fully connected layer of one neuron with a sigmoid activation function provides the output. Other activation functions were chosen to be ReLU, since this is a commonly accepted default that often speeds up the training process.}
	\label{fig:NetworkArchitucture}
\end{figure}

To make full use of the rich information present in GC-MS data, we propose that the network should compare not only their retention times, but also the chromatogram segments, the peak profile, and mass spectra of the peaks. To achieve that, the overall network architecture (Figure~\ref{fig:NetworkArchitucture}) employs three separate Siamese sub-networks to compare these aspects of the peaks. In other words, each sub-network will be composed of two branches, to which the input will be the relevant information from a pair of peaks (henceforth, we will refer to this as ``\textit{pairwise comparison}''). The output of the overall network is a probability indicating how likely these peaks should be aligned together. We will describe below the details of each sub-network. 

The mass spectrum sub-network encodes the information of all the m/z values at the RT where the peak is at maximum intensity. The mass encoder uses a standard fully connected structure with two layers of 64 neurons each before the final layer of 10 neurons. Dropout of 0.2 was present between each layer. This simple structure was used since the mass range detected by the instrument is usually a pre-set value and during pre-processing of the data the masses are rounded into their nearest integers, thus the size of the input vector to the mass encoder are consistent across all data that we gather. 

\begin{figure}[h]
	\centering
	\includegraphics[width = .75\textwidth]{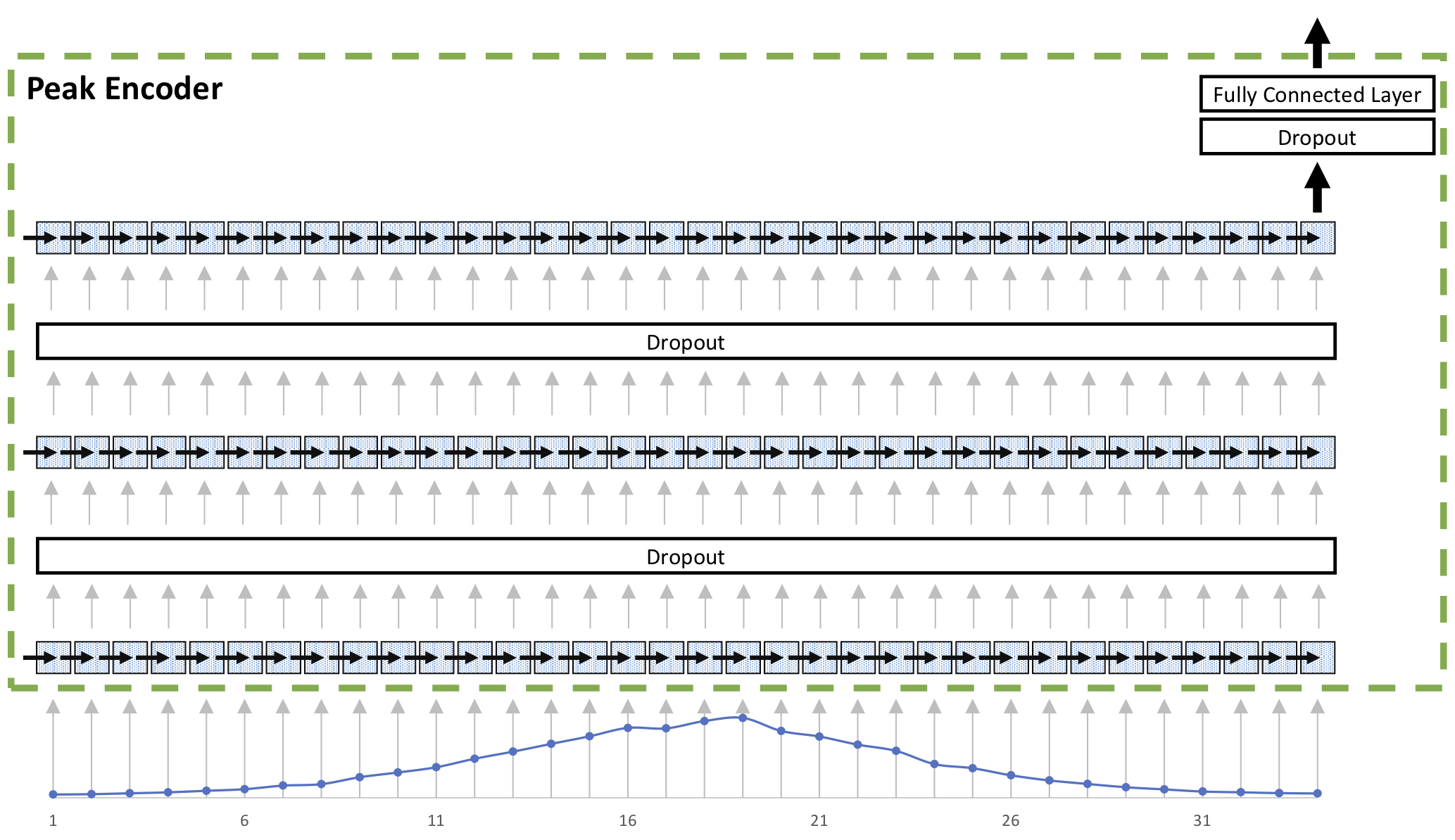}
	\caption{Architecture of the peak encoder component. For each recurrent layer, the blue shaded box represents the LSTM unit (containing 64 neurons for its internal state and 64 neurons for its memory). The elements of the input sequence are fed sequentially. The bidirectional aspect of the recurrent layers are omitted for compactness. For recurrent layers two and three, the output from the previous layer is used as the input.}
	\label{fig:PeakEncoder}

	\centering
	\includegraphics[width = .75\textwidth]{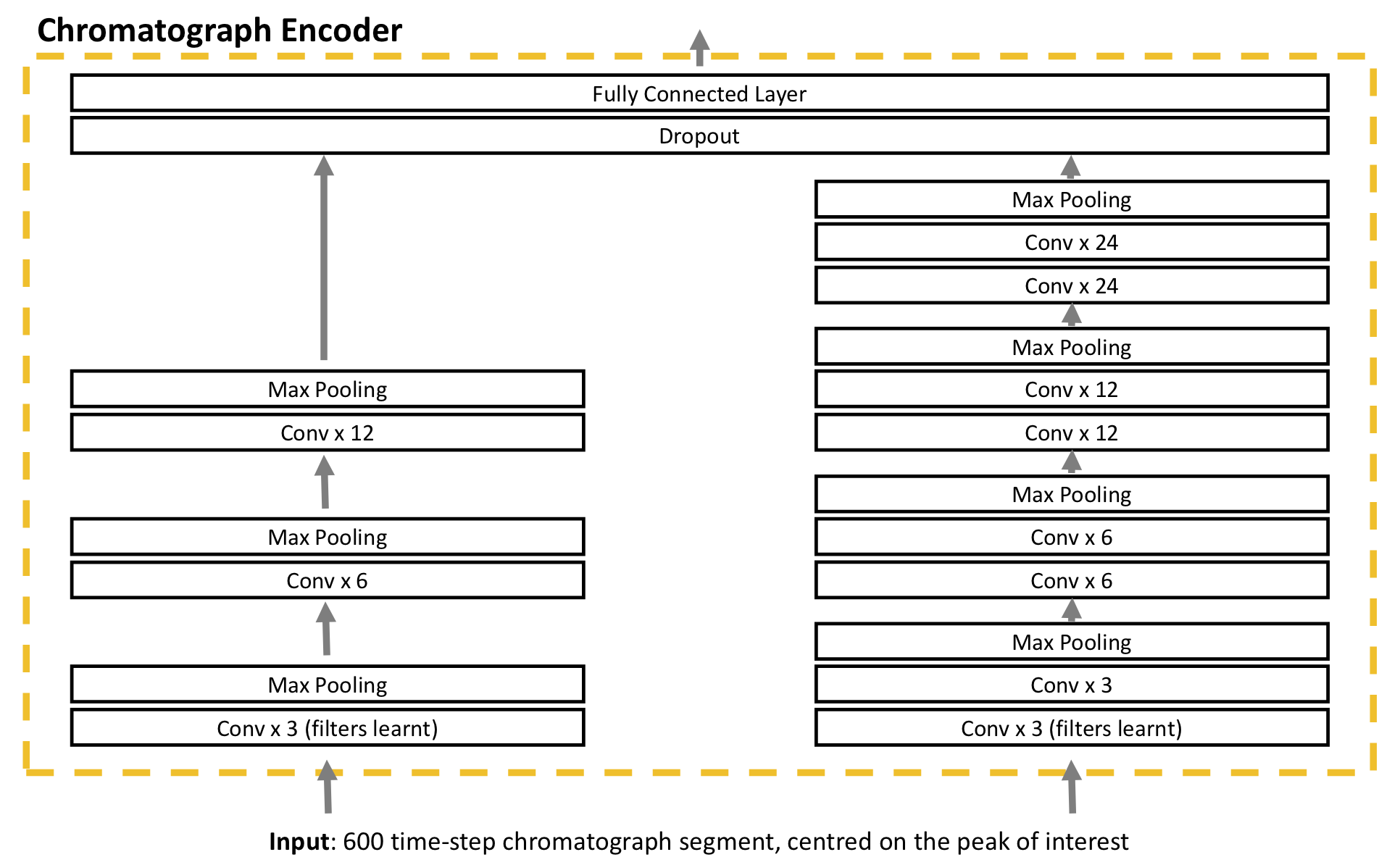}
	\caption{Architecture of the chromatogram encoder component. Each convolution layer uses a kernel size of 3. Each max pooling layer uses a kernel size and stride of 2, except the first which uses a kernel size and stride of 3.}
	\label{fig:ChomatographEncoder}
\end{figure}

The peak profile sub-network encodes the information along the peak's RT range to describe its shape -- that is, the intensity at all timepoints from the beginning to the end of each peak (see Section~\ref{subsec:expt_peak} for details of peak detection). The peak encoder (Supplementary Figure~\ref{fig:PeakEncoder}) uses a recurrent structure to deal with the variable duration of different peaks. Three recurrent layers were used, each separated by a dropout layer. The recurrent unit used is the LSTM cell, which keeps a separate memory on top of its internal state. A bidirectional structure is used, so the input for each layer is processed in both the forward and reverse directions and the output is concatenated together. After the three recurrent layers the output at the final time-step is passed on to the dropout and fully connected layers.
The input to the peak encoder is normalised to a maximum intensity of one.

Finally, we designed a sub-network to encode the information of a segment of the chromatogram centred on the peak. This is because often the surrounding peaks will inform us about the identity of the peak in question. The chromatogram encoder (Figure~\ref{fig:ChomatographEncoder}) makes use of 1D convolutions for feature extraction.
There are two stacks of layers, one with a greater number of convolution and pooling operations than the other. The deeper stack has a wider receptive field, allowing it to pick out large scale features from the chromatogram. Meanwhile, the stack with fewer convolution operations can focus on the finer details of the peaks. Inspiration for this multi-level approach was taken from the Inception module in the GoogLeNet~\cite{Szegedy2015}. Each convolution is kept at a kernel size of 3, opting for more layers rather than a larger kernel size. The result is a similar receptive field but greater non-linearity, a technique successfully demonstrated by VGG Net~\cite{Simonyan2014}.

Note that the inputs to the chromatogram encoder and peak encoder overlap. Inclusion of the peak encoder in the overall network to emphasise the peak component was considered valuable. As the preprocessing steps and network architectures differ between the two encoders, unique information is extracted by each.

Each of the encoders end in a fully connected layer of 10 neurons, giving a representation of the input as a vector of 10 values. The two branches of the Siamese structure are combined by taking the absolute difference between each element of the two vectors, producing an output of 10 neurons for each of the three sub-networks. These 30 neurons are concatenated together along with the absolute value of the difference in retention time between the two peaks, which is provided as an additional input.

This concatenated combination is passed through a dropout layer, where a dropout rate of 20\% is applied. Since the RT difference is input as a single neuron this means that there is a 20\% chance in each training iteration of this information being absent from the model. This may support the training of the encoders at an early stage (when they are not yet accurate), instead of the network simply using the RT difference as the basis for alignment. It is likely for simple examples that the retention time difference is the most important piece of information in deciding whether or not two peaks match (see Section~\ref{subsec:expt_data} for discussion on data complexity). However, in complex cases the retention times can vary by more than one minute. The information from the encoders may be more consistent than RT difference across these different cases. Encouraging reliance on a mixture of information from the different encoders should lead to better generalisation of the model across different data sets. The concern of generalisation may also be naturally resolved as more training data is used, capturing a variety of cases.

\subsection{Getting Groups from Pairwise Comparisons}
\label{subsec:method_groups}

For alignment, first all pairs of eligible peaks are compared using the trained model to get a prediction value, then peaks are allocated to \emph{groups} according to these prediction values. Thus the result of alignment is a set of groups, where each group contains peaks from multiple samples that are predicted to be the same compound. We will now describe the steps in detail. 

When comparing peaks between multiple chromatograms, a cut-off on the retention time difference is imposed to limit the number of comparisons. This can be set at a value around 1.5 minutes in practice, since it is highly unlikely to find retention times shifted by more than this amount even in complex scenarios. A larger value of three minutes was used during the testing of the algorithm. 

Outputs from the network are converted into distance values by taking the inverse, which then allow peaks to be grouped together using a hierarchical clustering algorithm. The `average linkage' method (also known as the UPGMA algorithm) was chosen, which combines groups based on the average unweighted distance of all the pairs within the groups. A cut-off distance of 2 was used, corresponding  to a probability of 0.5. 

A retention time is then assigned to each group. The group's RT is the average RT of the peaks in the group weighted by the peaks' intensity. The weighting helps avoid extreme shifts in RT values if noisy peaks of low intensity are mis-assigned to the group.

\section{Experimental Design}
\label{sec:expt}

\subsection{Data}
\label{subsec:expt_data}

Data from four different sources were used to train and evaluate the effectiveness of the algorithm. These included two sets of air data and two sets of human breath data. Detailed descriptions of the data and the relevant ethics approvals can be found in supplementary materials (Section~\ref{suppsec:data}). 
All samples were analysed using quadrupole time of flight instrument (GC-QTOF-MS, 7890B series, Agilent Technologies). 
Detailed description of data acquisition procedure can be found in previous studies~\cite{BernaJBR2018,WangJCB2018}. 
Note that QTOF-MS instruments give higher resolution mass spectrometry than standard GC-MS instruments. However, in exporting the data, the MS values were compressed into integer values, therefore, the data used here is similar to a standard GC-MS instrument's output, consequently, we will refer to them as simply GC-MS data. 

We used data from ambient air and human breath samples in our study due to the differences in their complexity. The ambient air samples are simpler and alignment of chromatographic peaks is straight forward, as there is minimal RT drift across multiple samples. Conversely, human breath samples are much more complex, with larger disparity in RT, making correct alignment of the peak data using existing algorithms problematic. Examples of unaligned chromatographic segments of the samples can be see in Supplementary Figure~\ref{fig:mz103}. 

Two of these sources of data will be used to generate training and validation samples (Section~\ref{subsec:expt_generatingTraining}) and two to generate test samples (Section~\ref{subsec:expt_generatingTest}). Validation samples are used during the training process to verify the increase in accuracy over training iterations, or epochs, using data the network hadn't encountered during training. Validation minimises the risk of overfitting, where networks fit too perfectly to the training data and cannot generalise to new data. 
The test samples are then used to fully evaluate networks trained on training data, as these are data the networks have never encountered.

\subsection{Peak Detection}
\label{subsec:expt_peak}

The data from the GC-QTOF-MS instrument were exported from the instrument's native file format to XML files, where each reading of the intensity was assigned non-integer m/z and RT values. These data are first converted into a csv file, where all the m/z are rounded to the nearest integer values. This step ensures the data are consistent with lower resolution mass spectral output, where the mass values are reported as integers, thus the peak detection and alignment algorithms can be applied to data from different mass analysers. 

An optional normalisation process was applied to the data, whereby a standard chemical is analysed on each day of analysis and used to normalised all data from that day~\cite{WangJCB2018}. 
The baseline drift of each single ion chromatogram (SIC) was corrected using Asymmetric Least Square (ALS)~\cite{Eilers2005}. 

Peak detection of the samples was achieved through an automated process~\cite{VivoTruyols2005133} for each SIC. A peak is located when the signal's second derivative is negative, and the start and end of each peak is found when the first derivative crosses the positive or negative threshold. The time at which a peak achieves its maximum intensity (i.e.~when the second derivative is minimum) is used to define the peak's RT. The area under the curve for each peak from the start to the end of the peak is calculated and attributed as the total count of the peak.

\subsection{Pre-processing for Alignment}
\label{subsec:expt_preprocessing}

The detected peaks are processed together with the raw GC-MS data to gather the three components (mass spectrum, peak profile and chromatogram segment) needed for the deep learning network described in Section~\ref{sec:method}. The mass spectrum was taken along the m/z axis at the peak's maximum RT. The peak profile is the section of chromatogram between the peak's start and end times along the time axis. The chromatogram segment was 600 time steps centred at the peak's RT, corresponding to 1.5 minutes in either direction. 
Supplementary Figure~\ref{fig:GCMS-Data-AssignmentProcess} shows how these different pieces of information can be used together to help identify peaks that should be aligned. 

We further processed all the data for use in the network: A log operation was performed on the chromatogram segments, ignoring any zeros. The minimum value was subtracted. 
The peak profile and mass spectrum associated with each peak were normalised to 1.

\subsection{Generating Training and Validation Data}
\label{subsec:expt_generatingTraining}

We used the ambient air samples from \emph{P. falciparum} CHMI trial and the healthy control human breath samples to generate training and validation data. 

We selected three masses (m/z = 103, 115, and 143 Da) with RTs between 14-15min from the air samples for training. These specific peak data were selected as the  m/z and RT combinations are indicative of biomarkers of interest in the related \emph{P. falciparum} CHMI trial~\cite{BernaJBR2018}. We selected two of these three masses from the healthy control human breath samples. The peaks identified in these segments were manually grouped based on their peak profiles, mass spectra and chromatogram segments. Detailed description of the manual grouping process can be found in the Supplementary Materials (Section~\ref{suppsec:manualGroups}). 

From the healthy control human breath data, we selected a further two masses (m/z = 73 and 88 Da) between RT 3-6 min for training and validation samples. These m/z and RT combination are indicative of three sulphur compounds (thioethers) in the GC-MS data, which were previously shown to increase in volunteers undergoing CHMI with \emph{P. falciparum} as infection progresses~\cite{BernaJID2015}. The RT of these early-eluting peaks 
can vary for more than 1 min, which is much larger than the duration of the individual peak. It is not unusual for the RT of a later peak in one sample to occur before the RT of an earlier peak in another sample. The large variation in the RT of these compounds makes alignment using traditional methods extremely difficult. The thioethers in the data were identified by expert interpretation\footnote{Henceforth, we use the word `detected' to describe a peak that is found by the automatic peak detection algorithm, and the word `identified' to describe a peak that has been manually verified by the domain expert.}. Detailed description of the analysis can be found in~\cite{BernaJBR2018}. 
Only the thioether peaks were used in the training data as positive pairs (see definition in the next paragraph), the other peaks were only used in the training data when combined with the thioether peaks to form negative pairs. 

\begin{table}
	\centering
	\caption{Single Ion Chromatogram (SIC) segments used to generate training examples. Note, for the last two breath data sets, only the professionally identified thioether peaks, of which there were 283 and 272 respectively in each data set, were given a group number. In the same region, there were 272 and 138 non-thioether peaks respectively, these were given `-1' as a group number and only used in training when paired with an identified peak as a negative pair. }
	\label{table:TrainingSamples}
	\resizebox{\textwidth}{!}{%
		\begin{tabular}{@{}clccccccc@{}}
			\toprule
			\textbf{\begin{tabular}[c]{@{}c@{}}Training \\Data set\end{tabular}} &  & Type & m/z & \begin{tabular}[c]{@{}c@{}}Retention Time\\ Segment (min)\end{tabular} & 
			\begin{tabular}[c]{@{}c@{}}No. of\\ Samples\end{tabular} & 
			\begin{tabular}[c]{@{}c@{}}Unique\\ Peaks ID'd\end{tabular} & 
			\begin{tabular}[c]{@{}c@{}}No. of\\ Groups\end{tabular} & 
			\begin{tabular}[c]{@{}c@{}}Training\\ Combinations\end{tabular} \\ \midrule
			Air103 &  & Air & 103 & 13.9 - 15.1 & 62 & 358 & 9 & 18,648 \\
			Air115 &  & Air & 115 & 13.9 - 15.5 & 62 & 501 & 12 & 25,752 \\
			Air143 &  & Air & 143 & 14.0 - 15.2 & 62 & 244 & 7 & 11,152 \\
			Breath103 &  & Breath & 103 & 13.9 - 15.3 & 98 & 536 & 13 & 43,723 \\
			Breath115 &  & Breath & 115 & 13.8 - 15.5 & 98 & 766 & 14 & 61,473 \\
			Breath73 &  & Breath & 73 & 3 - 6 & 98 & 283 & 3 & 21,920 \\
			Breath88 &  & Breath & 88 & 3 - 6 & 98 & 272 & 3 & 19,892 \\ \bottomrule
		\end{tabular}
	}
\end{table}

Seven different sets of training and validation data were generated (Table~\ref{table:TrainingSamples}). The name given to each data set is the source data `type' combined with the m/z (e.g.~Air103 or Breath73). Within each data set, positive and negative pairs were fed into the deep learning network. 
A positive pair comprises two peaks that should be aligned, while a negative pair is made of two peaks that correspond to different compounds that do not align. There are many more negative pairs possible than positive ones. To avoid an unbalanced training process, only the same number of random negative pairs were used as positive pairs existed. 

\begin{table}
	\centering
	\caption{The shaded cells highlight the combination of data sets which were used to train each of the seven models}
	\label{table:Dataset&Model}    
	\begin{tabular}{@{}llllllllll@{}}
		\toprule
		&  & \multicolumn{7}{c}{Model} &  \\ \cmidrule(l){2-10} 
		&  & A & B & C & D & E & F & G & \\ \midrule
		Air103 &  & \cellcolor[HTML]{9B9B9B} & \cellcolor[HTML]{5674F9} & \cellcolor[HTML]{9B9B9B} & \cellcolor[HTML]{5674F9} & \cellcolor[HTML]{9B9B9B} & \cellcolor[HTML]{5674F9} & \cellcolor[HTML]{9B9B9B} &\\
		Air115 &  & \cellcolor[HTML]{9B9B9B} & \cellcolor[HTML]{5674F9} & \cellcolor[HTML]{9B9B9B} & \cellcolor[HTML]{5674F9} & \cellcolor[HTML]{9B9B9B} & \cellcolor[HTML]{5674F9} & \cellcolor[HTML]{9B9B9B} &\\
		Air143 &  &  & \cellcolor[HTML]{5674F9} &  &  &  &  &  &\\
		Breath103 &  &  &  & \cellcolor[HTML]{9B9B9B} &  & \cellcolor[HTML]{9B9B9B} & \cellcolor[HTML]{5674F9} & \cellcolor[HTML]{9B9B9B} & \\
		Breath115 &  &  &  &  & \cellcolor[HTML]{5674F9} & \cellcolor[HTML]{9B9B9B} &  &  & \\
		Breath73 &  &  &  &  &  &  & \cellcolor[HTML]{5674F9} &  & \\
		Breath88 &  &  &  &  &  &  &  & \cellcolor[HTML]{9B9B9B} & \\ \bottomrule
	\end{tabular}
\end{table}

Seven models (A-G) were trained using different combinations of these data sets (Table~\ref{table:Dataset&Model}). This is because we want to evaluate model performances when they are trained on data sets with different complexities. We included both Air103 and Air115 data sets in all models, as air samples are the simplest, so they form the basis of training. Air143 data set was left out of all except one model to evaluate how well the models can perform on a simple data set when they are trained with data at various levels of complexity. Breath103 and Breath115 are data set at the next level of complexity from air samples. Therefore, we add different combinations of these two data sets to the training data to train models C, D and E. Finally, the Breath73 and Breath88 data sets are the most complex, therefore we add them to models F and G training data. Note that for models F and G, only the Breath103 data set was used in the training, so that we can evaluate the models trained at higher level of complexity on those with medium levels of complexity. 
As an initial evaluation, each model was used to align each of the data sets. Note that we evaluate the models on all data sets, even those in the training data, since not all data in the training sets will be used for training. 
Since these data sets originate from sources with different levels of alignment difficulty (e.g. varying RT drift early and later in the chromatogram), an assessment of how the accuracy changes for various types of training data can be made.

\subsection{Generating Test Data}
\label{subsec:expt_generatingTest}

We used the ambient air data from \emph{P. vivax} CHMI trial (40 samples) and the field samples of human breath (50 samples) to generate test data (Table~\ref{table:TestData}). 

From the air samples, two masses were selected (m/z = 92 Da between RT 14.5-17.5 min and m/z = 134 Da between RT 13.8-15.5 min). These specific m/z and RT combination were selected due to the identification of biomarkers of interest in the related \emph{P. vivax} trial~\cite{BernaJBR2018}. The peaks detected in these segments were then manually grouped based on their peak profiles, mass spectra and chromatogram segments. 

We selected three masses from the field samples (m/z = 73 and 88 Da between RT 3-6 min, and m/z = 134 Da between RT 13.8-15.5 min). As with the training and validation data sets the first two mass and RT combinations are associated with three thioether compounds, and the third combination is associated with three previously identified cymene compounds~\cite{BernaJBR2018}. The peaks for thioethers and cymene in the field samples were identified by an expert. Other peaks in these segments were assigned into the `un-identified' group. 

\begin{table}
	\centering
	\caption{Single Ion Chromatogram (SIC) segments used to generate test data. Note, for the field breath data, only the professional identified peaks were given a group number; un-identified peaks are given -1 as group number. }
	\label{table:TestData}
	\resizebox{\textwidth}{!}{%
		\begin{tabular}{@{}clcccccccc@{}}
			\toprule
			\textbf{\begin{tabular}[c]{@{}c@{}}Test \\Data set\end{tabular}} &  & Type & m/z & \begin{tabular}[c]{@{}c@{}}Retention Time\\ Segment (min)\end{tabular} & 
			\begin{tabular}[c]{@{}c@{}}No. of\\ Samples\end{tabular} & 
			\begin{tabular}[c]{@{}c@{}}ID'd\\ Peaks\end{tabular} & 
			\begin{tabular}[c]{@{}c@{}}No. of\\ Groups\end{tabular} & 
			\begin{tabular}[c]{@{}c@{}}un-ID'd\\ Peaks\end{tabular} & 
			\begin{tabular}[c]{@{}c@{}}Combinations\end{tabular} \\ \midrule
			Air92 &  & Air & 92 & 14.5 - 17.5 & 41 & 539 & 21 & 7 & 148,785 \\
			Air134 &  & Air & 134 & 13.8 - 15.5 & 41 & 270 & 10 & 9 & 38,781 \\
			Field73 &  & Breath & 73 & 3 - 6 & 50 & 26 & 3 & 244 & 36,585 \\
			Field88 &  & Breath & 88 & 3 - 6 & 50 & 27 & 3 & 93 & 7,140 \\ 
			Field134 &  & Breath & 134 & 13.8 - 15.5 & 50 & 112 & 3 & 203 & 50,403 \\
			\bottomrule
		\end{tabular}
	}
\end{table}

\subsection{Metrics Used for Assessment}
\label{subsec:expt_metric}

The performance of the models were assessed using the metrics of true and false positive rates, and ROC curves and their AUC values.

The model generates a probability of alignment which is converted into a binary classification using a threshold of 0.5. The rate of true positive is the proportion of positive cases (peaks that should be aligned) that are correctly determined as such. The rate of false positives is the proportion of negative cases that were identified as positive cases. These are peaks that should not be aligned together but were incorrectly predicted as such by the model. 

Receiver Operating Characteristic (ROC) curves are useful for visualising the performance of the models in a compact manner without the need to set an arbitrary threshold. ROC graphs plot false positive rate (x-axis) against true positive rate (y-axis), showing the trade-off as the threshold varies. When comparing models, one is said to be better, if it is towards the upper left hand side of the ROC space, where more true positives are produced than false positives. The performances between models can be compared by their respective measure of Area Under the Curve (AUC), the AUC value is between 0 and 1, a model with AUC=1 is said to give perfect performance.

\section{Results}
\label{sec:results}

\subsection{Training Process}
\label{subsec:res_training}

Neural networks' training processes are stochastic in nature. We therefore trained 10 repetitions of each model to give us a general idea of the performance of our models. In practice, multiple networks are trained and the best performing model is picked for use. We choose to randomly split the data in Table~\ref{table:TrainingSamples} into 80\% `training' and 20\% `validation' which is used to check the performance of the trained network at each epoch. See Supplementary Section~\ref{supp:training} for full details of the training process and results. 

The models were trained for 50 epochs. Depending on the size of the training set, this took between 2 and 10 hours using a single GPU node, or 6 to 27 hours using a single CPU node, on the supercomputer clusters.

\subsection{Accuracy and Alignment for Training Data}
\label{subsec:res_accuracyTraining}

\begin{figure}
	\centering
	\includegraphics[width = \figwidthfull]{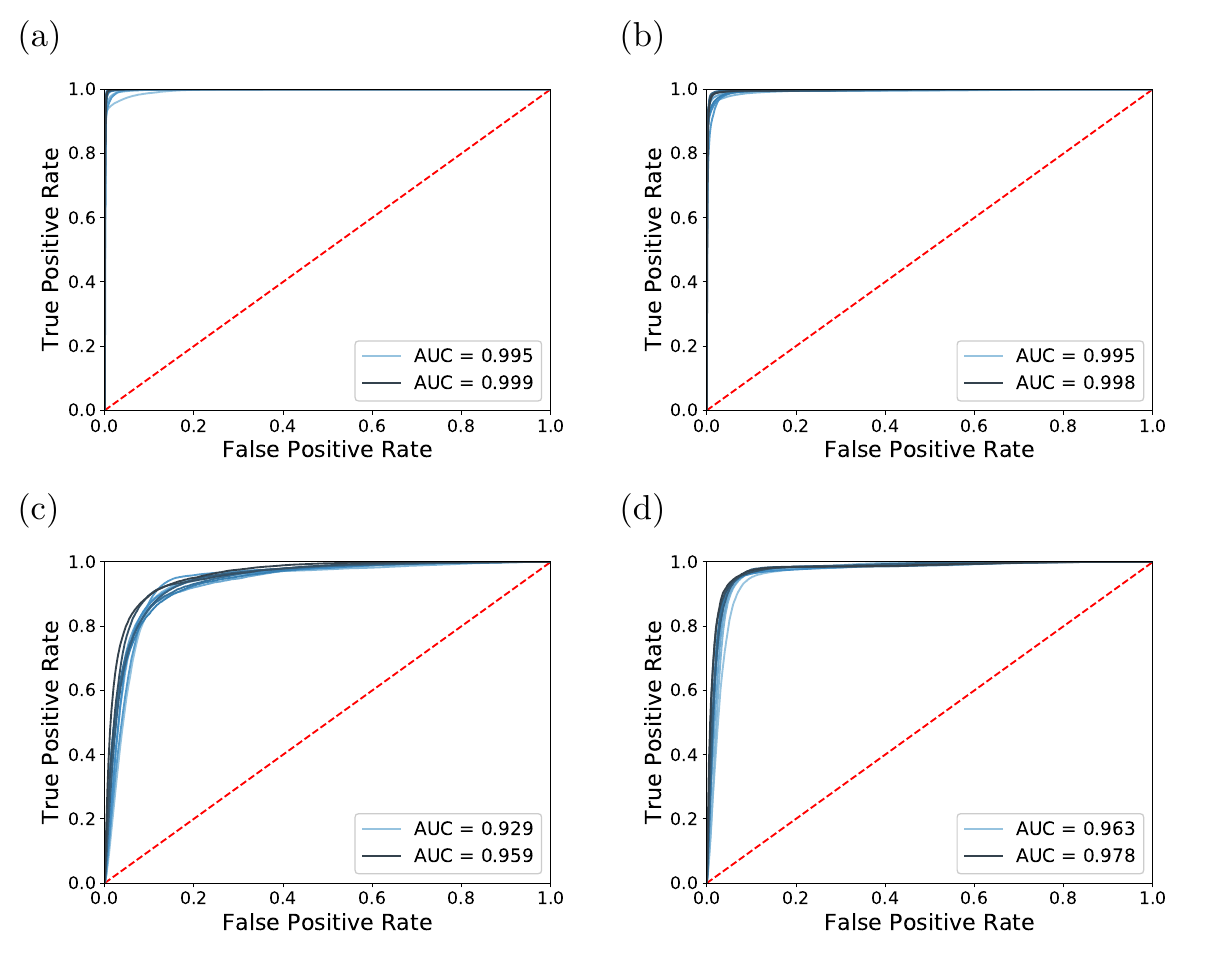}
	\caption{ROC Curves of a) 10 repetitions of model A, ran against data set Breath103, b) 10 repetitions of model A, ran against data set Breath115,
		c) 10 repetitions of model G, ran against data set Breath73 and d) 10 repetitions of model F, ran against data set Breath88. The lines are coloured in ascending order from lowest AUC value to highest AUC value, with the lowest and highest values labelled in the legend.}
	\label{fig:ROCTraining}
\end{figure}

\begin{figure}
	\begin{center}
		\begin{subfigure}[b]{\figwidthfull}
			\centering
			\caption{} \vspace{-0.3cm}
			\includegraphics[width=\figwidthfull]{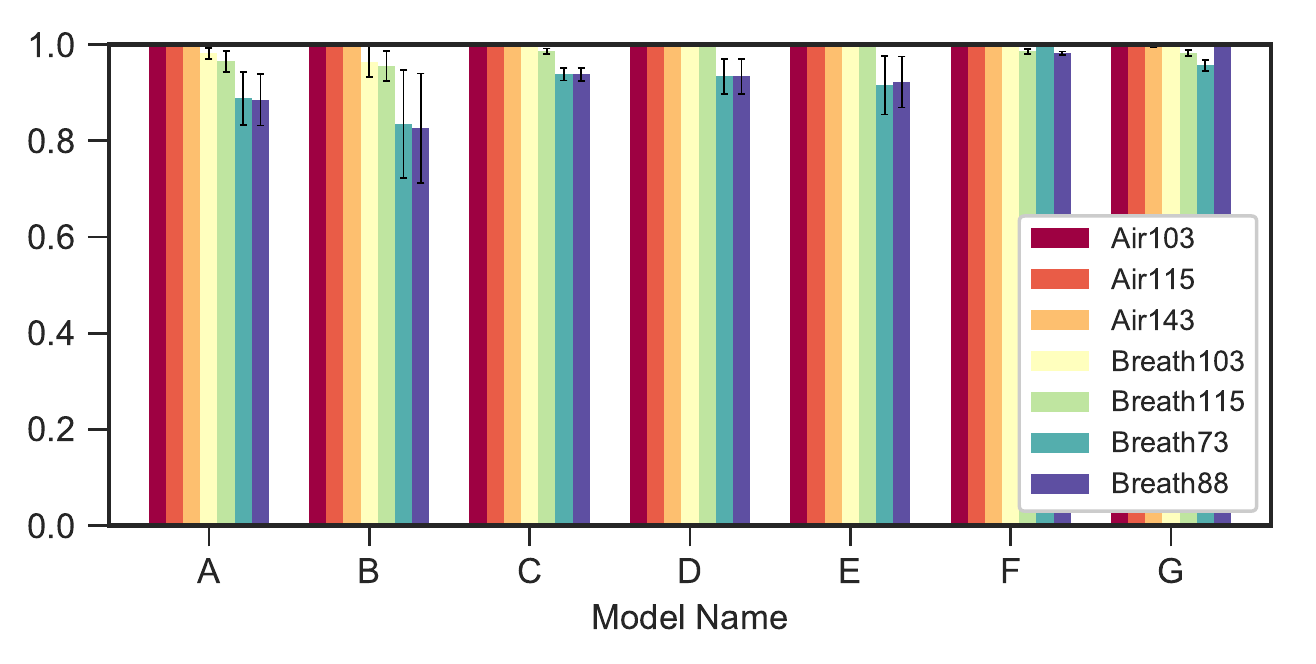}
		\end{subfigure}
		
		\begin{subfigure}[b]{\figwidthfull}
			\centering
			\caption{} \vspace{-0.3cm}
			\includegraphics[width=\figwidthfull]{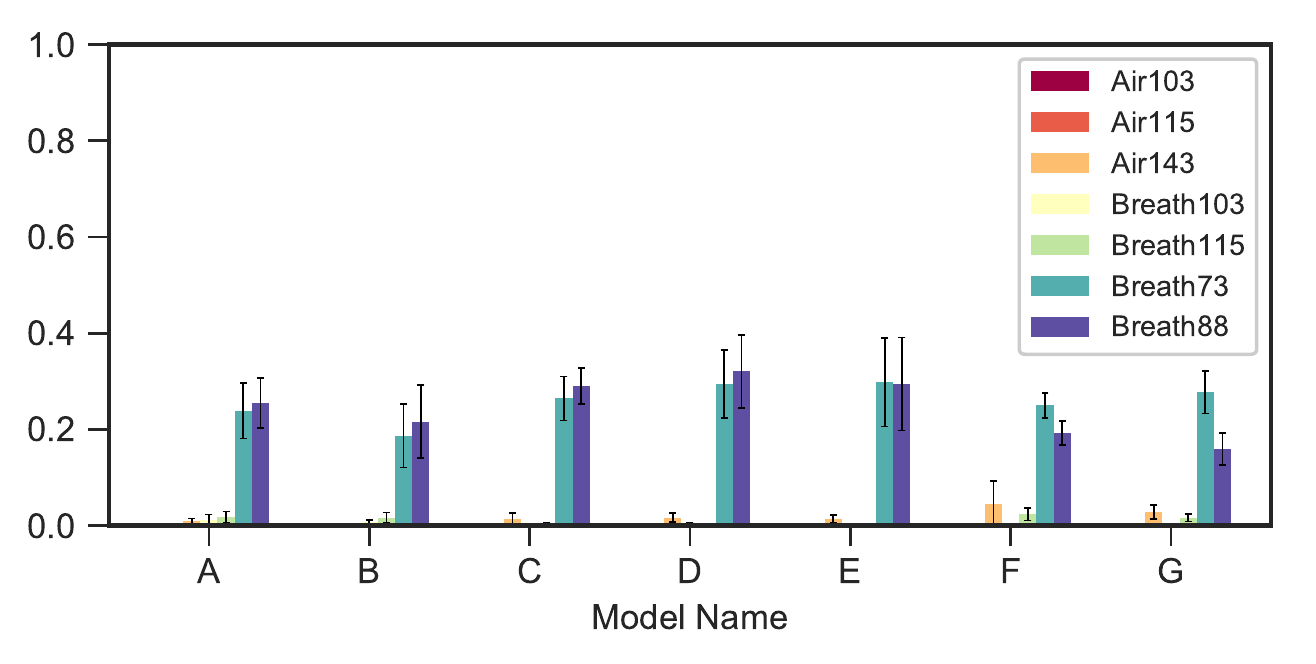}
		\end{subfigure}
	\end{center}
	\caption{Rates of (a) true positives and (b) false positives of all models tested against all data sets. The error bars show the standard deviation from ten repetitions}
	\label{fig:TrueAndFalsePositives}
\end{figure}

We first evaluated all seven trained models (Table~\ref{table:Dataset&Model}) using all the training data sets (Table~\ref{table:TrainingSamples}). This allowed us to observe the effectiveness of the models on data sets on which they were trained\footnote{Note that due to the validation split of the data set as well as the balancing of the positive and negative samples, not all samples in each data set were used for training.}, as well as on other data sets. We report the average true and false positive rates (Supplementary Table~\ref{table:TrueFalsePositives} and Supplementary Figure~\ref{fig:TrueAndFalsePositives}) and the average AUC (Table~\ref{table:AUC_training}) for each model tested on each of the data sets\footnote{For the data sets Breath73 and Breath88, only the expertly identified thioethers were used in the training data for true positive pairs. However, in predicting the alignment, all peaks in the data set were tested. This means when calculating the true positives, only the thioethers peaks were considered, since we do not know the true identity of the other peaks.}. Examples of ROC curves is shown in Figure~\ref{fig:ROCTraining}.

\begin{table}
	\centering
	\caption{AUC of predictions for each model against each training and validation data set. The values report the average over ten repetitions. Cells where the values are italic and in bracket indicate models that were trained with the data set.}
	\label{table:AUC_training}%
	\resizebox{\textwidth}{!}{%
		\begin{tabular}{@{}lccccccc@{}}
			\toprule
			& \multicolumn{7}{c}{Models} \\ 
			\cmidrule(lr){2-8}
			Data Set & A     & B     & C     & D     & E     & F     & G \\
			\midrule
			Air103 & (\textit{1.0000}) & (\textit{1.0000}) & (\textit{1.0000}) & (\textit{1.0000}) & (\textit{1.0000}) & (\textit{1.0000}) & (\textit{1.0000}) \\
			Air115 & (\textit{1.0000}) & (\textit{1.0000}) & (\textit{1.0000}) & (\textit{1.0000}) & (\textit{1.0000}) & (\textit{1.0000}) & (\textit{1.0000}) \\
			Air143 & 0.9999 & (\textit{1.0000}) & 0.9998 & 0.9999 & 0.9997 & 0.9996 & 0.9997 \\
			Breath103 & 0.9985 & 0.9965 & (\textit{1.0000}) & 0.9996 & (\textit{1.0000}) & (\textit{1.0000}) & (\textit{1.0000}) \\
			Breath115 & 0.9964 & 0.9942 & 0.9965 & (\textit{1.0000}) & (\textit{1.0000}) & 0.9942 & 0.9950 \\
			Breath73 & 0.9064 & 0.9105 & 0.9258 & 0.9175 & 0.9119 & (\textit{0.9857}) & 0.9413 \\
			Breath88 & 0.8977 & 0.8914 & 0.9130 & 0.9053 & 0.9101 & 0.9720 & (\textit{0.9879}) \\
			\bottomrule
		\end{tabular}%
	}
\end{table}%

The models generally had very good AUC values and high rate of true positives, meaning that the encoders were able to provide identifying information about the peak. We will not discuss the results for Air103 and Air115 data sets, since all models were trained with these two data sets, except to note that Models F \& G, which were trained with the more complex data sets, did not give perfect FP rates for these data sets. This might be that these models are relying more on the individual encoders' information than the difference between RT, and the few pairs of peaks that have very similar profiles in both the mass spectrum and the chromatogram segment can confuse the models. This FP will generally not matter too greatly for the final alignment since the near perfect TP rates mean the peaks will be aligned to their correct group. 

For the other five data sets, the models had near perfect AUC results for the `simpler' data sets of Air143, Breath103 and Breath115, and reasonable AUC (between 0.90 and 0.95) for the more complex Breath73 and Breath88 data sets. 

However, the rate of false positives was also high for some models, meaning that they were unable to distinguish between complex cases. This occurred most often when a model was trained on data sets of lower complexity compared to the data set with which it was evaluated. These models may have combined the different pieces of information inappropriately. For example, a model trained on the air data sets and evaluated with the breath data sets may have placed too much confidence in the RT difference as the most important piece of information. As we have seen, differences in RT can be significant for `complex' samples such as human breath, so a model that relies heavily on predictable differences in this component will likely perform poorly when presented with data samples in which RT is a less reliable indicator of peak identity. In such cases, peaks with a small difference in retention time may end up grouped together even if the other encoding elements suggest otherwise. 

\begin{figure}[h!]
	\begin{center}
		\begin{subfigure}[b]{\figwidthfull}
			\caption{} \vspace{-0.3cm}
			\includegraphics[width = \figwidthfull]{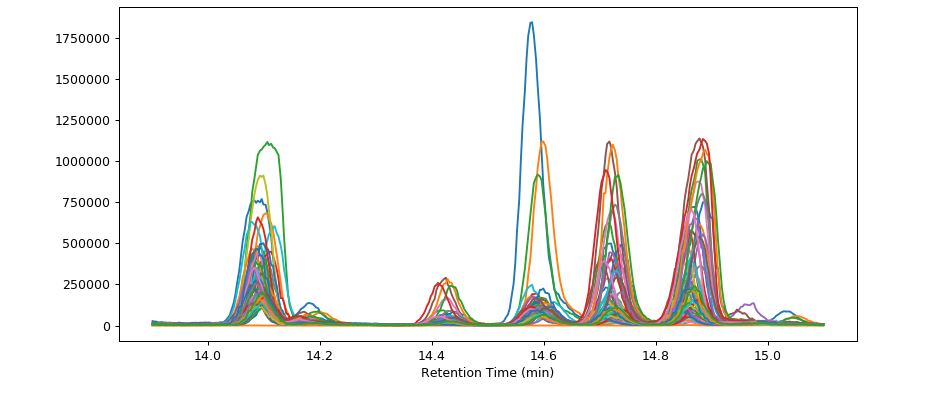}
		\end{subfigure}
		
		\begin{subfigure}[b]{\figwidthfull}
			\caption{} \vspace{-0.3cm}
			\includegraphics[width = \figwidthfull]{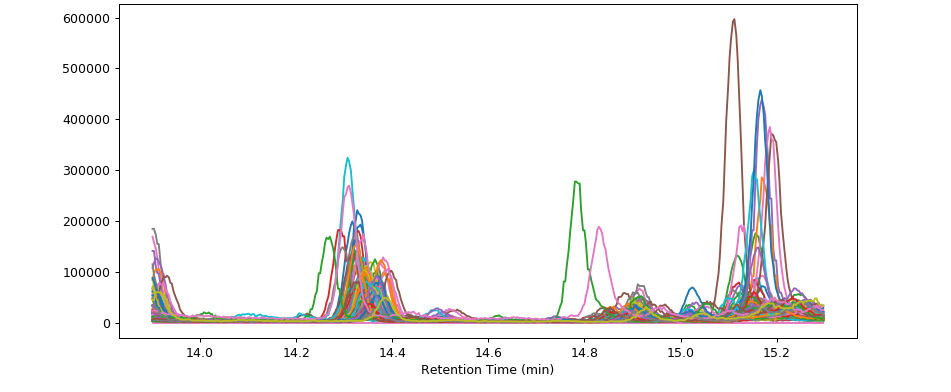}
		\end{subfigure}
		
		\begin{subfigure}[b]{\figwidthfull}
			\caption{} \vspace{-0.3cm}
			\includegraphics[width = \figwidthfull]{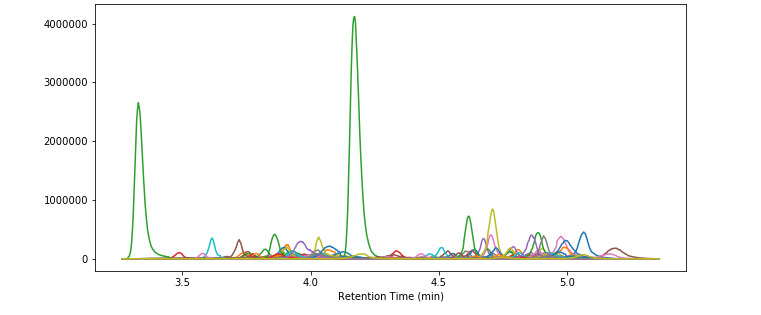}
		\end{subfigure}
	\end{center}    
	\caption{Chromatographic segment various difficulties prior to alignment: (a) 62 air samples at m/z = 103 Da between RT $\sim$14 and 15 minutes, (b) 98 breath samples at the same m/z and RT combination of (a), and (c) 98 breath samples at m/z = 73 Da between RT $\sim$3 and 6 minutes. }
	\label{fig:mz103}
\end{figure}

We observe in the results that Breath103 and Breath115 had fairly low FP rates, even when the prediction was made using models A and B, which were trained using just the air data sets. Conversely, the results for Air143 had high FP rates in comparison. This was unexpected because air samples were considered relatively `simple' data. This shows while the unaligned SIC for the two data sets are very different (Supplementary Figure~\ref{fig:mz103}, where we can easily see the groups of peaks in the air samples but not in the breath samples) there is enough information about these peaks in the air samples to align the breath samples well. At the same time, the chemical compounds in the Air143 data set are different enough from the other two air samples, that the encoded information in the network is insufficient to align these peaks well. 

\begin{figure}
	\centering
		\includegraphics[width=\figwidthfull]{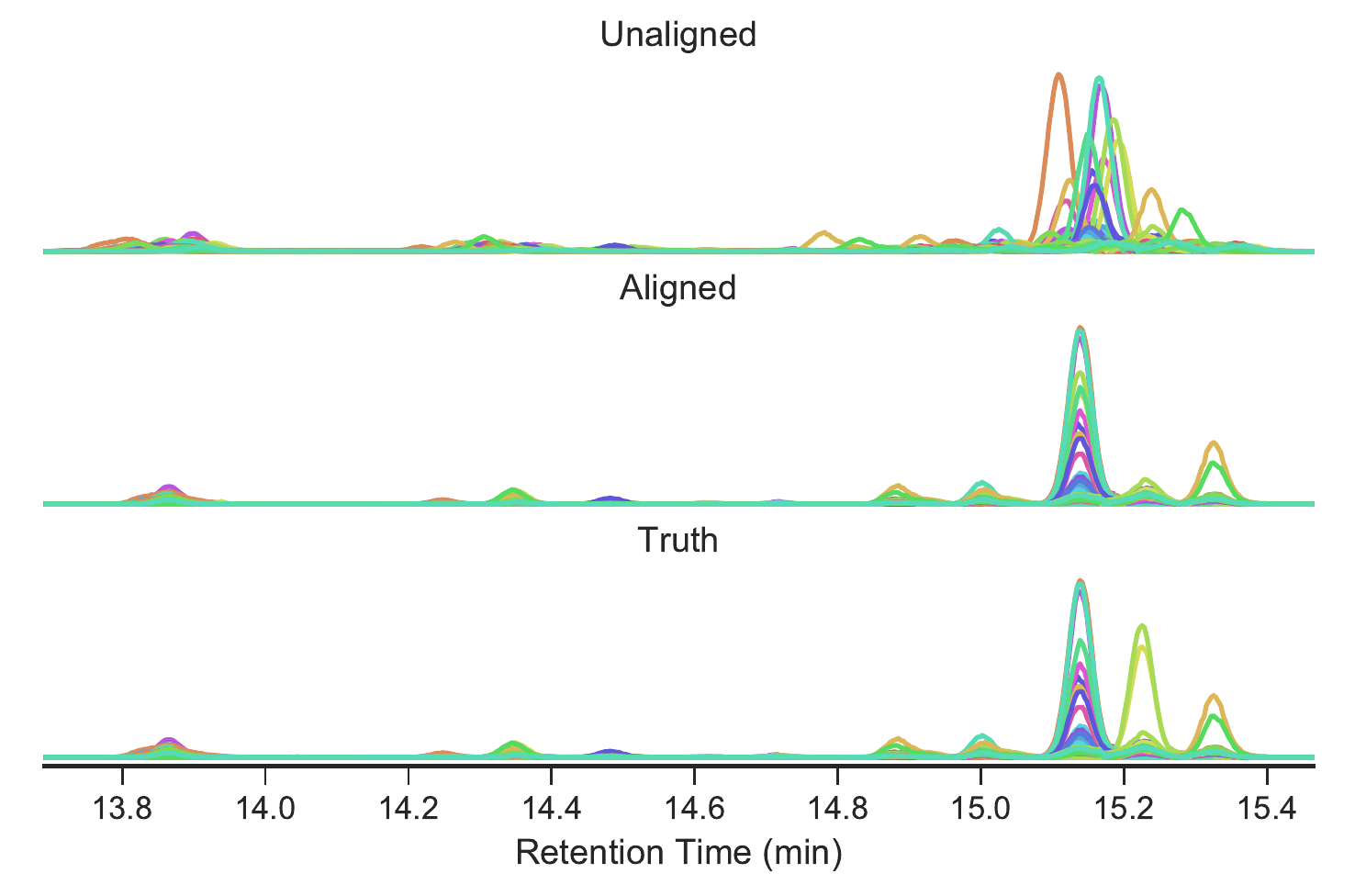}
	\caption{Alignment outcome for the data set Breath115 using the best performing network from model C. For this network, the true positive rate was 0.992 and the false positive rate was 0.004. (See Supplementary Figure~\ref{fig:AlignBreath115_C01} for the chromatographic image of the alignment.)} 
	\label{fig:AlignBreath115_C01_spec}
\end{figure}

Figure~\ref{fig:AlignBreath115_C01_spec} shows an example of the alignment outcome, using model C on data set Breath115. Here the result is from the network with the best TP rate on the data set. From the chormatographic image in the result, we can see the overall alignment of all 98 breath samples is a very close match to the true alignment positions for all compounds in the chromatographic segment. We can see from the chromatographic plot in Figure~\ref{fig:AlignBreath115_C01_spec}, that there are a couple of very obvious misclassifications at RT=15.23min. However, when we analysed the two mis-classified peaks, their mass spectra and their chromatogram segments were different from the other peaks in the same group of the ground truth. This means these peaks were mis-identified by human error and the network had correctly aligned them into the group of peaks at RT=15.14min. This example shows the power of using the mass spectral information in alignment, as the two samples in question here do not have a peak at RT=15.23min, and the deep learning network correctly matched them to the relevant peaks.  

We observe that the false positive rates for data sets Breath73 and Breath88 were consistently higher than other data sets (Table~\ref{table:TrueFalsePositives}). The alignment results for these two data sets seem to match well with the ground truth (see Supplementary Figure~\ref{fig:AlignBreath88_F01} for an example). Indeed, if all non-thioether peaks in these data sets are ignored, the false positive rates from models F and G are almost 0. Therefore, it is the non-thioether peaks that cause the high false positives in the results. As the negative training examples from these data sets do not include the other peaks, the trained network can not fully distinguish a negative pair of peaks from a positive pair. We tested this hypothesis by including the non-thioether peaks in the training data for negative pairs, and achieved a much reduced FP rate of around 0.05. 

A common mis-classification occurs when two adjacent peaks\footnote{We compare peaks within the same SIC due to the way the peak detection algorithm works, when sometimes noise in the signal will cause a compound to have a double peak in the chromatogram, this will in turn cause the peak detection algorithm to separate the peak.}, have very similar mass spectra, with significant contributions from higher molecular weight m/z than the small masses associated with the thioethers. This type of mis-classification is due to the information we used in the mass encoder of the network. In future work, we will investigate the maximum m/z threshold to be used in the encoder. 

From the results in Table~\ref{table:TrueFalsePositives}, the main difficulty is the high false positive rate. It can also be seen from Figure~\ref{fig:StandardModelLossComponents} that it is difficult to train the peak encoder. To see if changes to the model architecture could alleviate these two problems, 30 different modifications to model D were tested. Section~\ref{suppsec:modelVariations} (Supplementary Materials) describes all the variations to the model as well as their performances. We found in general, the peak encoder does not add to the model and can be removed without impacting the performance, further, the variants without the peak encoder have much faster training and prediction time than the base model. Additionally, modifying the architecture of the mass or chromatogram encoder can slightly improve the overall performance.

\subsection{Accuracy and Alignment of Test Data}
\label{subsec:res_accuracyTest}

\begin{figure}
	\begin{center}
		\begin{subfigure}[b]{\figwidthfull}
			\centering
			\caption{} \vspace{-0.3cm}
			\includegraphics[width=\figwidthfull]{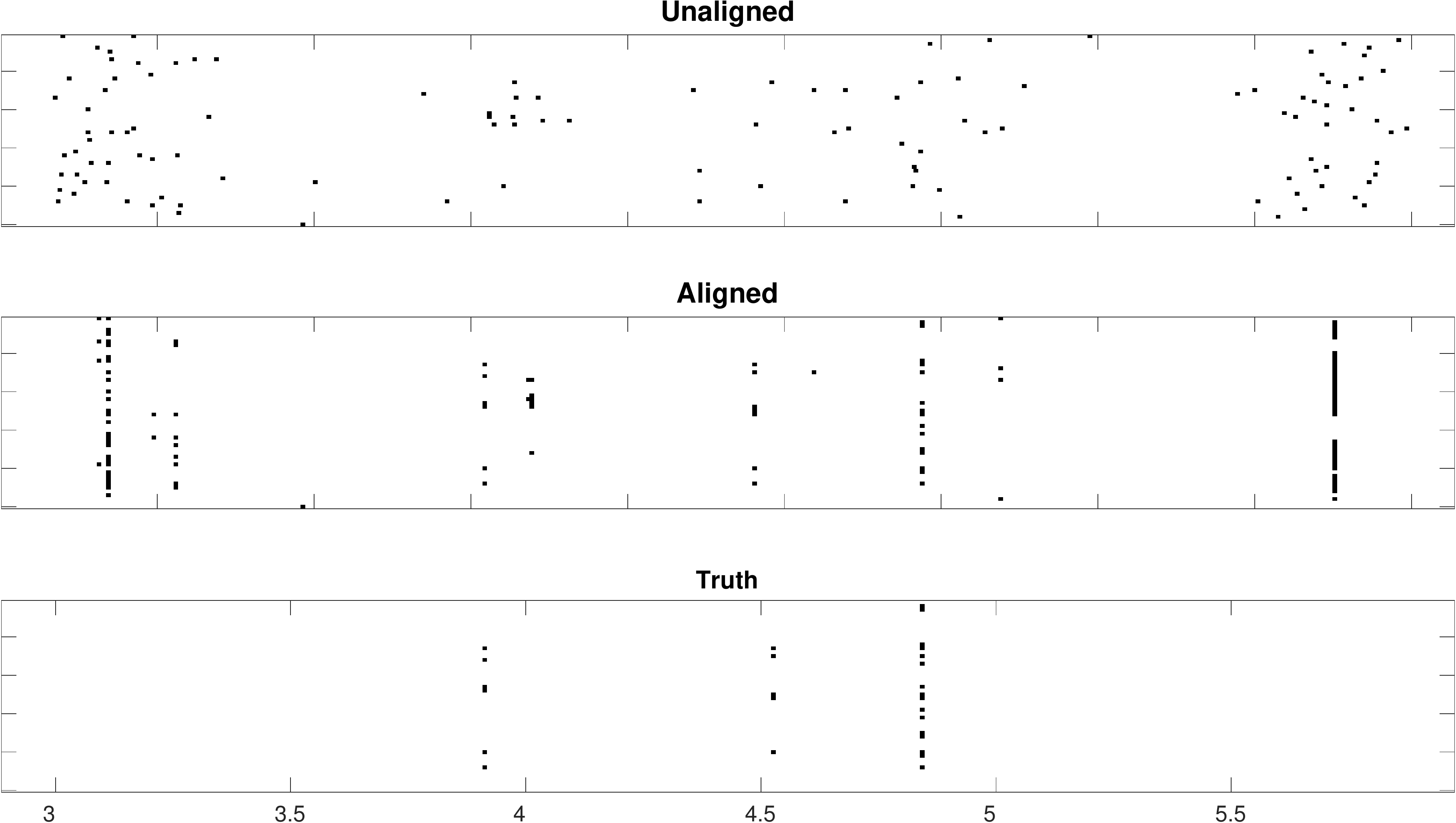}
		\end{subfigure}
		
		\begin{subfigure}[b]{\figwidthfull}
			\centering
			\caption{} \vspace{-0.3cm}
			\includegraphics[width=\figwidthfull]{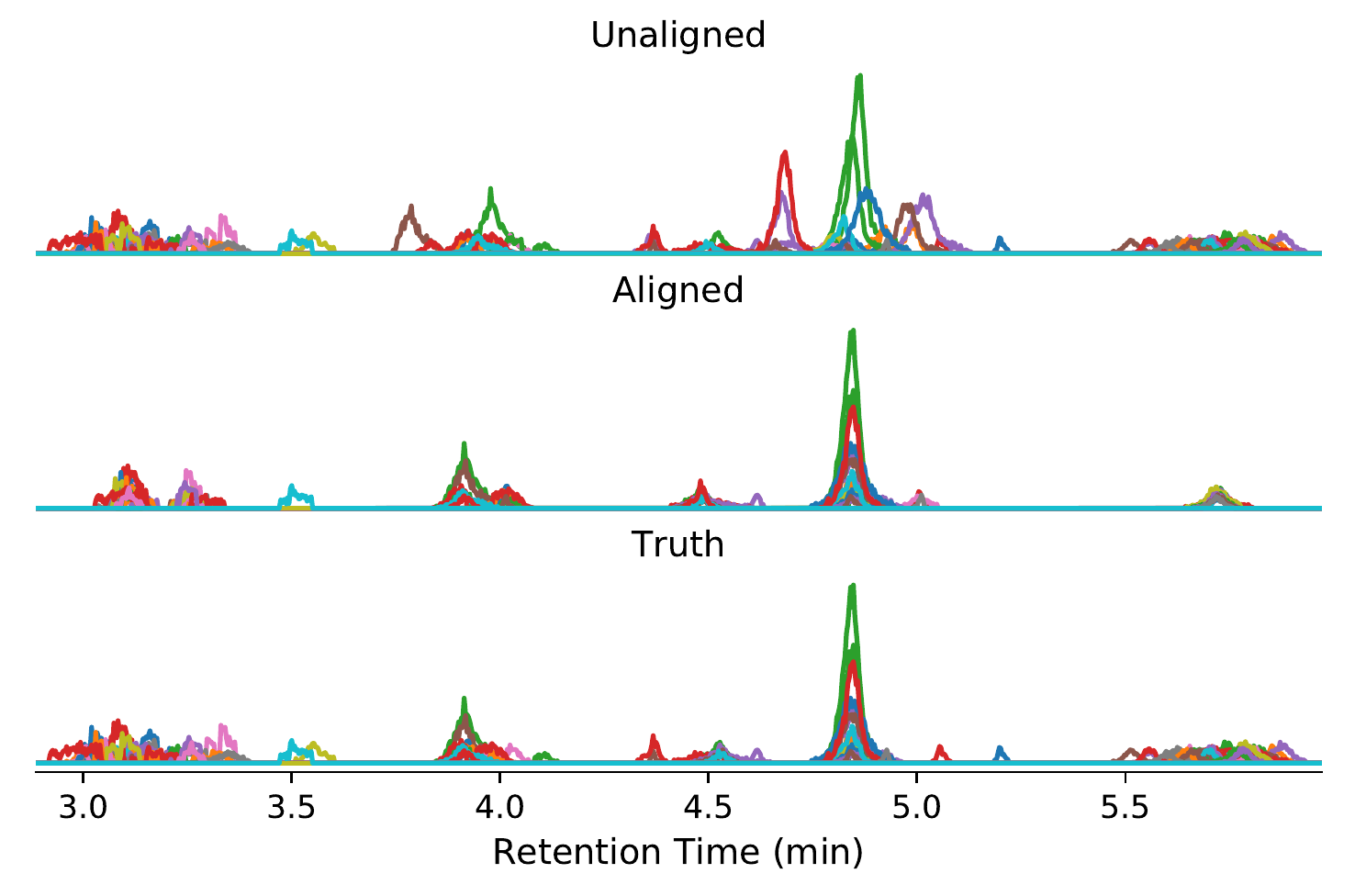}
		\end{subfigure}
	\end{center}
	\caption{Alignment outcome for the data set Field88 using the best performing network of model H-21. For this network, the true positive rate was 0.993 and the false positive rate was 0.134. (a)~Chormatographic image. (b)~Chormatographic plot. Note, only the three thioethers peaks were expertly identified and thus aligned in the ground truth plot, and in ground truth image only these peaks are shown. 
	} 
	\label{fig:AlignField88}
\end{figure}

\begin{figure}
	\centering
	\includegraphics[width = \figwidthfull]{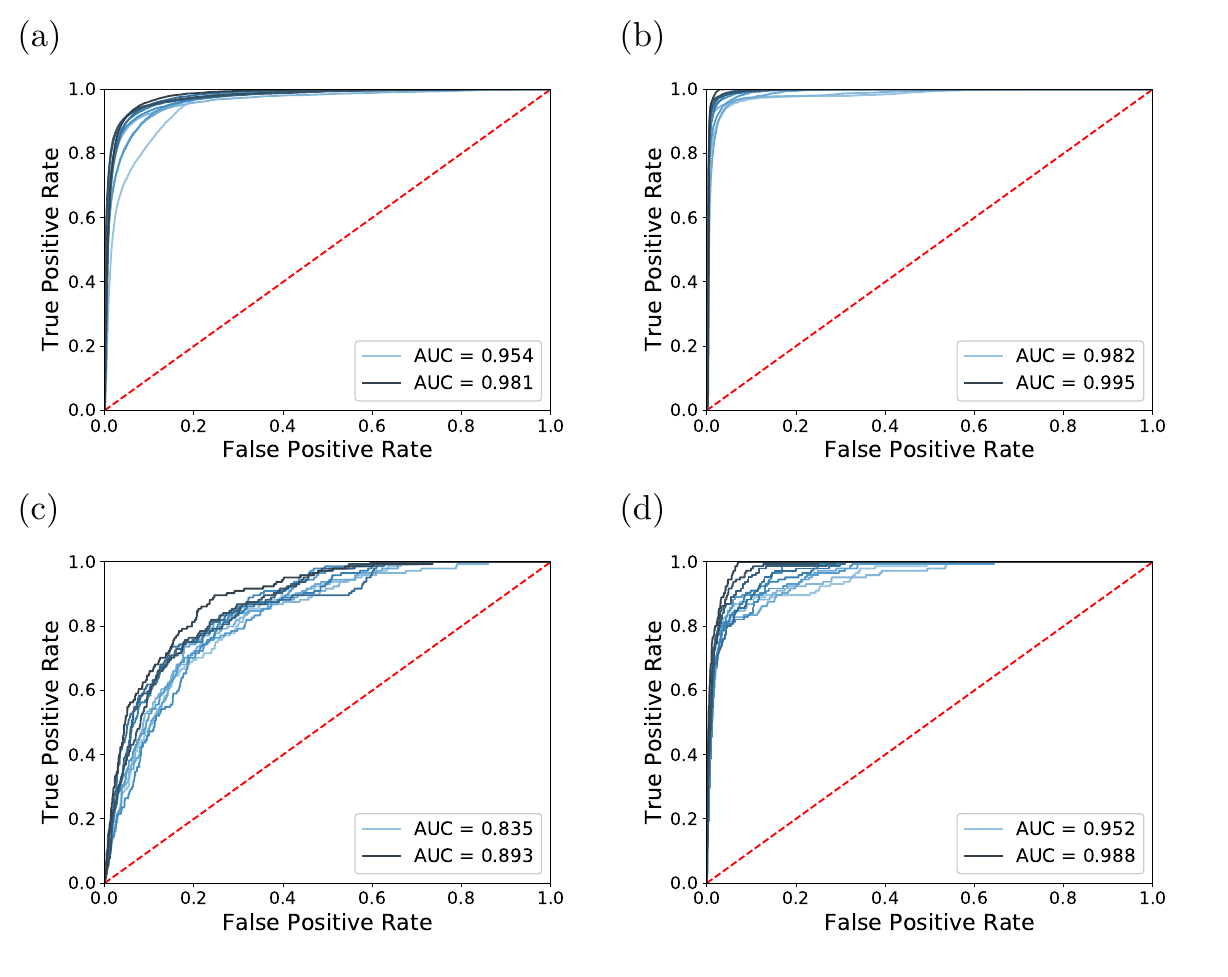}
	\caption{ROC Curves from four test sets ran on 10 repetitions of model H-20. The test sets are: a) Air92, b) Air134, c) Field73 and d) Field88. The lines are coloured in ascending order from lowest AUC value to highest AUC value, with the lowest and highest values labelled in the legend.}
	\label{fig:ROCTest}
\end{figure}

\begin{figure}
	\begin{center}
		\begin{subfigure}[b]{\figwidthfull}
			\centering
			\caption{} \vspace{-0.3cm}
			\includegraphics[width=\figwidthfull]{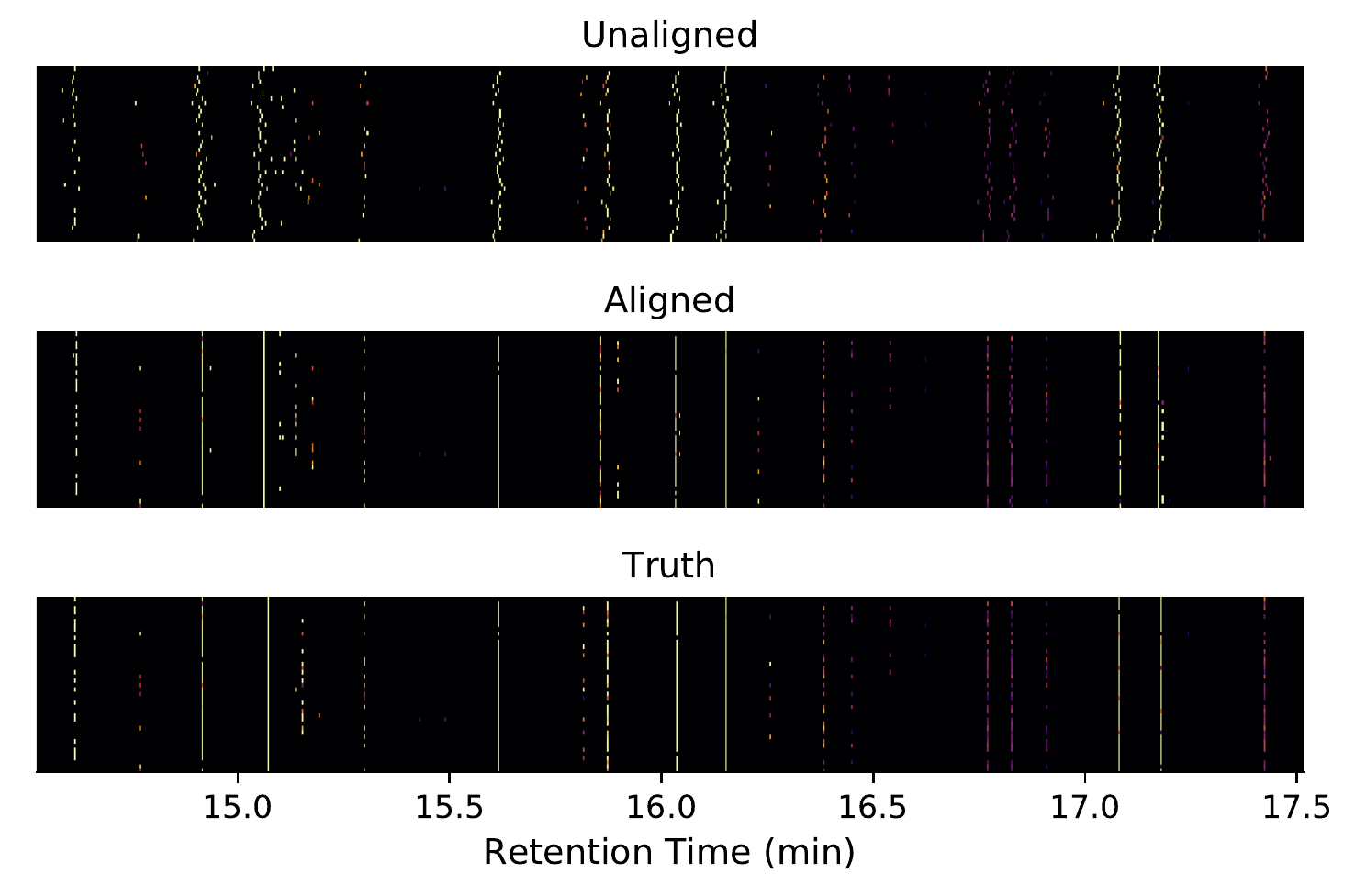}
		\end{subfigure}
		
		\begin{subfigure}[b]{\figwidthfull}
			\centering
			\caption{} \vspace{-0.3cm}
			\includegraphics[width=\figwidthfull]{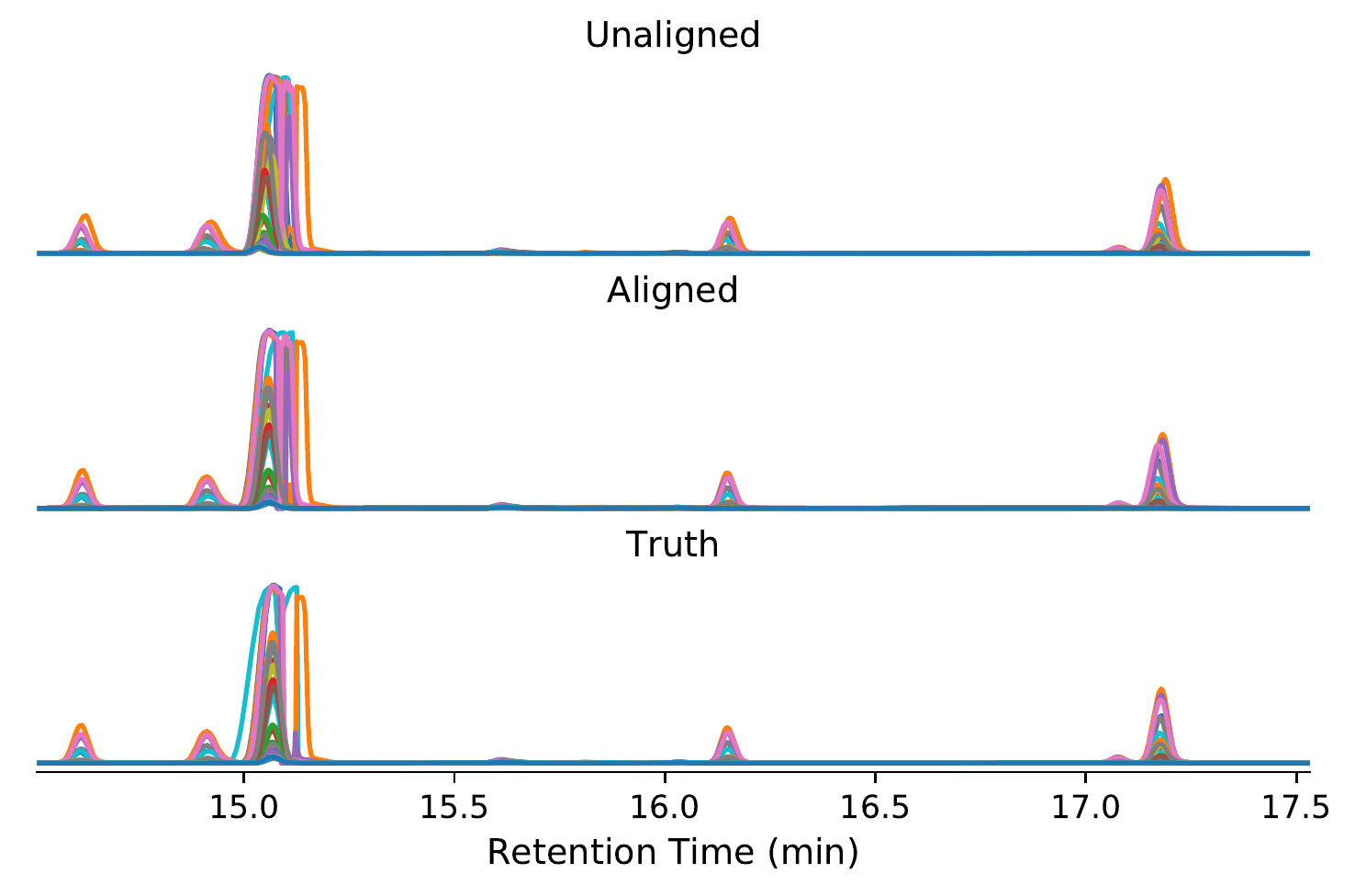}
		\end{subfigure}
	\end{center}
	\caption{Alignment outcome for the data set Air92 using the best performing network of model H-21. For this network, the true positive rate was 0.997 and the false positive rate was 0.172. (a)~Chormatographic image. (b)~Chormatographic plot. 
	} 
	\label{fig:AlignAir92_H21}
\end{figure}

\begin{figure}
	\begin{center}
		\begin{subfigure}[b]{\figwidthfull}
			\centering
			\caption{} \vspace{-0.3cm}
			\includegraphics[width=\figwidthfull]{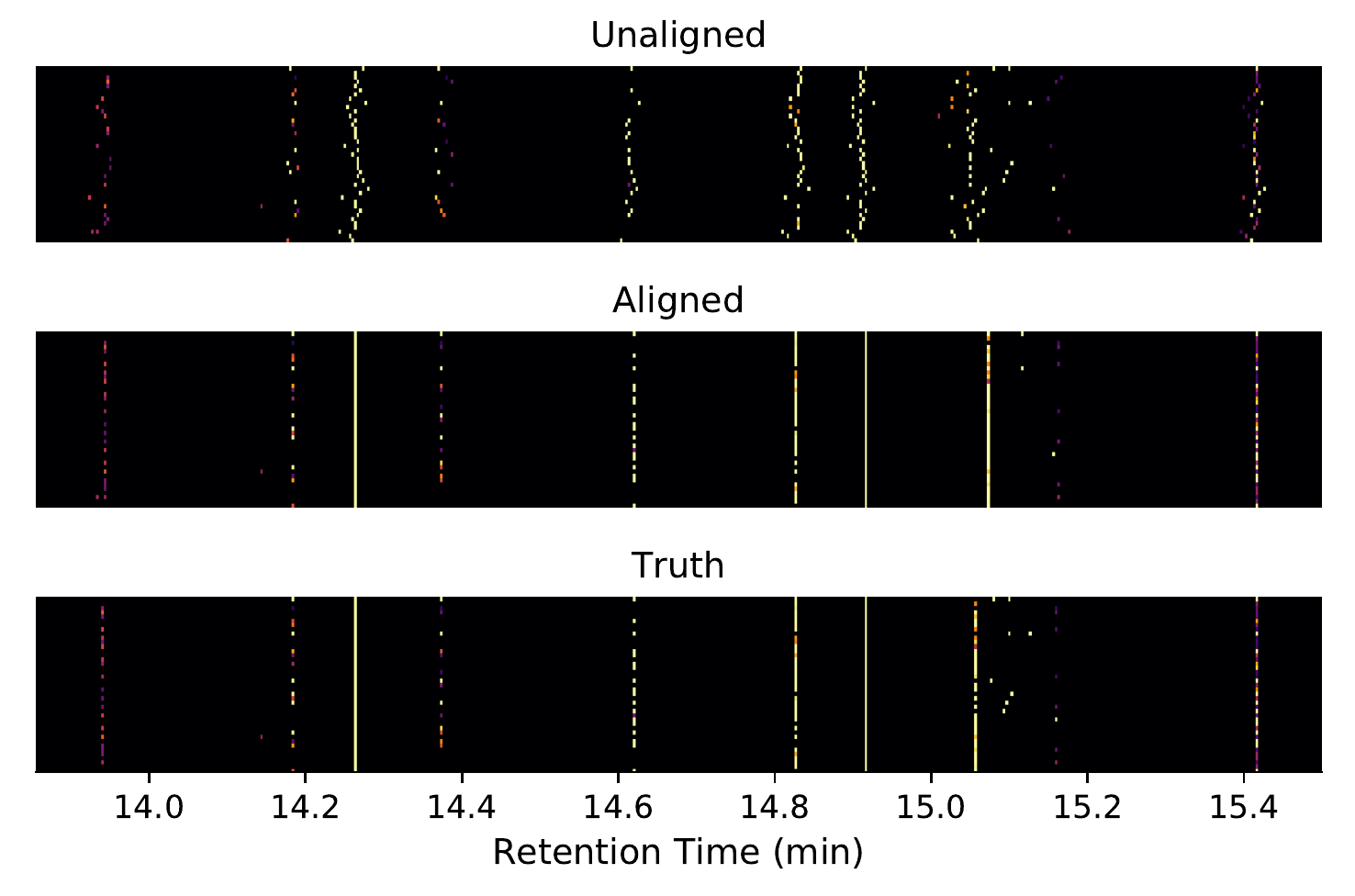}
		\end{subfigure}
		
		\begin{subfigure}[b]{\figwidthfull}
			\centering
			\caption{} \vspace{-0.3cm}
			\includegraphics[width=\figwidthfull]{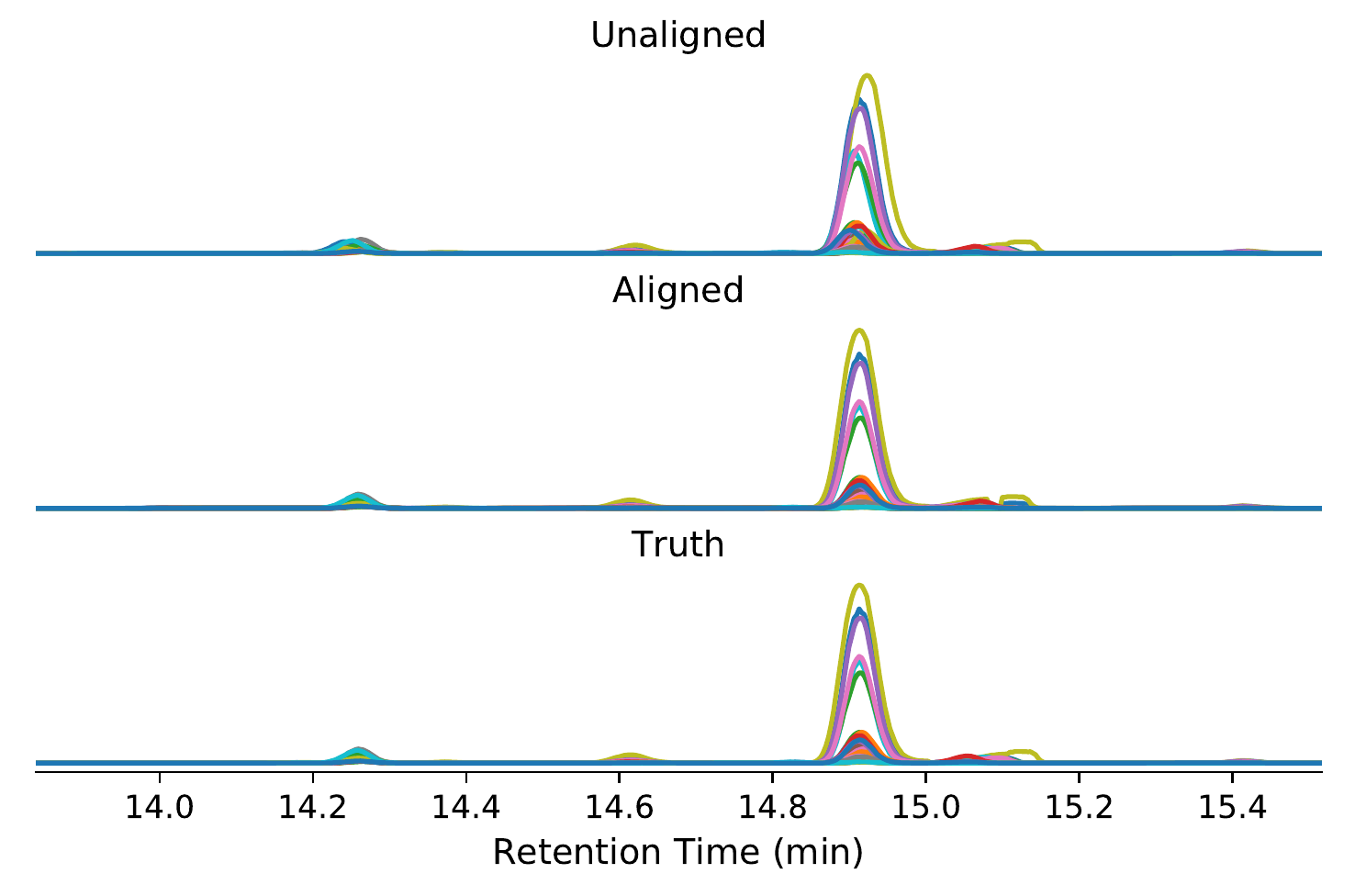}
		\end{subfigure}
	\end{center}
	\caption{Alignment outcome for the data set Air134 using the best performing network of model H-23. For this network, the true positive rate was 0.999 and the false positive rate was 0.074. (a)~Chormatographic image. (b)~Chormatographic plot. 
	} 
	\label{fig:AlignAir134_H23}
\end{figure}

\begin{figure}
	\begin{center}
		\begin{subfigure}[b]{\figwidthfull}
			\centering
			\caption{} \vspace{-0.3cm}
			\includegraphics[width=\figwidthfull]{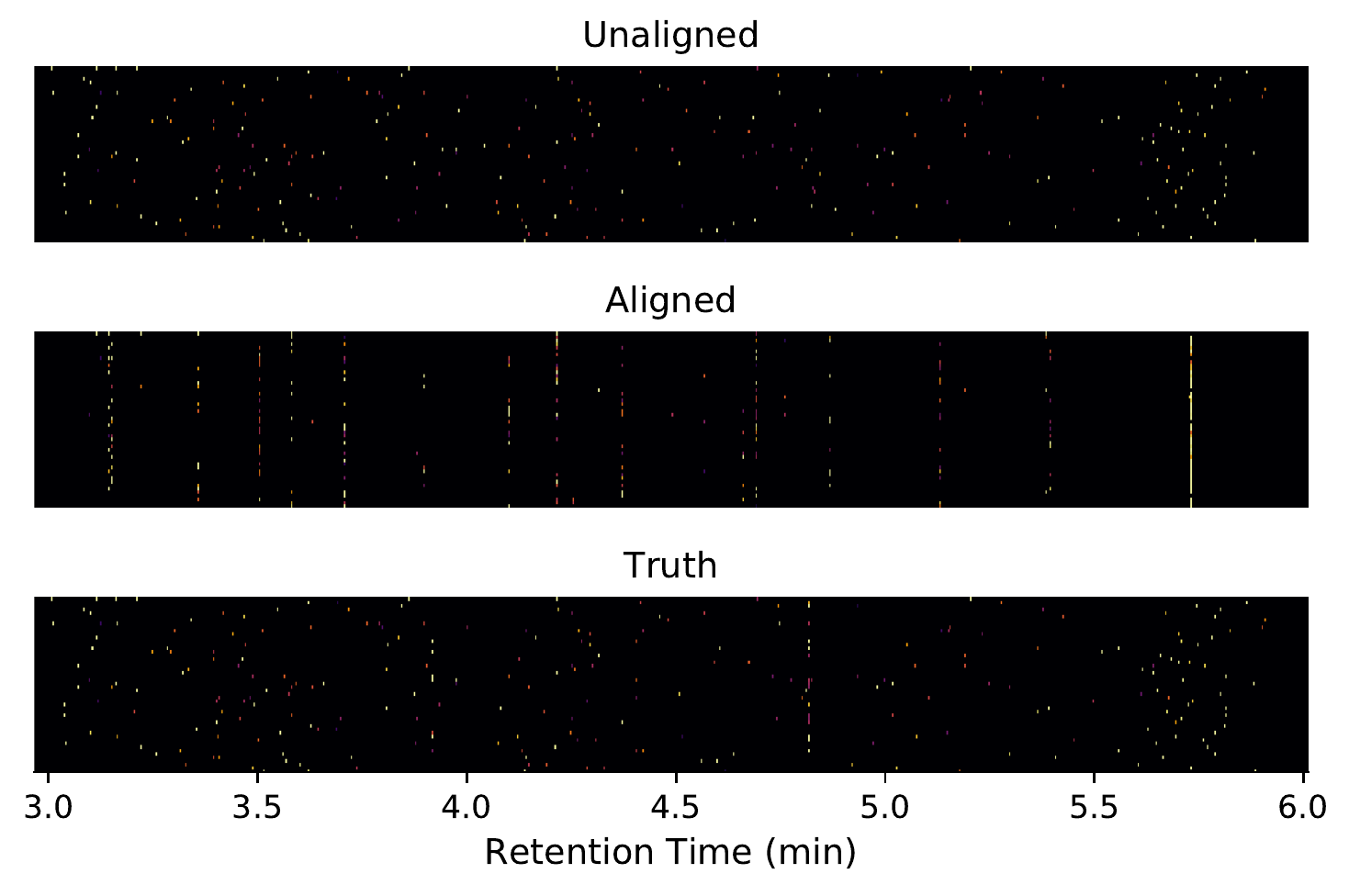}
		\end{subfigure}
		
		\begin{subfigure}[b]{\figwidthfull}
			\centering
			\caption{} \vspace{-0.3cm}
			\includegraphics[width=\figwidthfull]{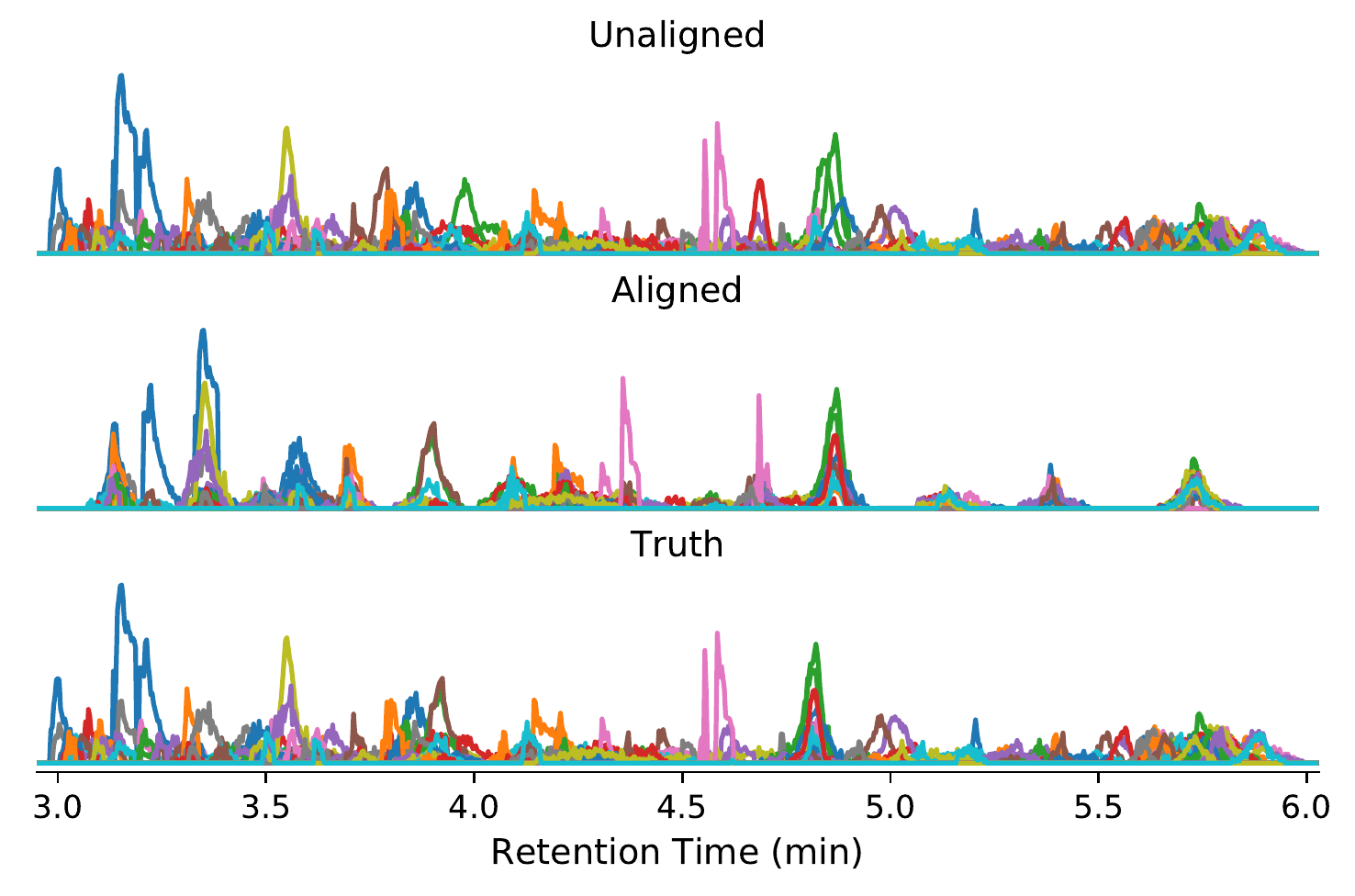}
		\end{subfigure}
	\end{center}
	\caption{Alignment outcome for the data set Field73 using the best performing network of model H-24. For this network, the true positive rate was 0.965 and the false positive rate was 0.381. (a)~Chormatographic image. (b)~Chormatographic plot. Note, only the three thioethers peaks were expertly identified and thus aligned in the ground truth image and plot, these peaks are at RT 3.9 (7 peaks), 4.3 (3 peaks), and 4.8 (16 peaks). 
	} 
	\label{fig:AlignField73}
\end{figure}

\begin{figure}
	\begin{center}
		\begin{subfigure}[b]{\figwidthfull}
			\centering
			\caption{} \vspace{-0.3cm}
			\includegraphics[width=\figwidthfull]{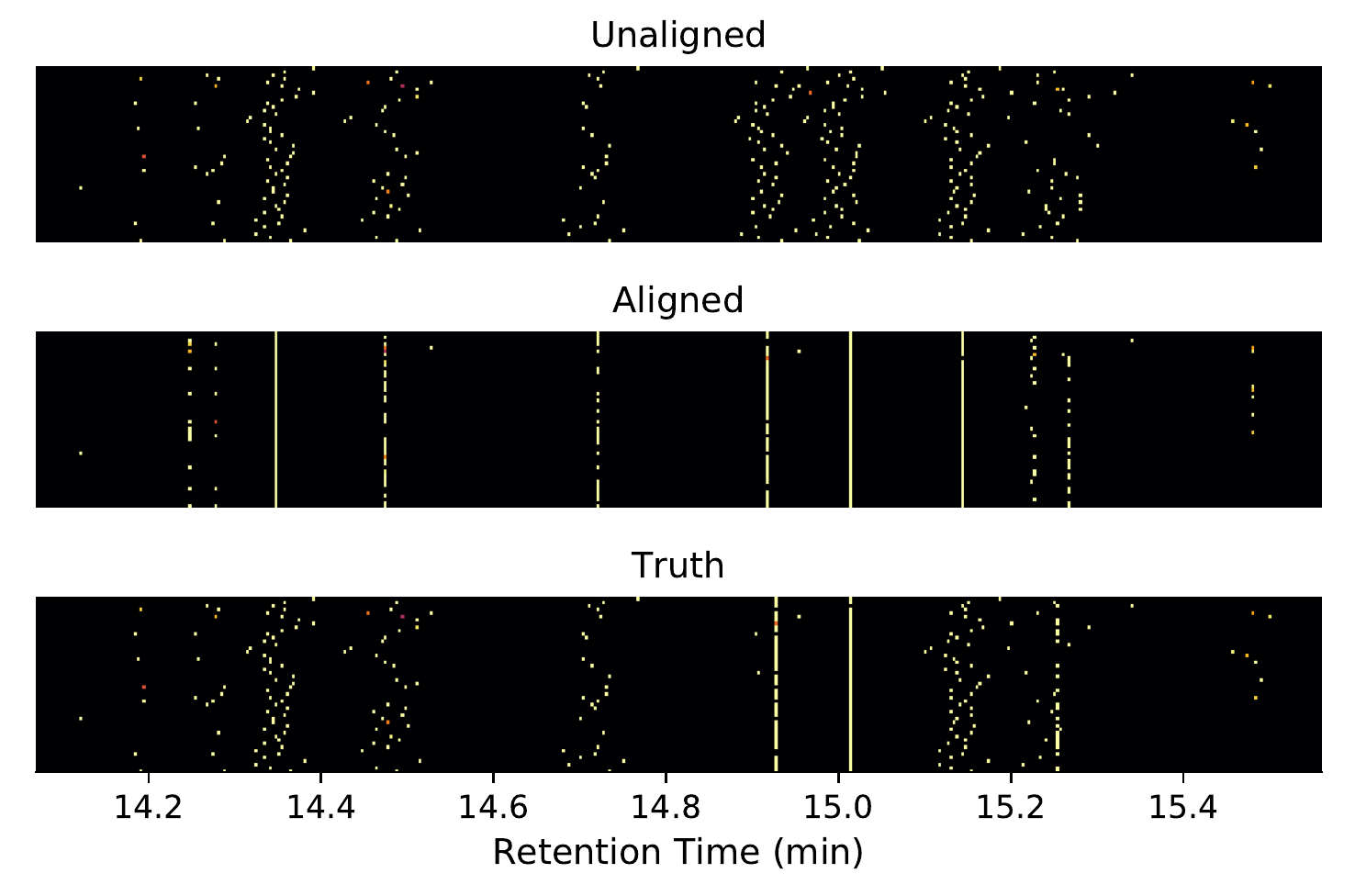}
		\end{subfigure}
		
		\begin{subfigure}[b]{\figwidthfull}
			\centering
			\caption{} \vspace{-0.3cm}
			\includegraphics[width=\figwidthfull]{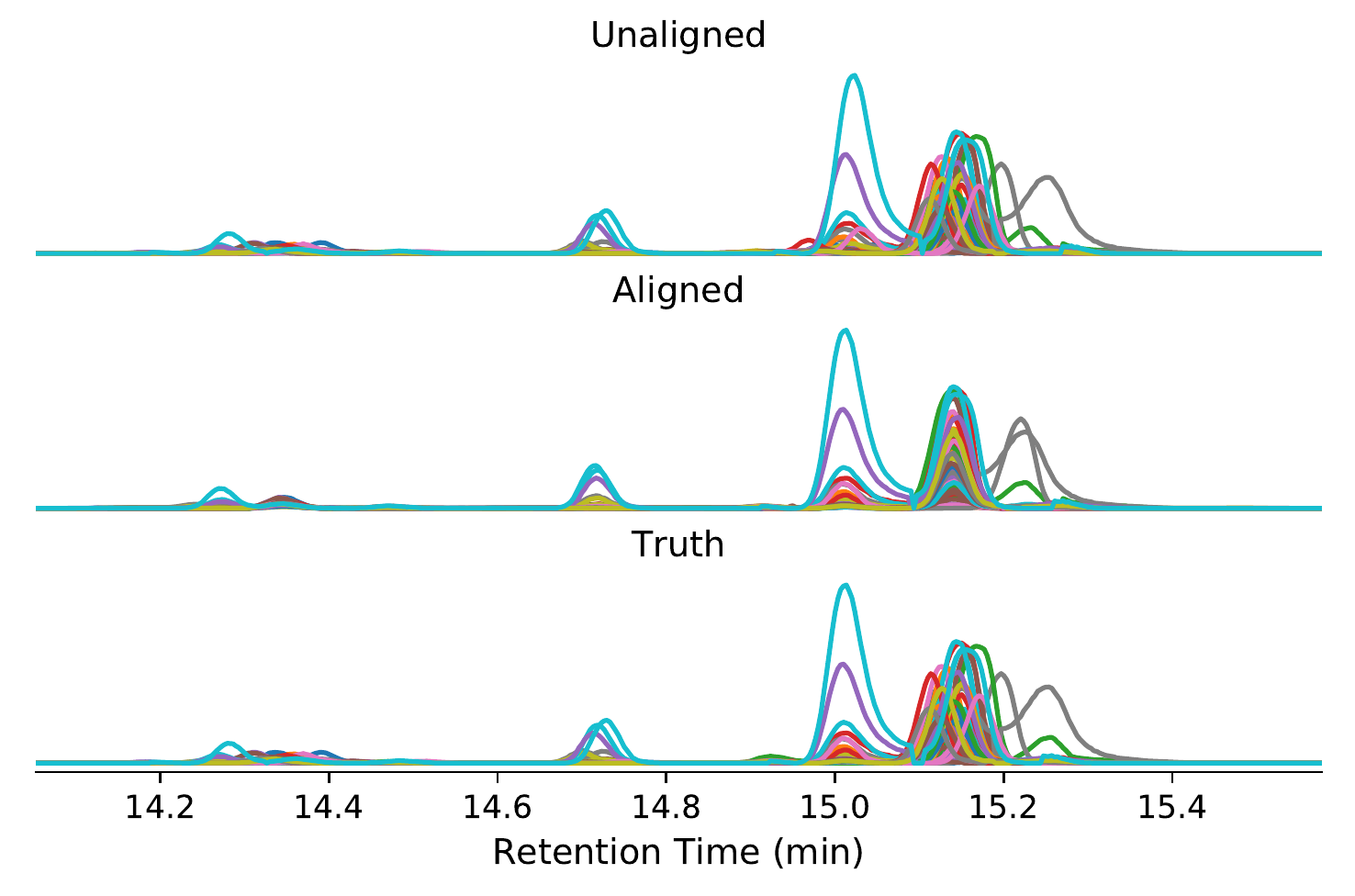}
		\end{subfigure}
	\end{center}
	\caption{Alignment outcome for the data set Field134 using the best performing network of model H-24. For this network, the true positive rate was 0.972 and the false positive rate was 0.039. (a)~Chormatographic image. (b)~Chormatographic plot. Note, only the three cymene peaks were expertly identified and thus aligned in the ground truth image and plot, these peaks are at RT 14.9, 15.0 and 15.25. 
	} 
	\label{fig:AlignField134}
\end{figure}

We now train a new model, H, using all training data from Table~\ref{table:TrainingSamples}. 
We trained six variants of the model: variant 01, the basic model as described in Section~\ref{subsec:method_network}, and five different variants that do not have the peak encoder due to their training time and performance (see Section~\ref{suppsec:modelVariations} for details). Ten networks of each variant were again trained. The training took between 17 and 23 hours for variant 01 and around 30 min for the rest of the variants. 
We report the models' AUC values (Table~\ref{table:AUC_test}), and average TP and FP performances (Supplementary Table~\ref{table:resultsTest}) on each test data set. Examples of ROC curves are shown in Figure~\ref{fig:ROCTest}. We also show alignment results of each test data set using the best performing network (Figure~\ref{fig:AlignField88} and Supplementary~\Cref{fig:AlignAir92_H21,fig:AlignAir134_H23,fig:AlignField73,fig:AlignField134}). 

\begin{table}
	\centering
	\caption{AUC values for six variant of model H (trained using all training data sets) against each test data set. The values report the average over ten repetitions. The values in bold are those with the best performance for the data set.}
	\begin{tabular}{@{}lcccccc@{}}
		\toprule
		& \multicolumn{6}{c}{Models} \\
		\cmidrule(lr){2-7}
		Data Set & H-01 & H-02 & H-20 & H-21 & H-23 & H-24 \\
		\midrule
		Air92 & 0.9669 & 0.9652 & \textbf{0.9718} & 0.9695 & 0.9702 & 0.9711 \\
		Air134 & 0.9923 & 0.9903 & 0.9912 & 0.9907 & \textbf{0.9933} & 0.9929 \\
		Field73 & 0.8532 & 0.8514 & 0.8557 & 0.8385 & 0.8566 & \textbf{0.8581} \\
		Field88 & 0.9587 & 0.9628 & 0.9679 & \textbf{0.9680} & 0.9632 & 0.9607 \\
		Field134 & 0.9846 & 0.9834 & 0.9830 & 0.9853 & 0.9820 & \textbf{0.9859} \\
		\midrule
		Overall & 0.9512 & 0.9506 & \textbf{0.9539} & 0.9504 & 0.9531 & 0.9537 \\
		\bottomrule
	\end{tabular}%
	\label{table:AUC_test}%
\end{table}%

The performances of model H variants on the test data show the models generally performed very well, with all models give AUC between 0.96 and 1 for all data sets except Field73. Overall, the variant H-20 had the best performance, closely followed by H-24. Both variants do not have the peak encoder and with slightly simplified architecture in either the mass encoder or the chromatograph encoder. These results show the peak encoder does not add to the overall performance of the model. 
All variants have high average true positive rates (almost 99\% for air data and between $\sim$ 84\% and 96\% for the field breath data). However, the false positive rates are also relatively high, under 10\% for three data sets (Air134, Field88 and Field134) and around 30\% for the Air92 and Field73 data sets. 

As masses 92 and 134 had not been included in the training data sets, the high AUC and TP results showed the model is able to generalise to data it hasn't seen before. The high FP results show that this generalisation has a limit, as the networks were not able to distinguish between neighbouring peaks with similar profiles. 
An example of this is the case of the Air92 results, which had a high false positive rate largely due to two cases of pairs of peaks occurring very close to each other at 15.8 and 15.9 min as well as 17.1 and 17.2 min (see Supplementary Figure~\ref{fig:AlignAir92_H21} for alignment results). 
Using the existing hierarchical clustering algorithm tends to merge the two peaks together in these cases. We thus introduced a condition in the grouping algorithm to separate peaks from the same sample into different groups based on the order of their retention times. Taking this into account means double peaks (due to possible co-elusion) will be split into different groups (as in groups of peaks at 15.1 min and 17.2 min). 
Future work may improve the resolution of the chromatogram encoder to distinguish these peaks occurring close together by learning a larger number of convolutional filters. These models used a small number of convolutional filters (e.g.\ 6 filters are learnt at the first layer) since it was only expected to provide supplementary information to the peak encoder.

There is a large difference between the FP rate of Field73 and Field 88 data sets, both containing the same three thioethers in the samples. This difference in FP is due to the difference in number of total peaks detected and the number of thioether peaks identified. For the Field73 data set, of the 270 peaks detected in the 50 samples, only 26 thioether peaks were identified (see Table~\ref{table:TestData}). In contract, while the Field88 data set also only had 27 identified thioether peaks, there were only 120 total peaks detected. There are almost five times as many combinations of peaks in Field73 compared to Field88, making it a more difficult data set for alignment.  

We observe that ordering the latter four variants according to their overall performance, we have H-20, H-24, H-23, H-21. 
This performance is slightly different from the results on training data (Table~\ref{table:VariationModels2}), these variants are ordered as H-20, H-23, H-21, H-24 due to their overall performance. 
The difference between the two sets of results show that performance is partially dependant on the data set, although this might improve when increased training data are used.

\subsection{Prediction Run-time Analysis}
\label{subsec:res_runtime}

\begin{figure}
	\centering
	\includegraphics[width=\figwidthfull]{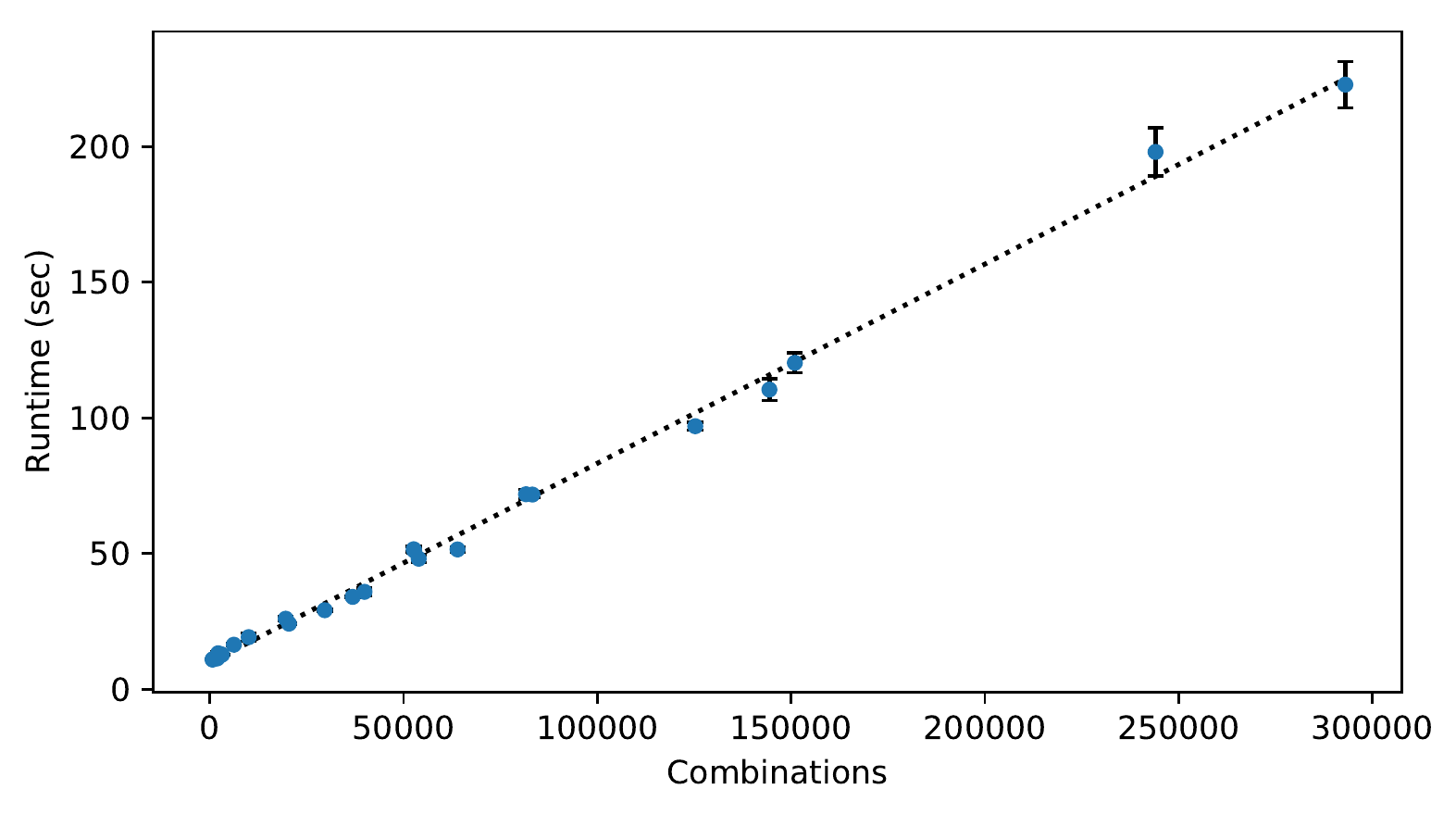}
	\caption{Number of combinations vs runtime (seconds). Data points are gathered from the running of data sets as well as the values in Table~\ref{table:RuntimeWholeSamples}}
	\label{fig:RuntimeVsCombinations}
\end{figure}

Once the model is trained and tested, it can be used on new data sets, therefore, the training time for the model is irrelevant to alignment, and only the prediction time is of concern to users of GC-MS data sets.  
The prediction time of the model depends linearly on the number of peak combinations (Supplementary Figure~\ref{fig:RuntimeVsCombinations}). This, in the worst case, is quadratically related to the number of peaks present. However for longer segments most of the peak combinations fall outside the 3 minute comparison cut-off and are immediately assigned a probability of 0. Table~\ref{table:RuntimeWholeSamples} shows the number of combinations that were found in two and ten SICs of air and breath. These combinations are given at a cutoff of one minute, three minutes and five minutes and shows that the speed of the algorithm can be adjusted by tuning the maximum range the retention time shifts are believed to occur within. The average run-time comparing two SIC took between 10 to 20 seconds.

\begin{table}
	\centering
	\caption{Runtime of whole SIC (full 50 minute retention time). This compares two SIC together and ten SIC together. These run-times are assessed using the simpler model D-02, since the peak encoder component does not seems to provide significant benefits. The computer used for assessment has a dual core 2.9GHz Intel i5-3380M processor with 8 GB of memory.}
	\label{table:RuntimeWholeSamples}
	\resizebox{\textwidth}{!}{%
		\begin{tabular}{@{}lrrrrrrrrrrr@{}}
			\toprule
			&  &  &  &  \multicolumn{2}{c}{1 Min Cutoff} & \multicolumn{1}{c}{} & \multicolumn{2}{c}{3 Min Cutoff} & \multicolumn{1}{c}{} & \multicolumn{2}{c}{5 Min Cutoff} \\ \cmidrule(lr){5-6} \cmidrule(lr){8-9} \cmidrule(l){11-12} 
			SIC& & Peaks &  & \multicolumn{1}{c}{Combinations} & \multicolumn{1}{c}{Time (sec)} & \multicolumn{1}{c}{} & \multicolumn{1}{c}{Combinations} & \multicolumn{1}{c}{Time (sec)} & \multicolumn{1}{c}{} & \multicolumn{1}{c}{Combinations} & \multicolumn{1}{c}{Time (sec)} \\ \midrule
			Air, m/z = 103&x2 & 271 &  & 2,100 & 13.3 $\pm$ 0.5 &  & 6,200 & 16.4 $\pm$ 0.4 &  & 10,000 & 19.2 $\pm$ 1.6 \\
			Air, m/z = 103&x10 & 1436 &  & 53,900 & 48.1 $\pm$ 1.5 &  & 151,000 & 120.3 $\pm$ 3.7 &  & 244,000 & 198.0 $\pm$ 8.9 \\
			Breath, m/z = 103&x2 & 155 &  & 800 & 11.1 $\pm$ 0.3 &  & 2,000 & 12.0 $\pm$ 0.2 &  & 3,100 & 12.7 $\pm$ 0.2 \\
			Breath, m/z = 103&x10 & 748 &  & 20,400 & 24.2 $\pm$ 0.2 &  & 52,600 & 51.6 $\pm$ 1.0 &  & 83,200 & 71.7 $\pm$ 1.3 \\
			Breath, m/z = 73&x2 & 172 &  & 700 & 10.9 $\pm$ 0.1 &  & 1,900 & 11.4 $\pm$ 0.5 &  & 3,000 & 12.8 $\pm$ 0.1 \\
			Breath, m/z = 73&x10 & 897 &  & 19,600 & 26.0 $\pm$ 0.9 &  & 52,700 & 51.2 $\pm$ 0.9 &  & 81,600 & 71.9 $\pm$ 1.8 \\ \bottomrule
		\end{tabular}%
	}
\end{table}

\subsection{Comparison with Existing Algorithm}
\label{subsec:res_compare}

We compare our alignment results with two existing algorithms: the correlation optimized warping (COW) algorithm~\cite{Nielsen1998,TomasiJChemom2004,SkovJChemom2007}\footnote{The MATLAB code for COW was downloaded from http://www.models.life.ku.dk/dtw\_cow}, 
and the GCalignR algorithm~\cite{Ottensmann2018}. 
We investigated many other existing algorithms, however, most are unsuitable to compare with our algorithm -- in some of the exiting algorithms the peak alignment module was not designed to be used on its own~\cite{SmithAC2006,DomingoAC2016}; some algorithms do not have source code available~\cite{FuJCA2017}; some requires a specific MATLAB toolbox that we do not own~\cite{YangJCA2018,ZhengJCA2013}; and another built on proprietary software~\cite{CouprieJCA2017}. 
Both of the methods we chose to compare our algorithm with have source code readily available online, and easily adapted to use on any datasets. 

The main difference between the implementation of the existing algorithms and ChromAlignNet is the user defined parameters. There are two main parameters to be defined in ChromAlignNet: the number of training epochs and the split in the training data set for validation during training. These parameters are relatively easy to choose and set. Another parameter to set is the batch size for training, which determines the efficiency of the algorithm on the computer, thus does not affect the final network. At prediction (i.e.\ alignment) time, there is only one parameter to set, that is the cut-off time between pairs of peaks to compare. This is relatively easy to set as most GC-MS users will know the maximum drift in a data set. Moreover, a generous cut-off time does not impact the result too much. 

COW requires the users to define two parameters: the segment length (the chromatograms are split in a number of segments during alignment) and the flexibility (how far the boundaries between segments are allowed to move). These parameters are selected on a trial and error basis by visual inspection of the sample chromatograms,
which means it is very time consuming for the user during implementation.
Efforts have been made at automating the parameter selection in COW~\cite{TomasiJChemom2004,SkovJChemom2007},
however, they still require iterative searches of the parameters by the algorithms. 
Additionally, 
COW also requires users to select a reference chromatogram as the target for alignment.
There are several methods for selecting the reference chromatogram~\cite{DaszykowskiJCA2007}, but no single method is optimal.

GCalignR requires three parameters corresponding to its three stage alignment process, which controls how far each peak is allowed to shift. The first stage is a linear shift of the full chromatogram in comparison to a reference sample, up to the value given by the parameter \verb|max_linear_shift|. The second step aligns each peak with the average RT of the surrounding peaks, grouping peaks together up to a maximum \verb|max_diff_peak2mean|. The third step uses another parameter \verb|min_diff_peak2peak| to account for adjacent groups that should be combined.
Although these time based parameters are more intuitive to set than those in COW, the strict nature of these cutoffs means that alignment outcome can vary significantly when these parameters are changed.

We implemented the existing algorithms on two data sets: (1) Air92, an easy data set, and (2) Field88, a complex data set of human breath. Both data sets are part of the test data as described in Table~\ref{table:TestData}. 
For COW, we implemented the automatic reference and parameter selection algorithms~\cite{TomasiJChemom2004,SkovJChemom2007}.
We selected the target chromatogram based on maximum cumulative product of correlation coefficients. For parameter selection, we used the default optimisation options with a custom range of slack and segment.
For GCalignR, we used the defaults provided by the algorithm.

To compare the alignment results, we modified the assessment metrics (see Supplementary Materiel Section~\ref{suppsec:metricGrouping} for full details). This is because for ChromAlignNet, we compare all possible pairs of peaks in the dataset, but to compare the final alignment results between the methods, we need to only compare how well a peak has aligned to the group. We calculate and present the TP rate and the False Discovery Rate (FDR) of the alignment results for the three methods (Table~\ref{table:compareMethods}). 

\begin{table}
	\centering
	\caption{Performance of three different alignment methods on three different data sets. The value $N$ next to the data set name indicates how many peaks in the data sets were given a group number, thus have ground truth in alignment. }
	\label{table:compareMethods}%
	\begin{tabular}{@{}lrrrrrrrr@{}}
		\toprule
		& \multicolumn{2}{c}{ChromAlignNet} &       & \multicolumn{2}{c}{COW} &       & \multicolumn{2}{c}{GCalignR} \\
		\cmidrule(lr){2-3} \cmidrule(lr){5-6} \cmidrule(l){8-9} 
		& \multicolumn{1}{c}{TP} & \multicolumn{1}{c}{FDR} &       & \multicolumn{1}{c}{TP} & \multicolumn{1}{c}{FDR} &       & \multicolumn{1}{c}{TP} & \multicolumn{1}{c}{FDR} \\
		\midrule
		Air 92 (N=525) & 0.952 & 0.010 &       & 0.442 & 0.013 &       & 0.962 & 0.008 \\
		Field 88 (N=27) & 0.963 & 0.071 &       & 0.556 & 0.063 &       & 0.481 & 0.000 \\
		\bottomrule
	\end{tabular}%
\end{table}%

The alignment results for Air92 data set show GCalignR has the best TP rate, with ChromAlignNet a close second, and both have much better performance than COW. The result for ChromAlignNet here is somewhat surprising, since it achieved a high TP rate in the pairwise comparison results for this data set. However, it also achieved a relatively high FP rate for this data set. The latter would have contributed to the final alignment TP rate to only be 95\% (from a TP rate of 99\% for pairwise comparison), that is, when multiple nearby pairs of peaks (where a common peak appears in all pairs) all have high prediction value of peaks belong to the same group, the grouping algorithm could misalign the peaks. 

For the Field88 data set, ChromAlignNet has the best results while both COW and GCalignR performed poorly. The samples in this data set differ substantially from each other. The RT of the same compound peak may be displaced by several peak widths in different samples.
In some cases, the consequent peak from one sample can have an earlier RT than the peak of interest in the second sample. This sort of complex data set is where rule based algorithm fails in peak alignment, as most algorithms assume the peaks are nearby and do not compare all pairs of peaks in a chromatogram segment.

\section{Conclusion}
\label{sec:discussion}

We introduced a deep-learning based method, ChromAlignNet, for alignment of GC-MS data. We quantitatively analysed the performance of our method and compared it with existing algorithms. 
In general, ChromAlignNet has very good alignment prediction outcomes when comparing two peaks, we achieved AUC between 0.95 and 1 for most of our test data sets. The performance of our model is only hampered when tested against very complex data sets, which give AUC $\sim0.85$. Comparing our final alignment of peaks in sets of chromatograms against existing methods, we found ChromAlignNet gives comparable results for simple data sets, and much better accuracy when aligning very complex data sets. 

Comparing to existing methods, ChromAlignNet's main advantage is that the model is learned from data. This means no assumptions were made when designing the algorithm other than the basics of GC-MS data. Unlike existing algorithms, we do not assume the samples for alignment are fairly similar (e.g. in algorithms such as GCalignR~\cite{Ottensmann2018} and CAM~\cite{ZhengJCA2013}), such that there is some correspondence between all peaks and there are few missing peaks. The model also isn't given information such as the largest single value after matrix decomposition represents chemical information (e.g. in FSA-MS~\cite{YangJCA2018}). This lack of assumption from ChromAlignNet means it is free to learn the representation from the data presented, and the models can improve with appropriate data as we see in the results in Table~\ref{table:AUC_training}. 

Another difference between ChromAlignNet and most existing algorithms is that no reference chromatogram is required for alignment, instead we compare peaks pair-wise. Since it is almost impossible to have a perfect reference chromatogram that contains examples of all peaks in the target chromatograms. The need for reference chromatogram is one of the reasons for low accuracy in existing algorithm when aligning very complex data sets, when the algorithms assume that every peak in target chromatogram need to be aligned to corresponding peak in the reference chromatogram. 

ChromAlignNet is also extremely easy to use with almost no user input of parameters compared to most existing algorithms. This is because the model learns the best parameters to compare peaks during training. Therefore users do not need to be familiar with the algorithm to run the prediction and alignment. 

However, the high false positive rates show there are still many areas for improvement. We discuss the shortcomings of the tool and future work of improvement in Supplementary Section~\ref{supp:future}.

\section*{Acknowledgement}

We would like to thank the CSIRO Scientific Computing team for the use of their supercomputer clusters in performing the experiments for this paper. We acknowledge the Sydney Informatics Hub and the University of Sydney’s high performance computing cluster Artemis for providing the high performance computing resources that have contributed to the research results reported within this paper.

We thank Dr Jacqui Poldy for her invaluable comments on the manuscript. We thank Ms Julie Cassells for demonstrating the current methods of peak detection and compound identification, as well as the identification of compounds and their retention times used in this paper. We thank Dr Alisha Anderson for help with ethics and permission for use of the data sets. XRW thanks Dr Amalia Berna for discussion of possible peak alignment ideas. 

\section*{Research Data}

The data for this paper is available online~\cite{Wang2019}. 

\section*{Software}

ChromAlignNet was implemented in Python 3.6 with the Keras library. The hierachical clustering algorithm for group assignment after pairwise alignment is from Python's SciPy library. The ROC curve and AUC calculations are from Python's scikit-learn library. The code is available online (https://github.com/mili7522/ChromAlignNet).

The code for converting from XML to csv files, and for peak detection in a SIC were implemented in Matlab. The codes are also available online (https://github.com/rosalind-wang/GCPeakDetection). 

\section*{Data Analysis}

Training and batch prediction (prediction using all models on all possible data sets) were conducted on CSIRO's GPU cluster and the Sydney University CPU cluster. The CSIRO GPU cluster are composed of dual 14 core compute nodes each with 256GB RAM and 4 Nvidia P100 GPUs. The Sydney University CPU cluster are composed of a mix of Intel's Haswell, Broadwell and Skylake compute nodes based on dual socket servers with up to 192GB RAM per node (excluding High-Memory nodes).

Further calculations, such as individual predictions, were conducted using a dual core 2.9GHz Intel i5-3380M processor with 8 GB of memory running Windows 10 Pro and a dual core 3.3GHz Intel Core i7 processor with 16 GB of memory running macOS 10.13.


\bibliography{reference,references-gcms}

\pagebreak
\beginsupplement

\section{Supplementary Text}

\subsection{Deep Learning}
\label{suppsec:deeplearning}

Deep learning~\cite{Lecun2015}, also called deep neural networks, is a type of artificial neural network and a specific subfield of machine learning. Since around 2012, it has become one of the most popular machine learning methods for many types of complex problems such as image classification~\cite{Lecun2010,Osadchy2007,Krizhevsky2012}, speech recognition~\cite{Raina2009,Hinton2012}, language translation~\cite{Sutskever2014}, handwriting transcription~\cite{Chellapilla2006a,Chellapilla2006b}, autonomous driving~\cite{Hadsell2009}, to name a few. The success of deep learning stems from its ability to learn successive layers of increasingly meaningful representation of the raw data. Deep learning systems build up complex concepts by composing simple but non-linear modules (called `neurons') that transform the simpler representation of one layer into the slightly more abstract representation of the next layer~\cite{Lecun2015,Goodfellow2016}. For example, in the field of computer vision, successive layers learn to detect edges, shapes and eventually entire objects given only pixel values. The key aspect differentiating deep learning from other conventional machine learning techniques is that the representations (or features) are not designed by engineers nor are they selected by experts, rather, they are learnt from the data through the neural network's learning algorithms~\cite{Lecun2015}. 

One reason for the success of deep learning networks in solving a large range of complex problems is the diverse set of models, or network architecture, in deep learning compared to traditional machine learning algorithms. The general structure can be described as neurons arranged into an input layer, an output layer and several intermediate, or hidden, layers~\cite{Goodfellow2016}. However, the number of hidden layers can vary up to several hundreds or even thousands, with any number of neurons in each layer. Further, the neurons can be connected in many different ways. This flexibility in network architecture allows researchers to build deep learning models that best suit their application.

\begin{figure}[h]
	\centering
	\includegraphics[width = \figwidthfull]{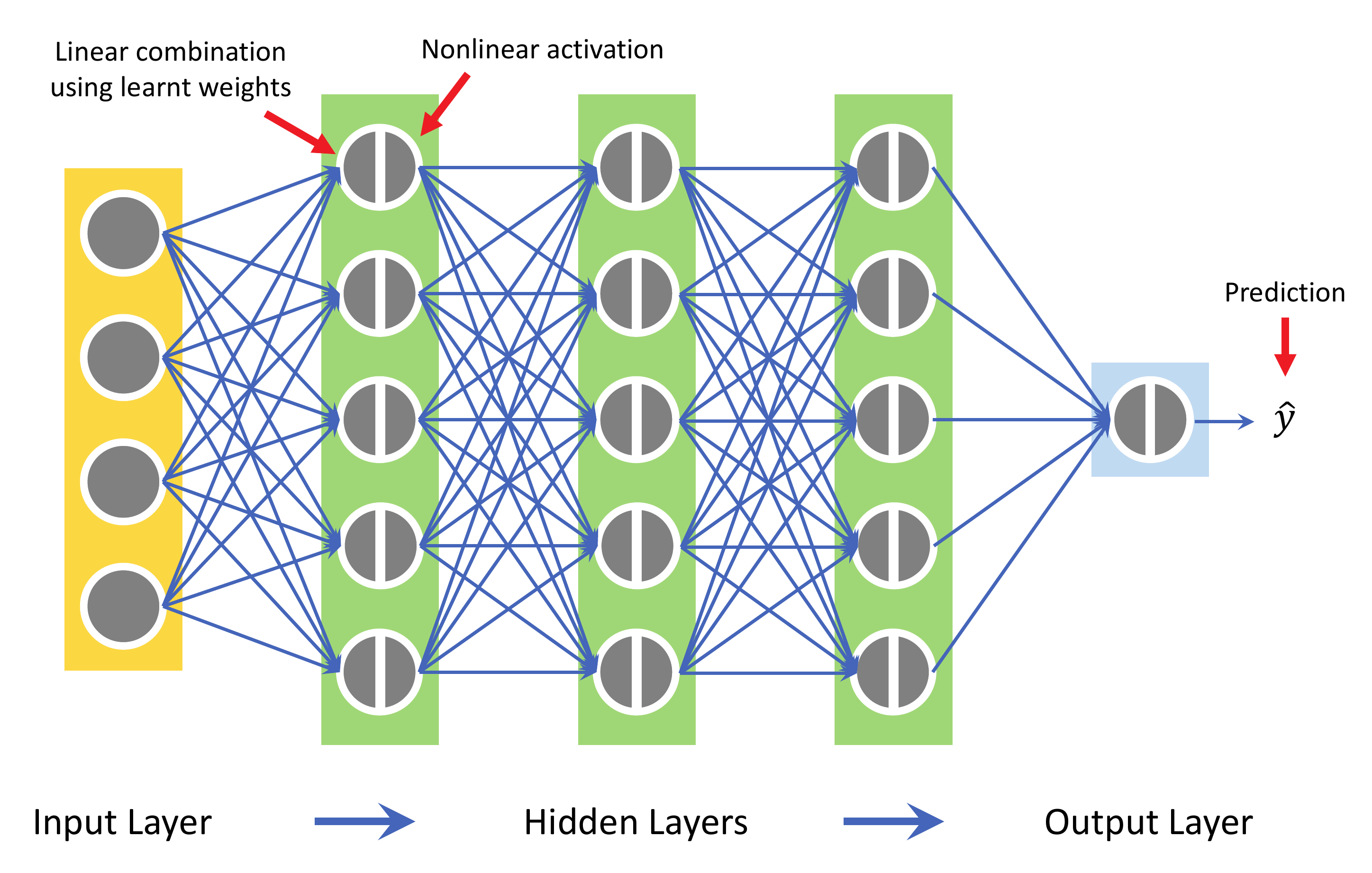}
	\caption{Fully connected, feedforward network structure, consisting of several layers of neurons, each of which is a linear combination of the previous layer followed by a non-linear activation function}
	\label{fig:FeedforwardNetwork}
\end{figure}

The simplest and most common neural network model is the \emph{fully connected feedforward network} (Figure~\ref{fig:FeedforwardNetwork}). Feedforward refers to the flow of information through the layers of the network in a single direction, from input to output. There are no cycles formed by connections between neurons in the same layer or by information passed back from later layers of the network.
Meanwhile, a fully connected network refers to all possible connections being present between the neurons of adjoining layers. That is, each neuron in the hidden and output layers receives inputs from all the neurons in the layer before it.
During computation, the neurons in each layer calculate a weighted sum of their inputs and pass this result through a non-linear `activation' function. 
These networks are then trained with data to learn the weights of the connections -- in other words how important the data from one neuron is to the connected neuron in the next layer. In a typical modern deep learning network, there may be hundreds of millions of weights to be learnt. 

By relaxing the requirements of either being fully connected or only feedforward, specialised deep neural network architectures can be formed, such as convolutional neural networks and recurrent neural networks. This generally reduces the number of weights that need to be learnt since the connections between layers are sparse and weights can be shared.
\emph{Convolutional neural networks} (CNN)~\cite{Lecun1989} are designed for data that come in the form of multiple arrays, for examples, 1D time series signals from various sensors, 2D coloured images or 3D coloured videos. 
\emph{Convolutional} layers in a CNN replace the weighted sum with the convolution operation as the mathematical linear operation between the layers. This means that instead of learning a unique weight between each pair of neurons, only a small number of weights arranged into a filter are required. Filters act locally and produce output `feature map' -- for example, an edge in one part of an image is the same as an edge in another part so the concept of an edge detector only needs to be learnt once to be applied across the whole image~\cite{Lecun2015}.
Convolution layers are often alternated with \emph{pooling} layers, whose role is to merge similar features, thus reducing the amount of information in order to train the model more efficiently. For example max pooling~\cite{Zhou1988} replaces a block of values with the maximum within that block. The many stages of convolution and pooling layers extract successively more features from the raw data, which are used by the final stage of standard (fully connected) neural network layers for tasks such as classification.

\emph{Recurrent neural networks} (RNN)~\cite{RumelhartNature1986} are a type of neural network that are specialised for processing sequential data, thus are common in fields such as speech recognition. 
Unlike feedforward networks, RNNs contain connections within the same layer, forming loops that allow them to make use of their current internal state to help process a sequence of inputs.
Since the same RNN block with the same weights is used over the whole sequence, the number of parameters that need to be learnt is reduced compared to treating different positions along the sequence as unique. One development in RNN is the Long Short Term Memory (LSTM) network~\cite{Schmidhuber1997}, which  maintains a memory unit as well as the internal state. The goal of this memory unit is to retain inputs for a long time. This is effective in machine translation tasks which must keep track of grammar components separated by many words. 

While all the different neural networks described above were introduced in the 1980s and 1990s, it wasn't until 2012 that deep learning gained wide spread notice and many successes. These successes were made possible by a combination of advancements in hardware and software. In terms of the hardware, the introduction and adaptation of graphics processing units (GPUs) in desktop and supercomputer clusters allowed researchers to train networks 10 or 20 times faster~\cite{Raina2009}. On the software front, the introduction of rectified linear units (ReLU)~\cite{Glorot2011} for neurons and a new regularisation technique called dropout~\cite{Srivastava2014} sped up the learning of networks with many layers. Dropout is a method that randomly removes a number of neurons in the network during each training iteration. This simplifies the network during training and forces redundant representations to be learnt.

\subsection{Data Used}
\label{suppsec:data}

\emph{Ambient air} samples obtained during trials of controlled human malaria infection (CHMI). Data from studies with two different malaria species were used. In the first trial, volunteers were infected with \emph{Plasmodium falciparum}, and in a second trial, volunteers were infected with \emph{P. vivax}. The details of the trials have been described previously~\cite{BernaJBR2018}. During these studies a total of 62 and 41 samples of ambient air were collected, respectively. The studies were approved by the Queensland Institute of Medical Research Human Research Ethics Committee (QIMR‐HREC) (P1479) and endorsed by the CSIRO Health and Medical Research, Human Research Ethics Committee (proposal numbers: 11/2016 and 19/2016).

\emph{Healthy control human breath} samples were collected from nine volunteers at different times of the day over several days. Details of the study were described in a previous paper~\cite{BernaJBR2018}. A total of 98 samples were collected overall. The study was approved by the CSIRO Health and Medical Research, Human Research Ethics Committee (proposal number: LR 2/2017). Written informed consent was received from participants prior to inclusion in the study. Further, the use of data for this paper was also approved by the same committee (LR 12/2018 and 2019\_059\_LR). 

\emph{Field samples of human breath} were collected from 50 febrile patients at the Public Out Patient Department of Lihir Medical Centre in Papua New Guinea between 29th July and 19th August, 2017. Medical diagnosis confirmed 38 patients were malaria positive. Details of the study will be published in the near future. Approval for this study was provided by The Medical Research Advisory Committee (MRAC) of Papua New Guinea (MRAC No.\ 17.12) and CSIRO Health and Medical Research Ethics Committee (RR 14/2017) prior to patient recruitment. All participants included in the study were informed and provided consent before sample collection. Furthermore, all participants provided consent for the samples to be used in future studies. The use of data for this paper was also approved by CSIRO Health and Medical Research, Human Research Ethics Committee (proposal number: 2019\_059\_LR)


\subsection{Manual Grouping of Peaks}
\label{suppsec:manualGroups}

To generate data for training and validation each peak was manually assigned a group so that all the peaks assigned the same group should be aligned together. Groups to be aligned are numbered with an integer starting with 0. Peaks which are instead given `-1' as group number are largely ignored in the training process, only being used when paired with an identified peak as a negative pair.

\begin{figure}[h!]
	\centering
	\includegraphics[width=\figwidthfull]{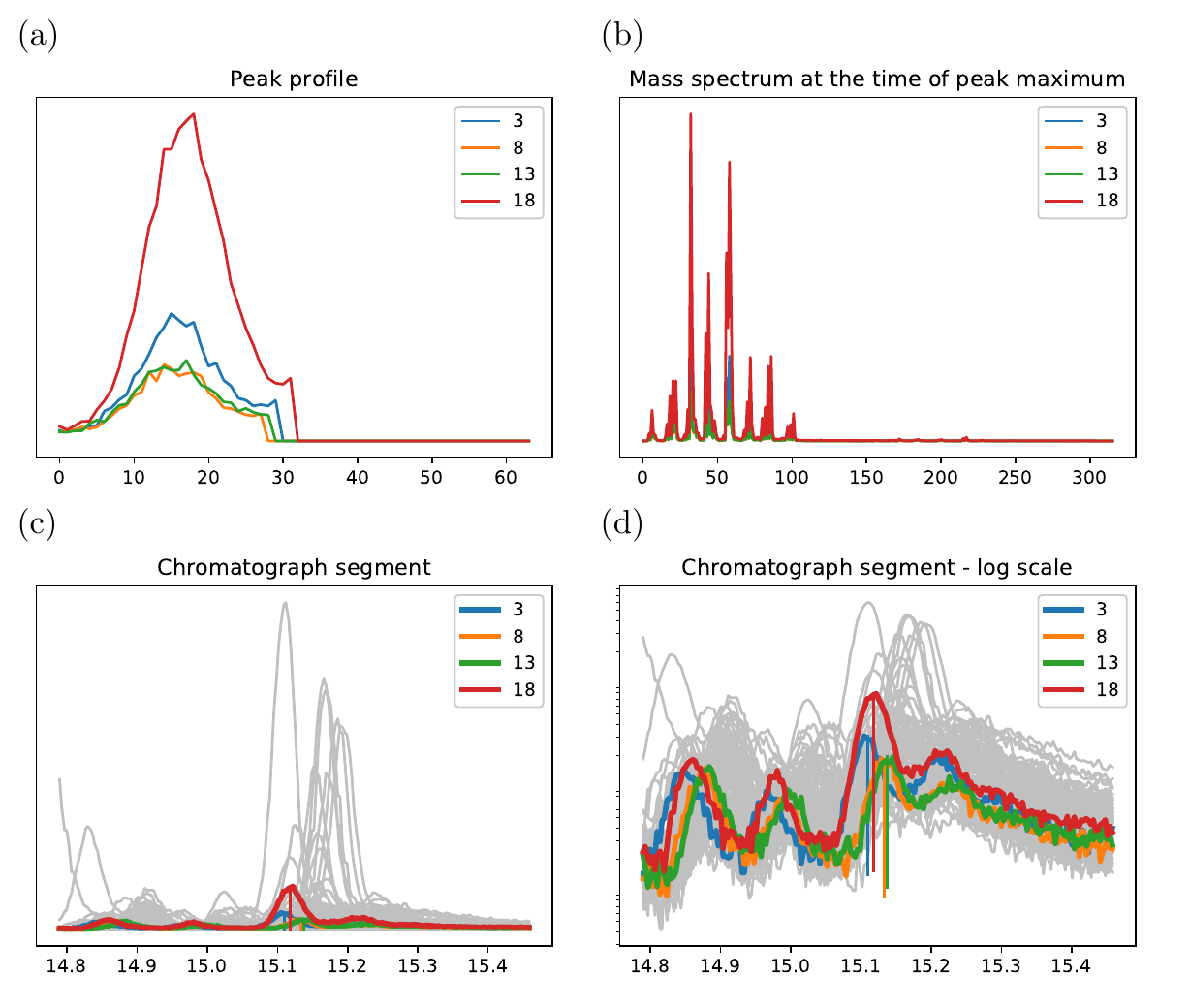}
	\caption{Information on a peak with m/z = 103 Da from four breath samples. Each plot shows a different piece of information that can be used together to decide if these peaks show the same compound. (a) The full peak as determined by the peak detection algorithm, x-axis shows time steps; (b) Mass spectrum at the time of the peak maximum, x-axis shows the m/z; (c) Chromatographic segment covering about half a minute to either side of the peak, x-axis shows the RT; and (d) Same chromatographic segment as (c) with intensity in log scale. The peaks under investigation have been indicated with vertical lines of the same colour at their RT. The grey traces in the background are the rest of the samples in the data set. }
	\label{fig:GCMS-Data-AssignmentProcess}
\end{figure}

Assignment of the group is done using the following procedure: 
\begin{enumerate}
	\item A custom excel template was used to highlight peaks which are close together in retention time to a particular selected peak -- that is within a specified cutoff window. This is used to find likely sets of peaks to compare. \label{itm:manualStepSelection}
	\item Figure~\ref{fig:GCMS-Data-AssignmentProcess} is plotted for several peaks in the highlighted group to allow visual inspection and comparison of the peak shape, mass spectra and chromatograph segment (the same components which are fed into the deep neural network). Any peaks deemed to be the same will be assigned the same group number. \label{itm:manualStepInspection}
	\item Step~\ref{itm:manualStepInspection} is repeated with several more highlighted peaks, including one or more which had already been assigned a group number. Peaks which match are assigned the same group number. This is repeated until all highlighted peaks have been visually inspected. \label{itm:manualStepRepeat}
	\item Steps~\ref{itm:manualStepSelection} to \ref{itm:manualStepRepeat} are repeated until all peaks have been checked. 
	\item Peaks which are not yet assigned a group are then compared to existing groups close by in retention time and with other unassigned peaks. Remaining lone peaks are either assigned a group number by itself or given a negative group number to remove them from the training set.
\end{enumerate}
The Excel template is available for download from the same URL as the data set.

\subsection{Detailed Training Process}
\label{supp:training}

The Adam optimiser was used at a learning rate of 0.001. Adam is an improvement on stochastic gradient descent that adapts the learning rate for each variable being optimised. Since it works well with default parameters on a wide range of deep learning models, alternative learning rates and other optimisers were not tested.

Our deep learning model has four separate outputs (Figure~\ref{fig:NetworkArchitucture}). Each sub-network's output 
(i.e.~the 10 neurons giving the difference between extracted features of the inputs) is used to conduct its own prediction.  For example, the output of the mass encoder sub-network is used to make a prediction of the alignment probability as if only information about the mass spectra of the two peaks were available. In addition, there is the main output of the overall network, which integrates all the information, and makes an overall prediction of alignment. Therefore, there are four predictions from the network, which are used during training. 

During the training process, the inputs to the network are pairs of peaks and the true labels indicating whether each pair should be aligned or not. The aim of training is to minimise the loss\footnote{Loss is a function of how much the predictions differs from the actual labels. In our case, since the network performs a binary classification an appropriate loss is the cross-entropy function.}, by using the gradient of the loss in backpropagation. There are four losses in the network (from the four predictions). Training minimises the sum of the four losses, 
with a weighting of 0.2 is given to each encoder's loss in the sum so that the emphasis remains on the main output.
The four predictions are compared with true labels to help users to better understand the performances of each part of the model. 
When using the network for alignment, only the main output from the network is considered.

\begin{figure}
	\centering
	\includegraphics[width = 0.85\textwidth]{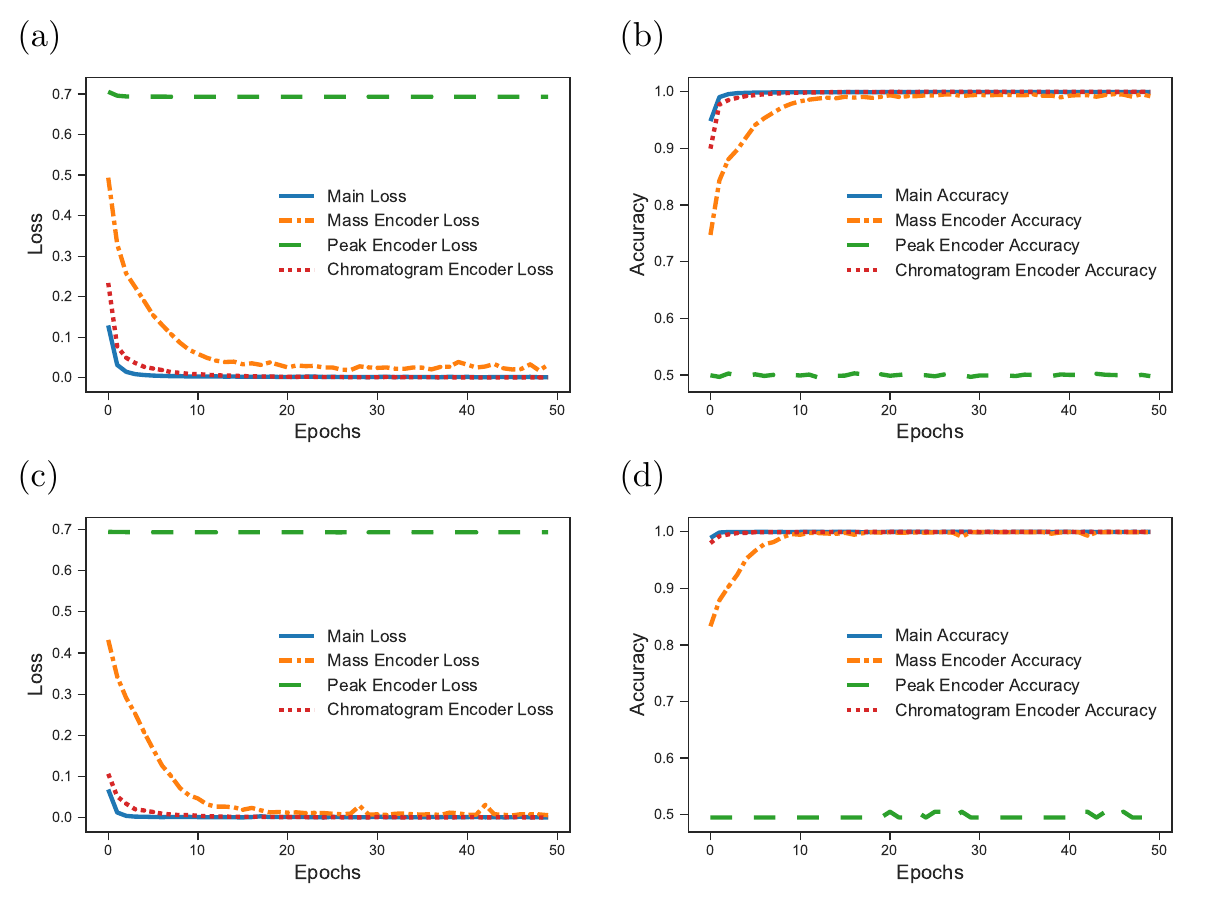}
	\caption{Loss and accuracy components of model G over 50 epochs. The components correspond to the main output from the network as well as the outputs from each Siamese sub-network. (a) Training loss components, (b) Training accuracy components, (c)Validation loss components, (d) Validation accuracy components.}
	\label{fig:StandardModelLossComponents}
\end{figure}

\begin{figure}[h!]
	\begin{center}
		\begin{subfigure}[b]{\figsemiwidth}
			\caption{} \vspace{-0.3cm}
			\includegraphics[width = \figwidthfull]{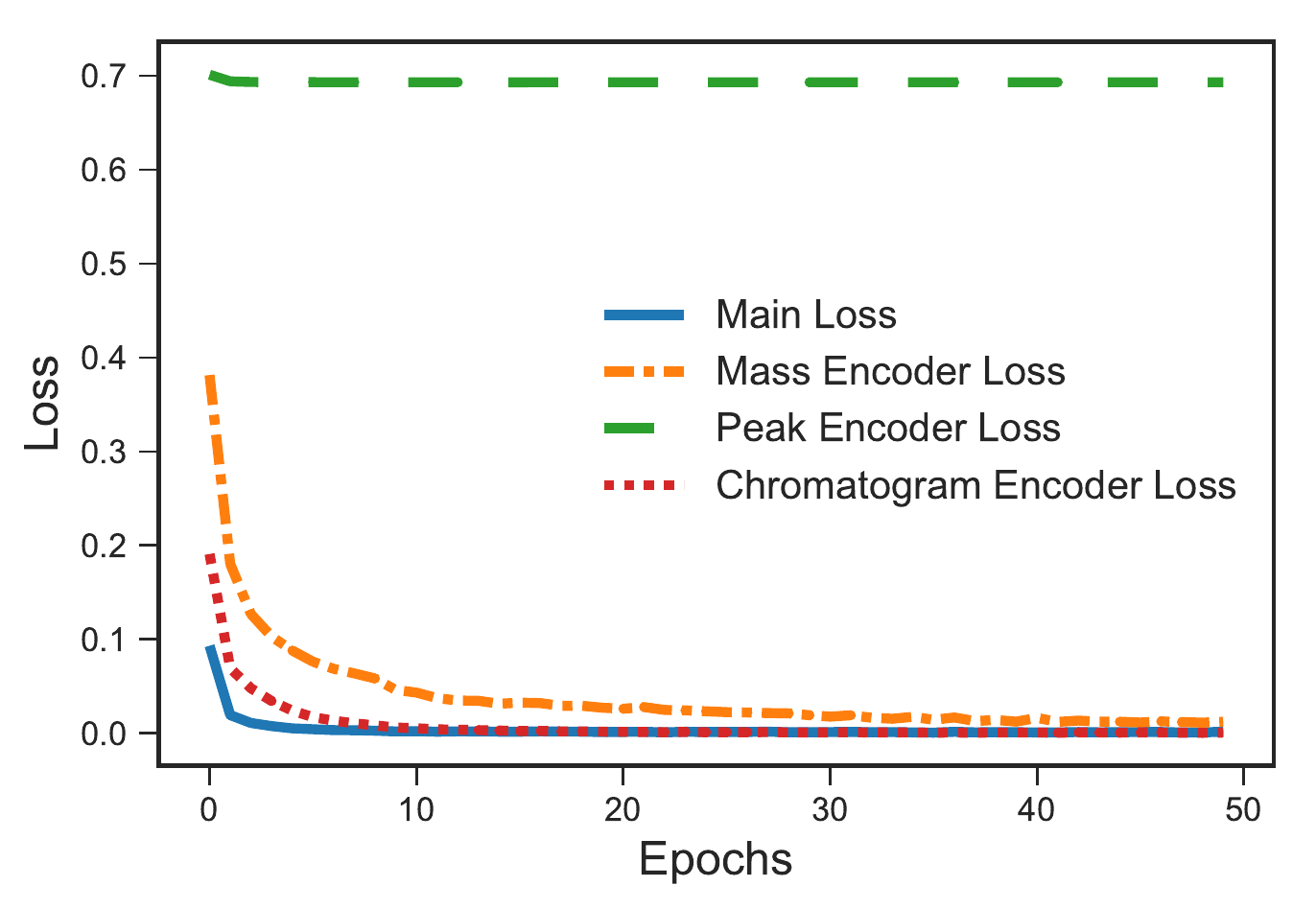}
		\end{subfigure} \hfill
		\begin{subfigure}[b]{\figsemiwidth}
			\caption{} \vspace{-0.3cm}
			\includegraphics[width = \figwidthfull]{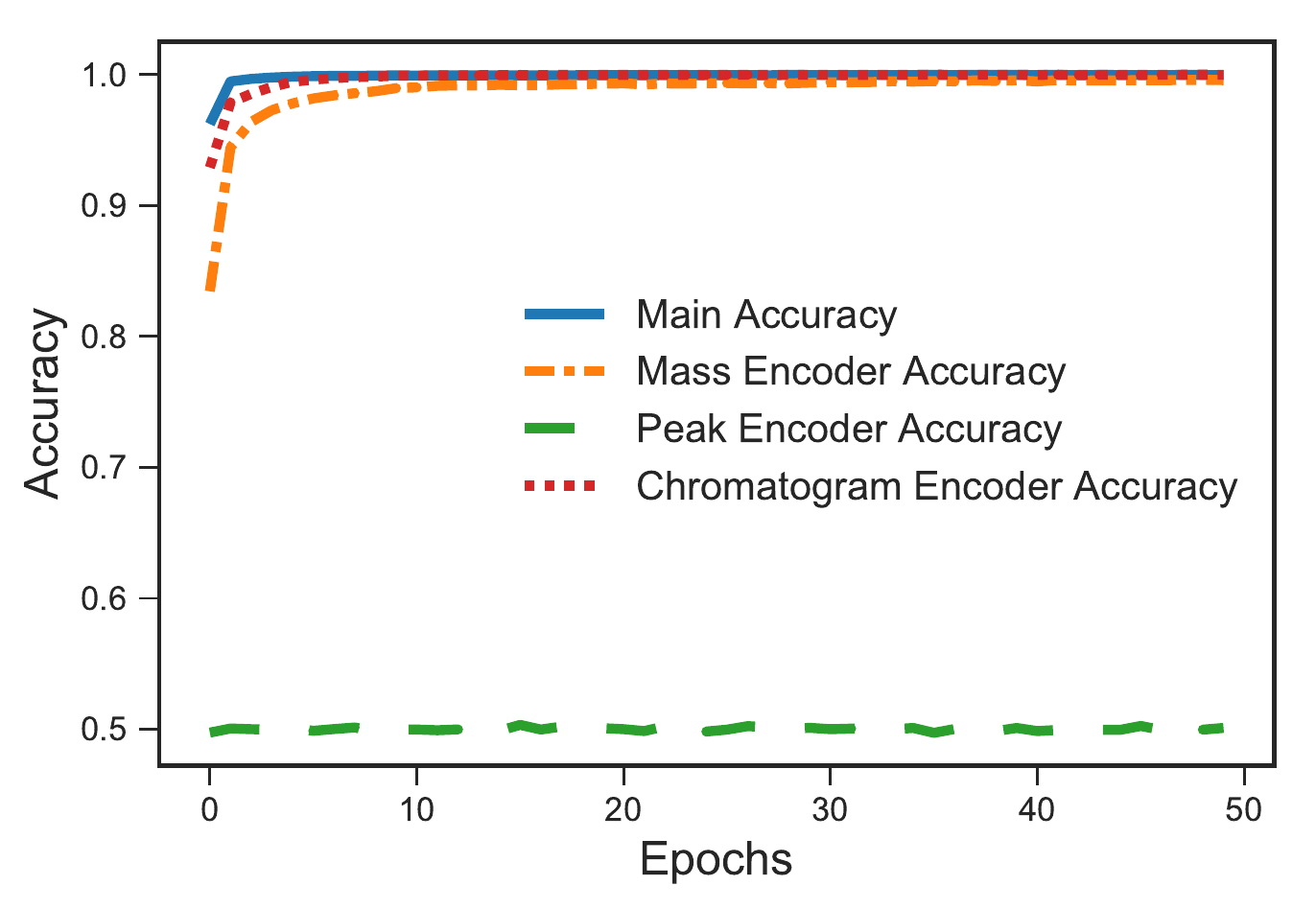}
		\end{subfigure}
		
		\begin{subfigure}[b]{\figsemiwidth}
			\caption{} \vspace{-0.3cm}
			\includegraphics[width = \figwidthfull]{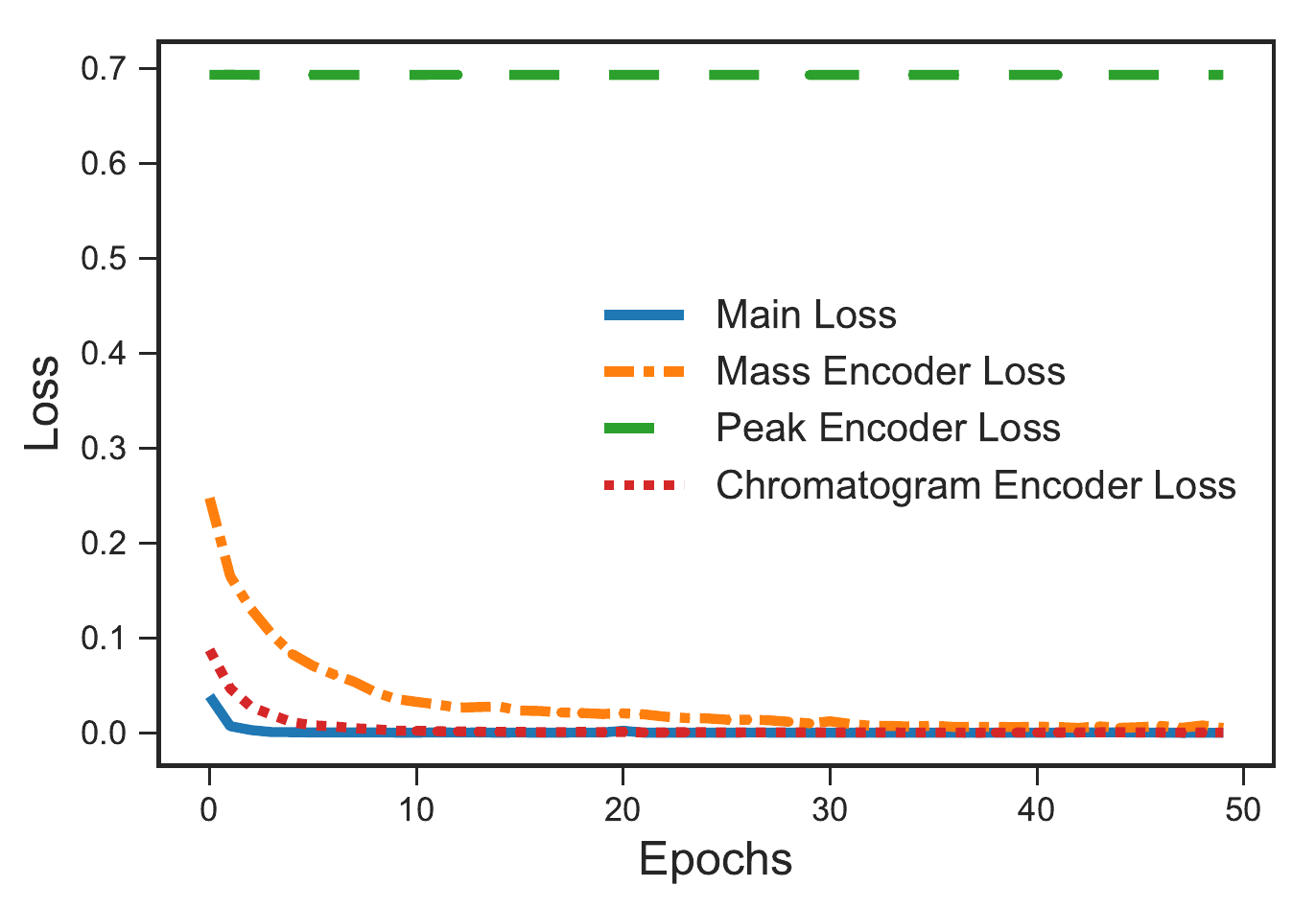}
		\end{subfigure} \hfill
		\begin{subfigure}[b]{\figsemiwidth}
			\caption{} \vspace{-0.3cm}
			\includegraphics[width = \figwidthfull]{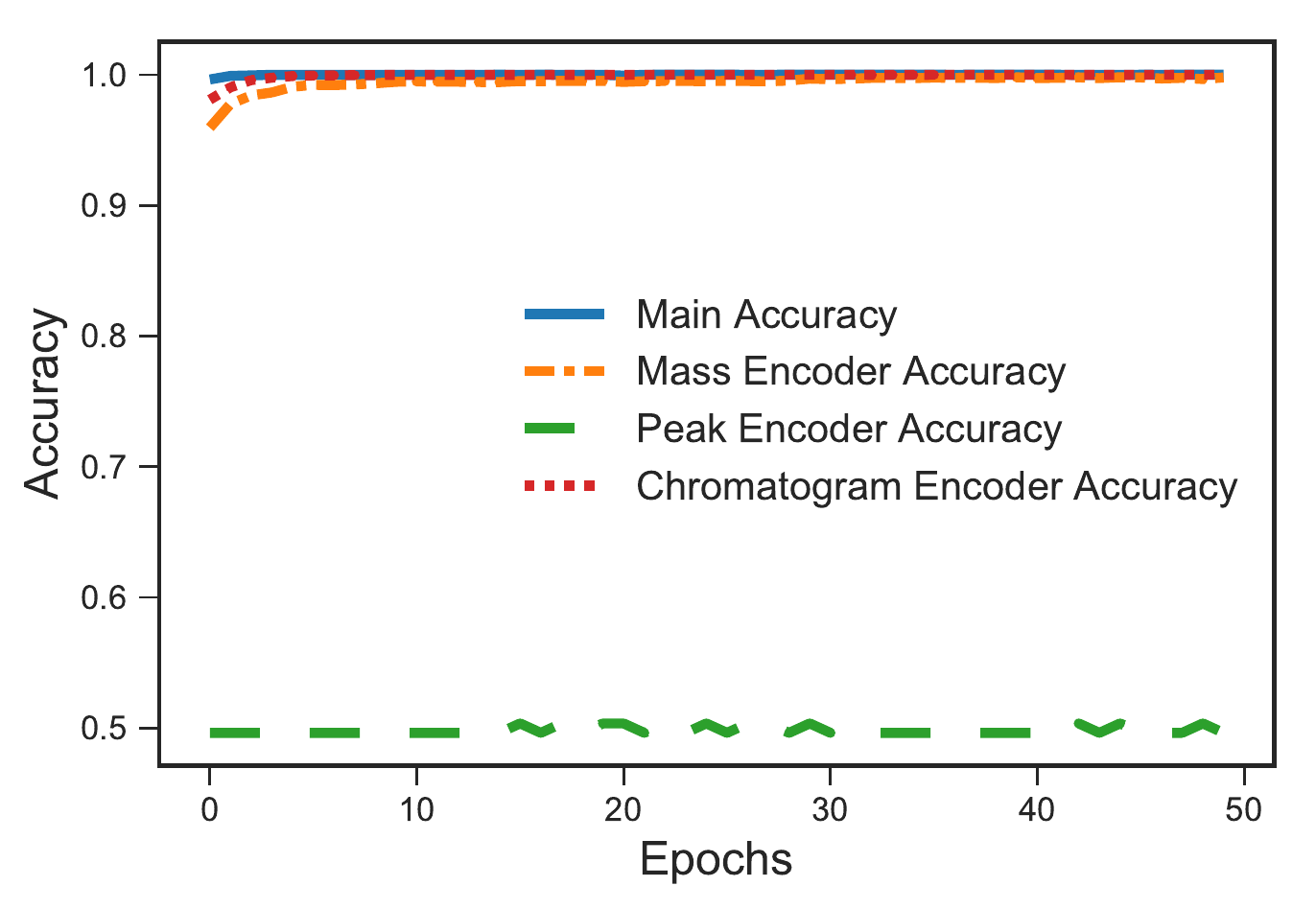}
		\end{subfigure}
	\end{center}    
	\caption{Loss and accuracy components of model E over 50 epochs. The components correspond to the main output from the network as well as the outputs from each encoder. (a) Training loss components, (b) Training accuracy components, (c) Validation loss components, (d) Validation accuracy components.}
	\label{fig:StandardModelLossComponents-E01}
\end{figure}

We can thus track the loss and accuracy of the network over the training epochs for each of the four outputs (Figure~\ref{fig:StandardModelLossComponents}). Most of the models exhibit similar patterns of behaviour for loss and accuracy over cumulative epochs (see Supplementary Figures~\ref{fig:StandardModelLossComponents-B01} and~\ref{fig:StandardModelLossComponents-E01} for examples of this plot for other models). The plots displayed in Figure~\ref{fig:StandardModelLossComponents} show that the loss from the peak encoder component stabilises shortly after training begins and its accuracy ceases to rise. There are two likely explanations for this observation. Either there is little information in the peak profiles beyond what is already provided by the chromatogram segments or the peak encoder requires more epochs to train to an acceptable level of accuracy compared to the other components. If the peak encoder's contribution simply added noise to the final output it would be cut off early in the training process.

Of the other two encoder components, the mass encoder took more epochs to train than the chromatogram encoder. This is a little surprising, as the identification of compounds by domain experts relies heavily on the information in the mass spectrum. This could be due to the different network architecture of the two encoders. The CNN architecture of the chromatogram encoder is better at finding features in the data for alignment comparison than the fully connected network of the mass encoder. 

\subsection{Variations to the Model}
\label{suppsec:modelVariations}

From the results in Table~\ref{table:TrueFalsePositives}, the main difficulty is the high false positive rate. It can also be seen from Figure~\ref{fig:StandardModelLossComponents} that it is difficult to train the peak encoder. To see if changes to the model architecture could alleviate these two problems, a number of modifications to model D were tested. Making changes to model D allowed greater reduction in false positive rate than models F and G, as the model wasn't trained with Breath73 and Breath88 data sets, which present the high FP rates in the results. 
The variations to the main architecture follow three general themes:
\begin{enumerate}
	\item Modifications to the peak encoder. For example removing the encoder altogether or making simplifications (where we replaced the LSTM units with Gated Recurrent Unit~\cite{ChungNIPS2014}) with the aim to make it easier to train.
	\item Changes to the encoders. Adding dropout and changing the number of encoding neurons. The aim is to reduce the rate of false positives by reducing overfitting.
	\item Changing the convolution layers in the chromatograph segment encoder, for example changing the number of layers or adding dropout between the layers. The aim is to improve the feature extraction from the chromatograph.
\end{enumerate}
From the three general themes described above, we proposed 18 variations to the model (Table~\ref{table:VariationModels}), each of which was trained and tested with ten repetitions. The networks were trained on the GPU and took under 20min for variant 2, under 3hr for variant 3, and under 6hr for the rest. 
We tested the resultant models on the Breath103, Breath73 and Breath88 data sets.

\begin{figure}
	\centering
	\includegraphics[width=\figwidth]{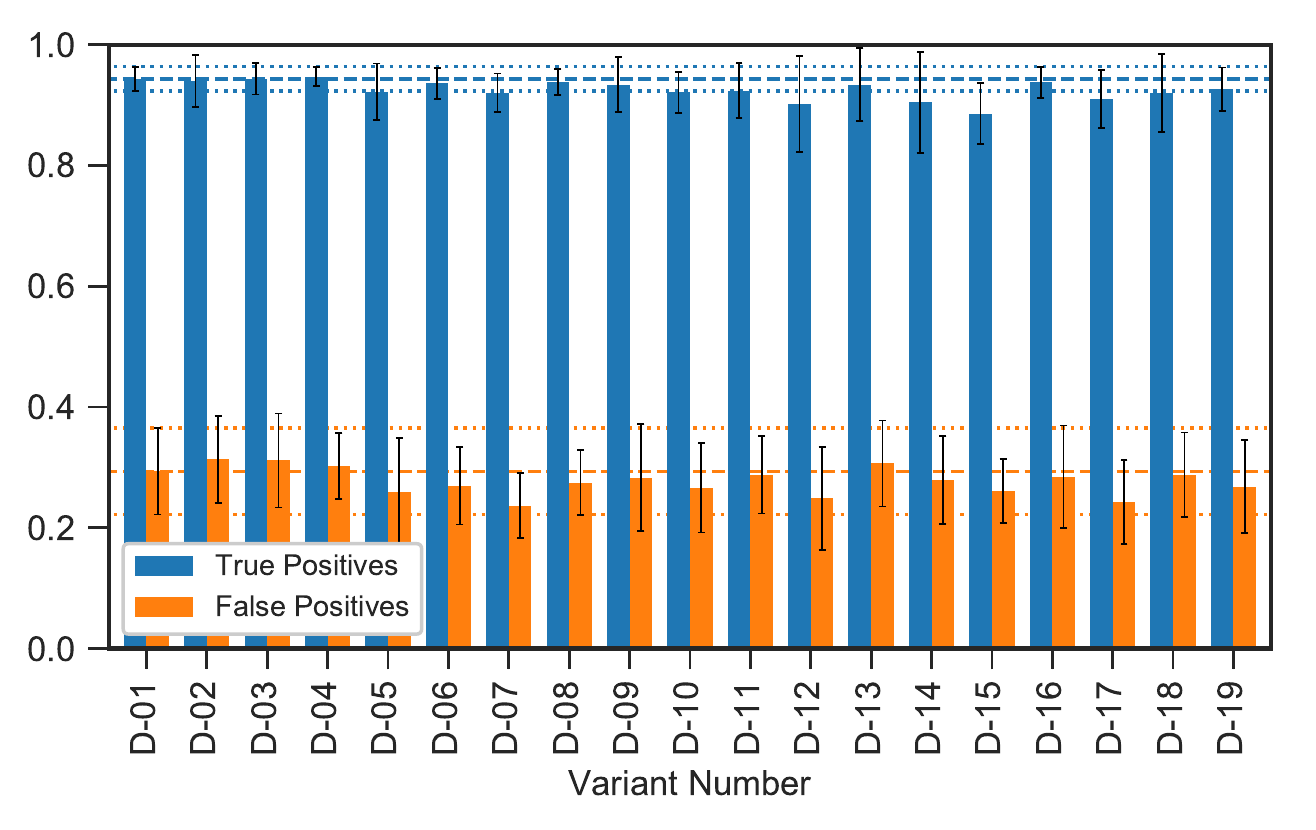}
	\caption{Performance of model alternatives on Breath73. The average and the standard deviation of 10 repetitions are shown. Model D-01 is the benchmark against which all changes were made.}
	\label{fig:ModelAlternatives-b73}
\end{figure}

\begin{figure}
	\centering
	\includegraphics[width=\figwidth]{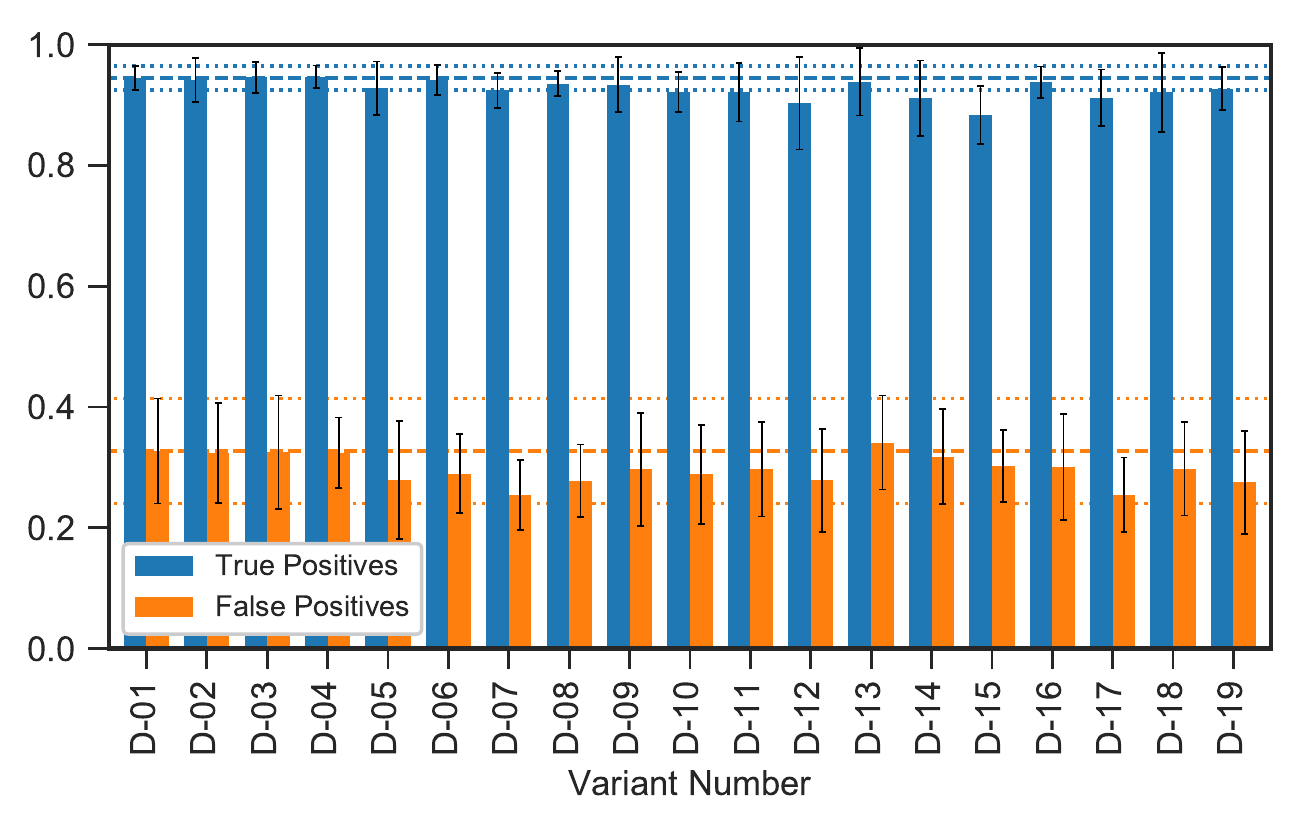}
	\caption{Performance of model alternatives on Breath88. The average and the standard deviation of 10 repetitions are shown. Model D-01 is the benchmark that all changes were made against}
	\label{fig:ModelAlternatives-b88}
\end{figure}

The average alignment results are show in Table~\ref{table:VariationModels}. The results for Breath73 (with standard deviations) are show in Figure~\ref{fig:ModelAlternatives-b73}, Breath88 showed a similar outcome (Figure~\ref{fig:ModelAlternatives-b88}). Although there is no significant improvement in true positive rate, the false positive rates show slight reductions in some variants, most notably in D-07, D-12 and D-17. However, in all of these cases, there is a slight drop in TP rate, demonstrating that the networks' ability to distinguish between positive and negative pairs is closely related. These models may be simply making some trade off between the TP and FP rate instead of having an overall improvement.

\input{table_resultsDVariants_part1}

\begin{figure}
	\begin{center}
		\begin{subfigure}[b]{\figsemiwidth}
			\centering
			\caption{} \vspace{-0.3cm}
			\includegraphics[width = \figwidthfull]{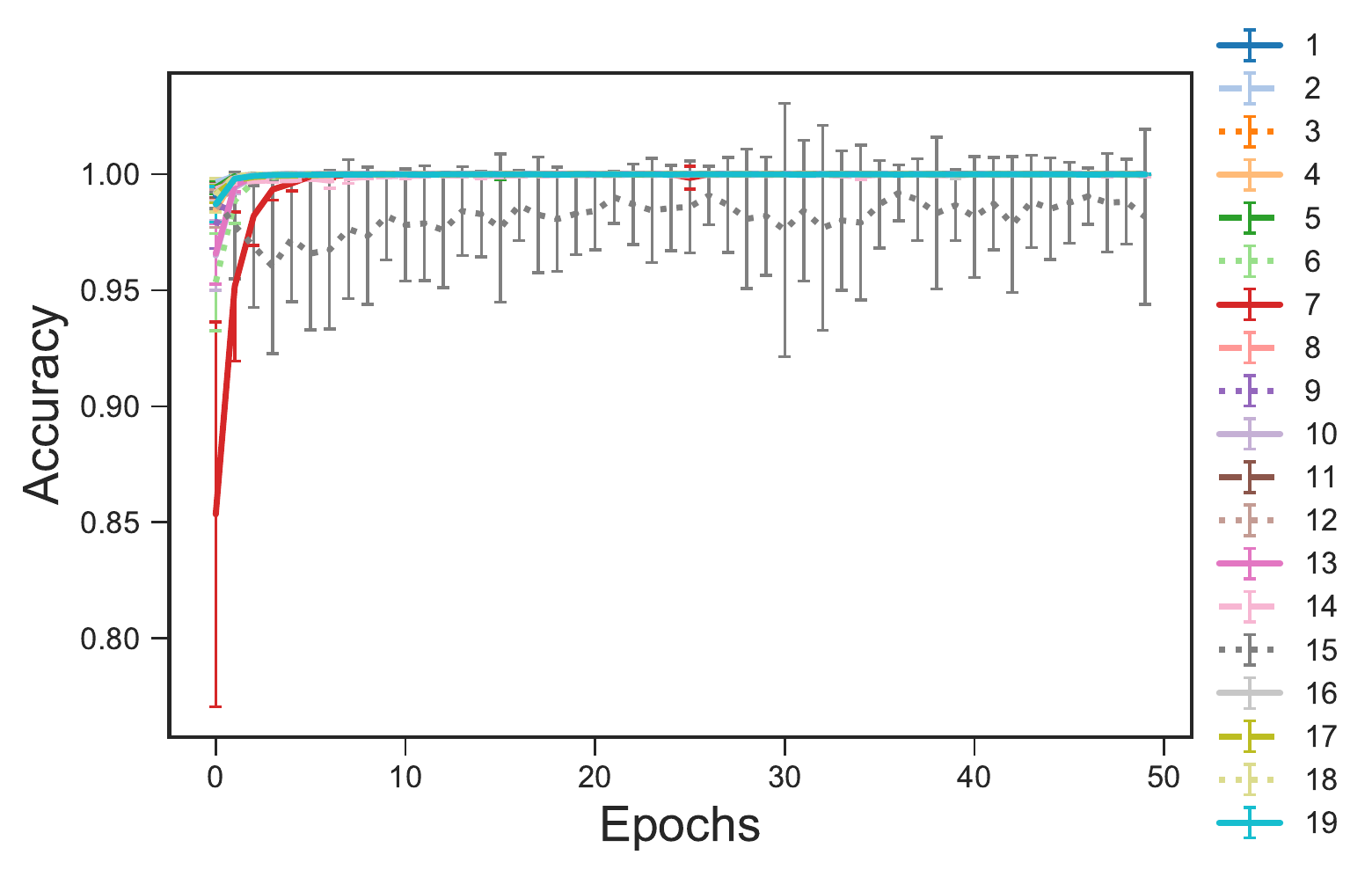}
		\end{subfigure} \hfill
		\begin{subfigure}[b]{\figsemiwidth}
			\centering
			\caption{} \vspace{-0.3cm}
			\includegraphics[width = \figwidthfull]{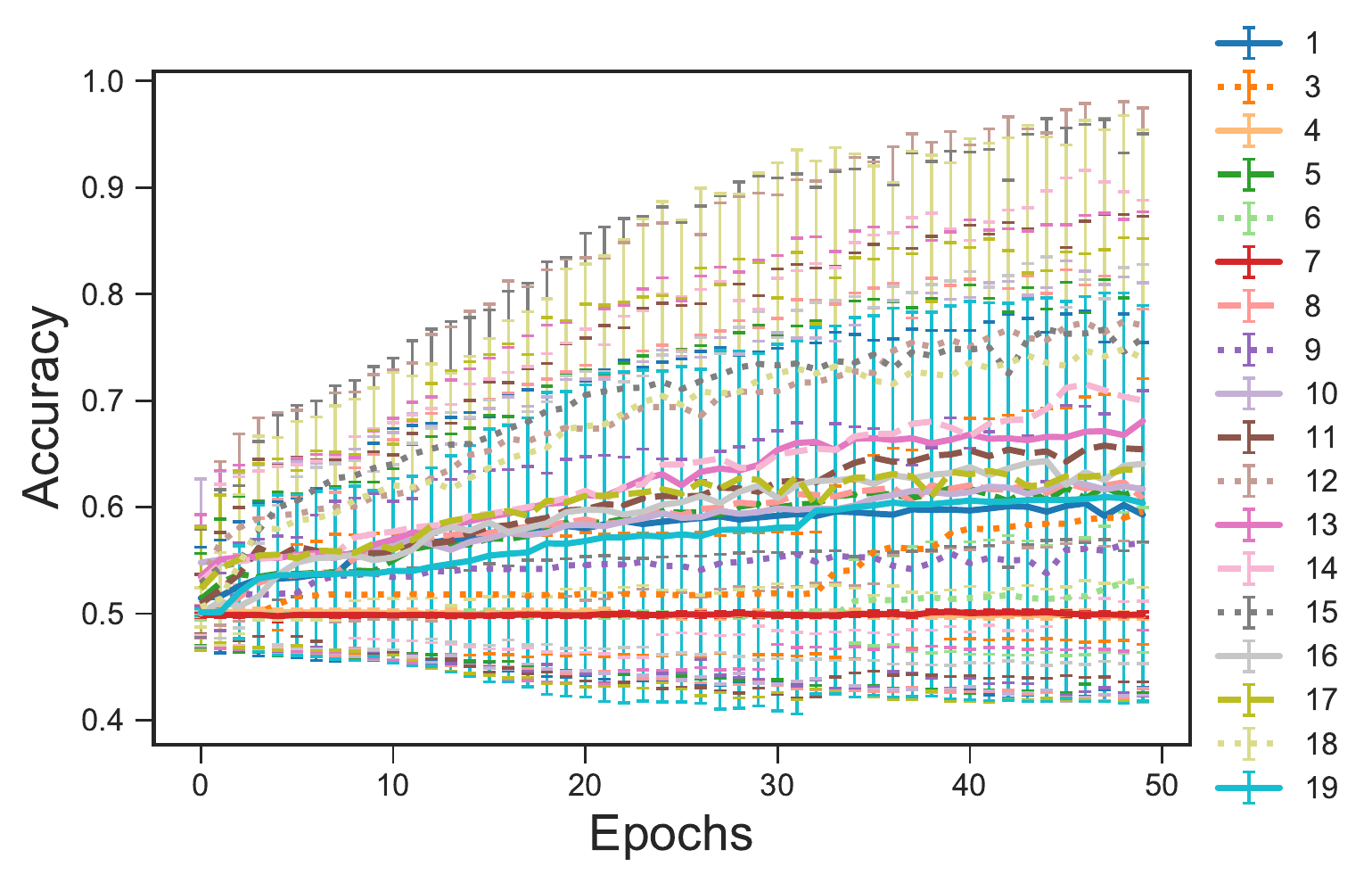}
		\end{subfigure}
		
		\begin{subfigure}[b]{\figsemiwidth}
			\centering
			\caption{} \vspace{-0.3cm}
			\includegraphics[width = \figwidthfull]{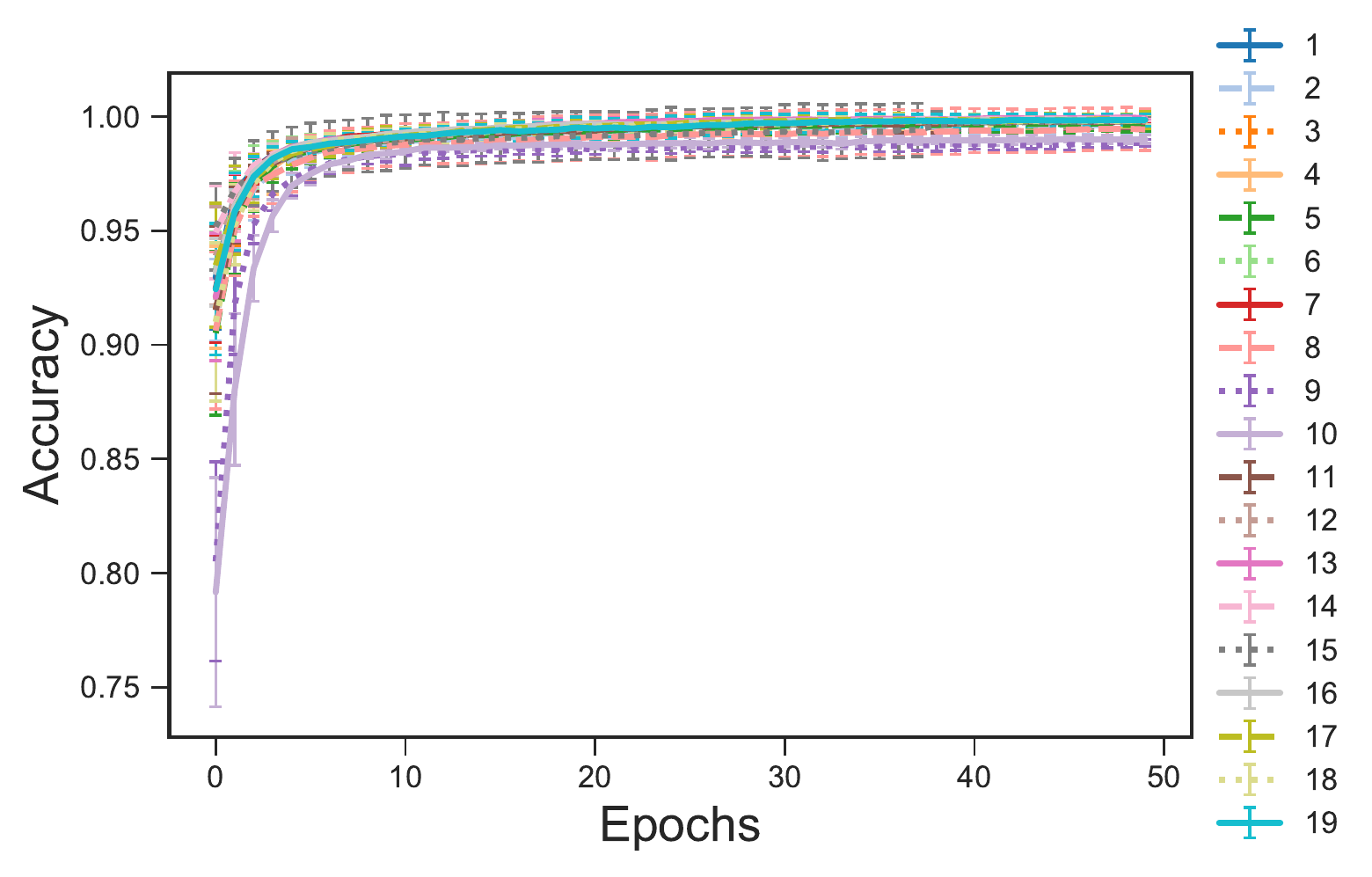}
		\end{subfigure} \hfill
		\begin{subfigure}[b]{\figsemiwidth}
			\centering
			\caption{} \vspace{-0.3cm}
			\includegraphics[width = \figwidthfull]{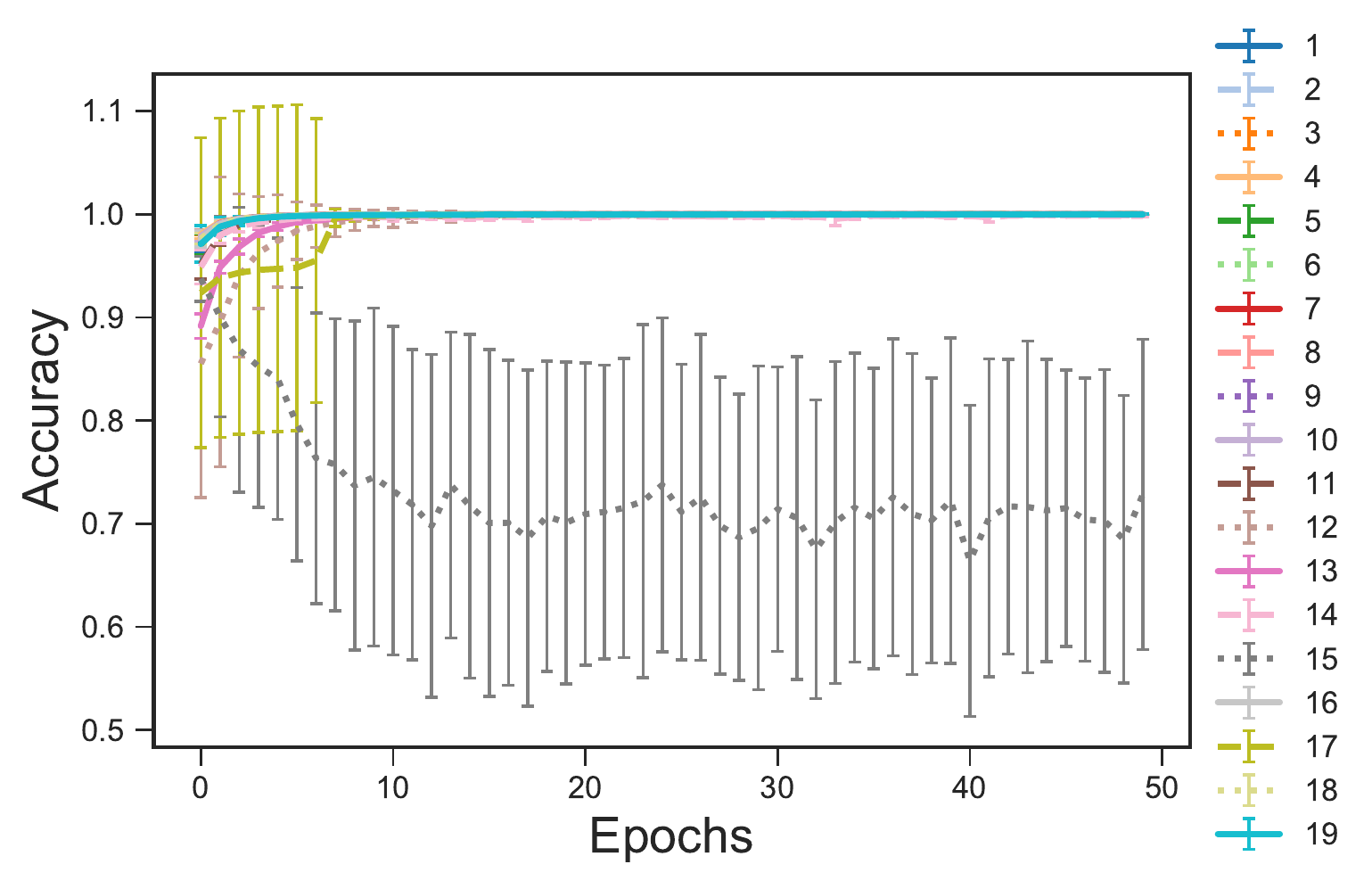}
		\end{subfigure}
	\end{center}   
	\caption{Validation accuracy for model D's variants 1-19 plotted over the training epochs: (a) Main Accuracy, (b) Peak Encoder Accuracy, (c) Mass Encoder Accuracy, and (d) Chromatogram Encoder Accuracy.  Note that variant 2 does not have a peak encoder, thus its accuracy is not plotted in (b).}
	\label{fig:ValAccComponentsOfExpModels}
\end{figure}

We again analysed  the training of the new network variations over 50 epochs (Figure~\ref{fig:ValAccComponentsOfExpModels}). Looking at the validation accuracy of the four outputs, the peak encoder has the most inconsistent results. While the peak encoder of some variants was more accurate than others, the large variation in accuracy means the peak encoder can not be trained effectively and consistently. These results indicate the peak encoder may be unnecessary for the deep learning model of chromatogram alignment. Furthermore, the substantial increase in training time for the peak encoder compared to the other encoders suggests it would be advantageous to remove this component, especially when the size of the training data becomes much larger than the current set. 

The other notable result in the validation accuracy was variant 15's chromatogram encoder, which had a much worse performance than any other variant. In variant 15, the dropout rate between the convolutional layers was set to a high value of 0.5 (compare to no dropout in the original). The poor performance of its chromatogram encoder and its overall decreased TP rate, demonstrate the necessity of complex connectivity in the convolution layers to fully describe the chromatogram segment. 

Given the performance of the peak encoder in all variants, we devised a further 12 second generation variants of model D based on a combination of the first generation variants 2 or 3 with one of six other variants (Table~\ref{table:VariationModels2}). The six variants were chosen for their performances (mainly low FP rates). The training times for variants based on variant 2 were under 15 min, and the training times for variants based on variant 3 were between 2.5 to 3 hours. In comparison, the training time for variant 1 (the original model D) was around 5.5 hours. 

\input{table_resultsDVariants_part2}

We observe from the performance of these 12 variants, that those based on variant 9 (D-21 and D-27) have the best TP rates, which were also an improvement on the original model (D-01). Conversely, those based on variant 17 (D-24 and D-30) have the lowest FP rate, but also some of the lowest TP rates. Moreover, we note that variants based on variant 2 (D-20 to D-25) have overall better performance (higher TP rates and lower FP rates) than those based on variant 3 (D-26 to D-31), despite the former lacking the peak encoder in the network. The performance of these variants combined with their much faster training (and consequently prediction) time confirms it is preferable to remove the peak encoder from the overall network.

\subsection{Metrics for Overall Alignment}
\label{suppsec:metricGrouping}

To compare the results from various different methods, we need to modify the assessment metric. This is because in our assessment of pairwise comparison, we compare all possible pairs of peaks (in a cut off segment) in the data set, which no other methods have done. A fairer comparison between the methods would be how well each peak has aligned to the other peaks in the final alignment. 

For each group of peaks in the ground truth alignment, we identify the range of RT for these peaks in the aligned result. From this range, we find the RT with the most number of peaks from the real group, this will be used as the actual aligned RT. Count the number of peaks from the real group that is aligned to this RT, this is added to the number of true positive peaks. All other peaks that are aligned to this RT will be counted towards the number of false positive peaks. 

The total number of actual positive peaks is the total number of identified peaks in the data sets, thus we can calculate the TP rate. For the false positive values, instead of calculating the FP rate, we will be using the False Discovery Rate (FDR), which is the proportion of predicted positive cases that are false positives. The FDR is a much more meaningful value here as we do not have a good estimate of the total number of negative cases.

\subsection{Future Work}
\label{supp:future}

We built a network that compares the retention time, chromatogram segments, peak profile and mass spectra at peak maximum between two peaks. This allows the network to make use of all the information about a peak in the comparison. We also designed 18 alternative network structures and 12 combinations of the alternatives to test whether alterations to individual component of the network can improve its performance. The main issue we observed from these results is that none of the model alternatives was able to consistently train the peak encoder. In fact, when we removed the peak encoder from the network and altered other parts of the network, we achieved better true positive \emph{and} false positive rate, as well as much faster training and prediction times. 

Investigating the performance over epochs of the peak encoder, we notice it seems to train much more slowly than the other network components and stop improving after a certain point, suggesting that its contribution to the prediction was switched off. It may be effective to pretrain the peak encoder as an autoencoder so that it can make useful contributions to ChromAlignNet right from the first training epoch. Training as an autoencoder consists of combining an encoder network with a decoder network and assessing the ability to reconstruct the original peak from the encoding. This is an unsupervised training process and it is much easier to extract a large amount of unlabelled peak data (i.e.\ peaks that have not been manually identified) to pretrain the autoencoder compared to labelling the peaks into groups. 

Furthermore, recent development in deep learning has shown that CNNs can be more effective than RNNs in many sequence based tasks traditionally seen as the domain of recurrent structures~\cite{Bai2018}.
Even when sticking to the basic versions of each architecture, the performance of CNNs versus RNNs on sequential data is highly task dependent~\cite{Yin2017}. In natural language processing, CNNs outperforms RNNs where detection of specific features or key words is important, such as in question and answer systems. Conversely, comprehension of long-range semantics is better done with an RNN~\cite{Yin2017}.
In addition, there has been explorations of hybrid CNN-RNN structures which aims to combine the benefits of both~\cite{Gross2017,Lin2017}.
In future work, we will test the use of CNN and other architectures for the peak encoder. 

We will also investigate improvement to the mass spectrum encoder. Currently this encoder takes as input all mass values from the GC-MS instrument, however, close inspection of the mass spectra shows that not all masses have useful information about a peak. This is also evident in the performance of variants 09, 10, 21 and 27, which all have very high drop outs (0.5) in the mass encoder. In future work, we will investigate the use of smaller mass spectral segments as inputs to the mass encoder. We will also investigate other network structures that can extract better features from the mass spectra. 

The overall false positive rate of prediction result is relatively high and that will impede the use of ChromAlignNet. One approach which may reduce FP rate is to replace some parts of the loss function from the current standard cross entropy with the triplet loss which is commonly used for Siamese architecture networks. The triplet loss uses three encodings simultaneously instead of two, comparing both a positive and negative example. Often these are hand chosen to be difficult cases. The triplet loss measures the difference between the encodings of the positive pair and the negative pair and actively works to increase the gap between these cases. The Siamese components would be pre-trained with the tripled loss to produce a good encoding. The full network would then be assembled and trained as a whole to fine tune each component. Since the optimisation of neural networks is a non-convex problem this would help the network avoid getting stuck in a local optima that does not adequately distinguish true negatives from false positives.

The prediction from ChromAlignNet is a value between 0 and 1, i.e.\ the likelihood of two peaks being the same. During alignment, we allocate peaks to the same group when the pairwise prediction value is greater than and equal to 0.5. This might be a threshold that is too low in practice, as the current FP rate suggests many peaks that don't belong to the same group will be merged together. Currently, we employ various rules in the grouping stage to separate these peaks, in future work, we will aim to make this process more robust. 

Moreover, using the hierarchical clustering algorithm to assign groups may be a simple solution but is not ideal.
If incorrect fragmentation or combination of groups occur due to prediction errors being made on certain pairs of peaks
this will impact the average distance between clusters. Since this determines which further clusters are combined, such errors can have flow-on effects. 
A more accurate method of assigning groups may be to simultaneously modify the pairwise probability based on the other members in the proposed group. A group should have a high pairwise probability between all members and a peak which shows a high probability of alignment with a single peak in the group but low probability of alignment with the other peaks should not be aligned with the group. This may be developed as a custom clustering algorithm or performed as a check after initial clusters are obtained and further iterations of clustering can be performed.
We will explore this in future work. 

\clearpage

\section{Supplementary Figures}


\begin{figure}[h!]
	\begin{center}
		\begin{subfigure}[b]{\figsemiwidth}
			\caption{} \vspace{-0.3cm}
			\includegraphics[width = \figwidthfull]{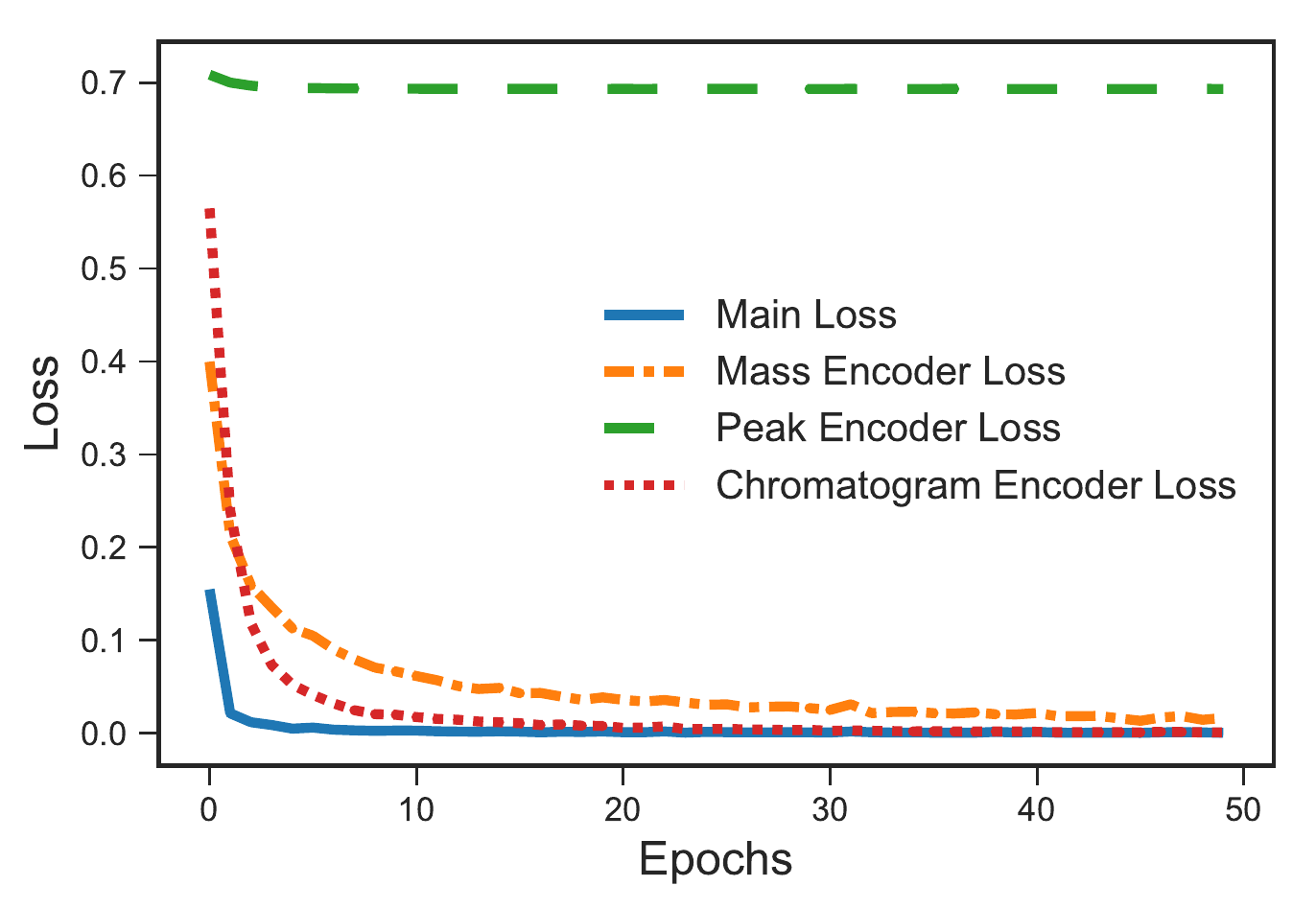}
		\end{subfigure} \hfill
		\begin{subfigure}[b]{\figsemiwidth}
			\caption{} \vspace{-0.3cm}
			\includegraphics[width = \figwidthfull]{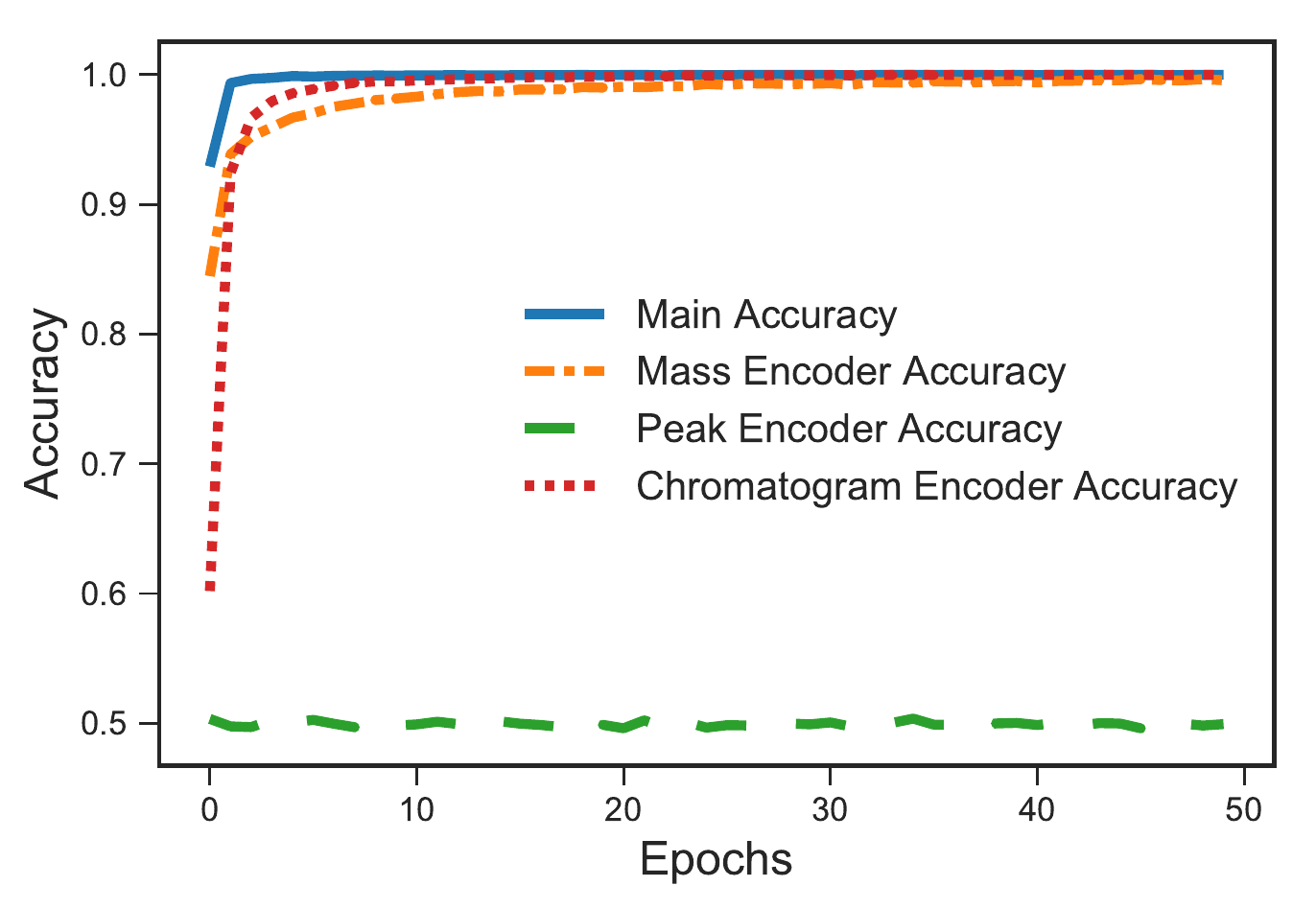}
		\end{subfigure}
		
		\begin{subfigure}[b]{\figsemiwidth}
			\caption{} \vspace{-0.3cm}
			\includegraphics[width = \figwidthfull]{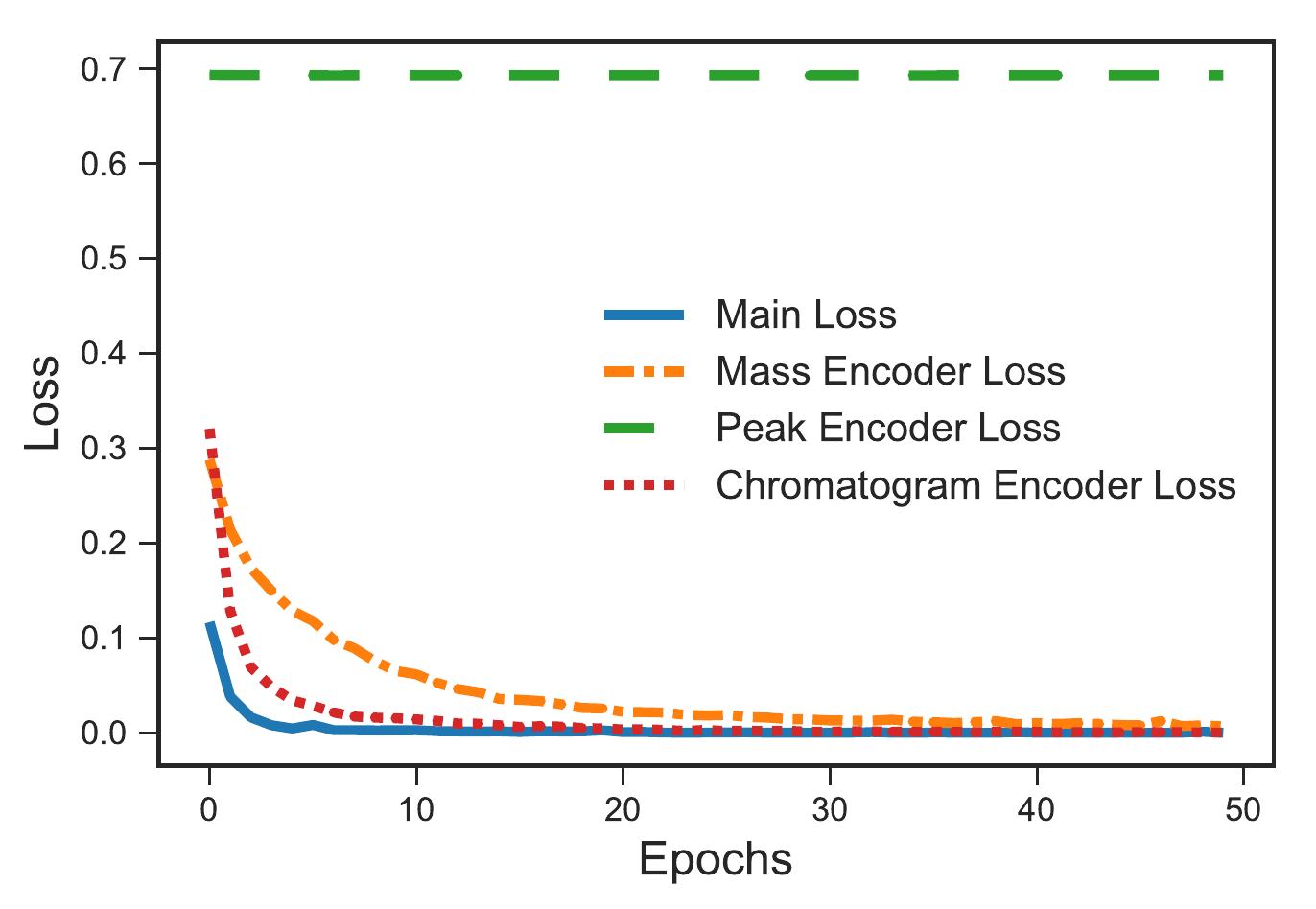}
		\end{subfigure} \hfill
		\begin{subfigure}[b]{\figsemiwidth}
			\caption{} \vspace{-0.3cm}
			\includegraphics[width = \figwidthfull]{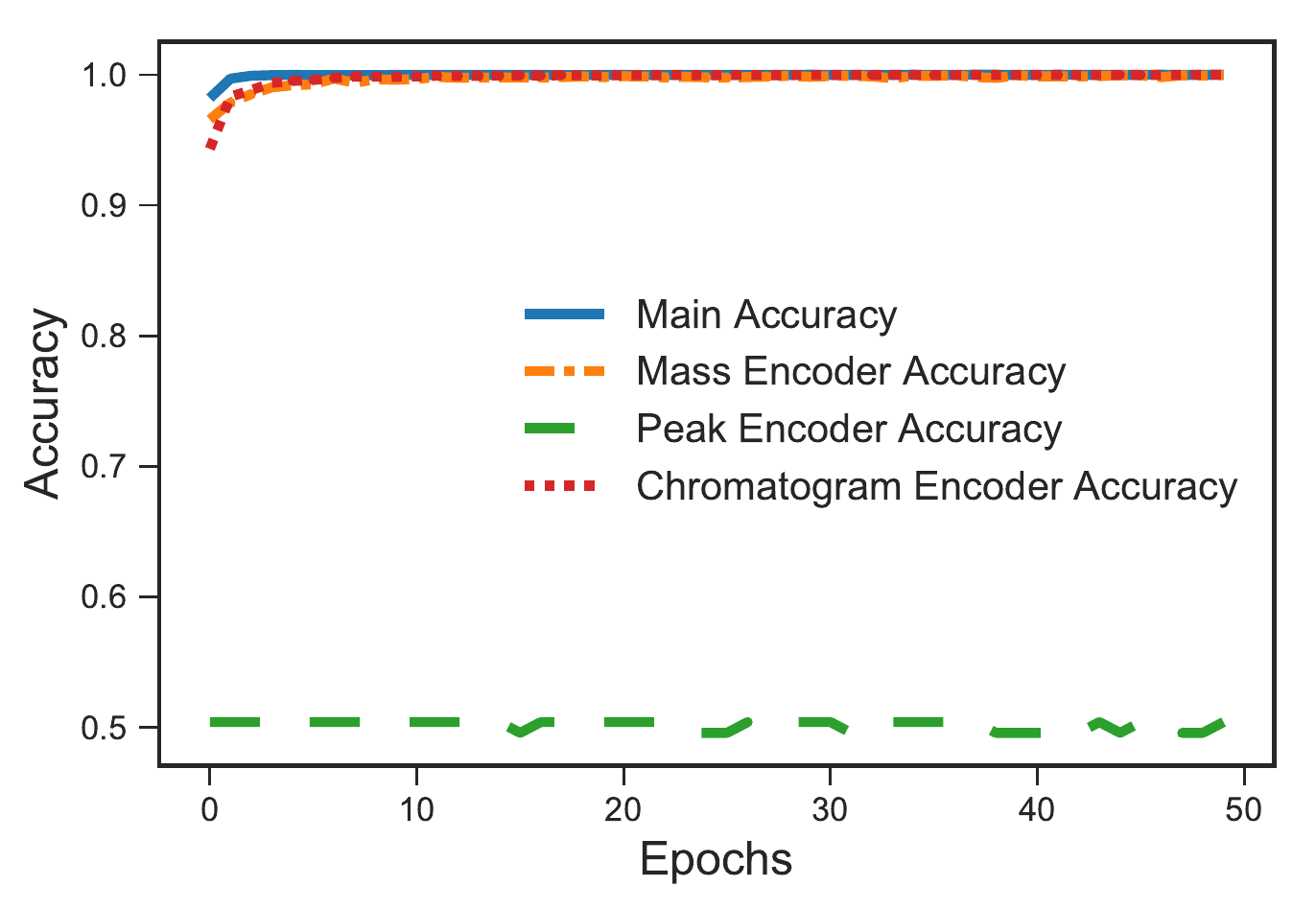}
		\end{subfigure}
	\end{center}    
	\caption{Loss and accuracy components of model B over 50 epochs. The components correspond to the main output from the network as well as the outputs from each encoder. (a) Training loss components, (b) Training accuracy components, (c) Validation loss components, (d) Validation accuracy components.}
	\label{fig:StandardModelLossComponents-B01}
\end{figure}

\begin{figure}
	\begin{center}
		\begin{subfigure}[t]{\figwidthfull}
			\centering
			\caption{} \vspace{-0.3cm}
			\includegraphics[width=\figwidthfull]{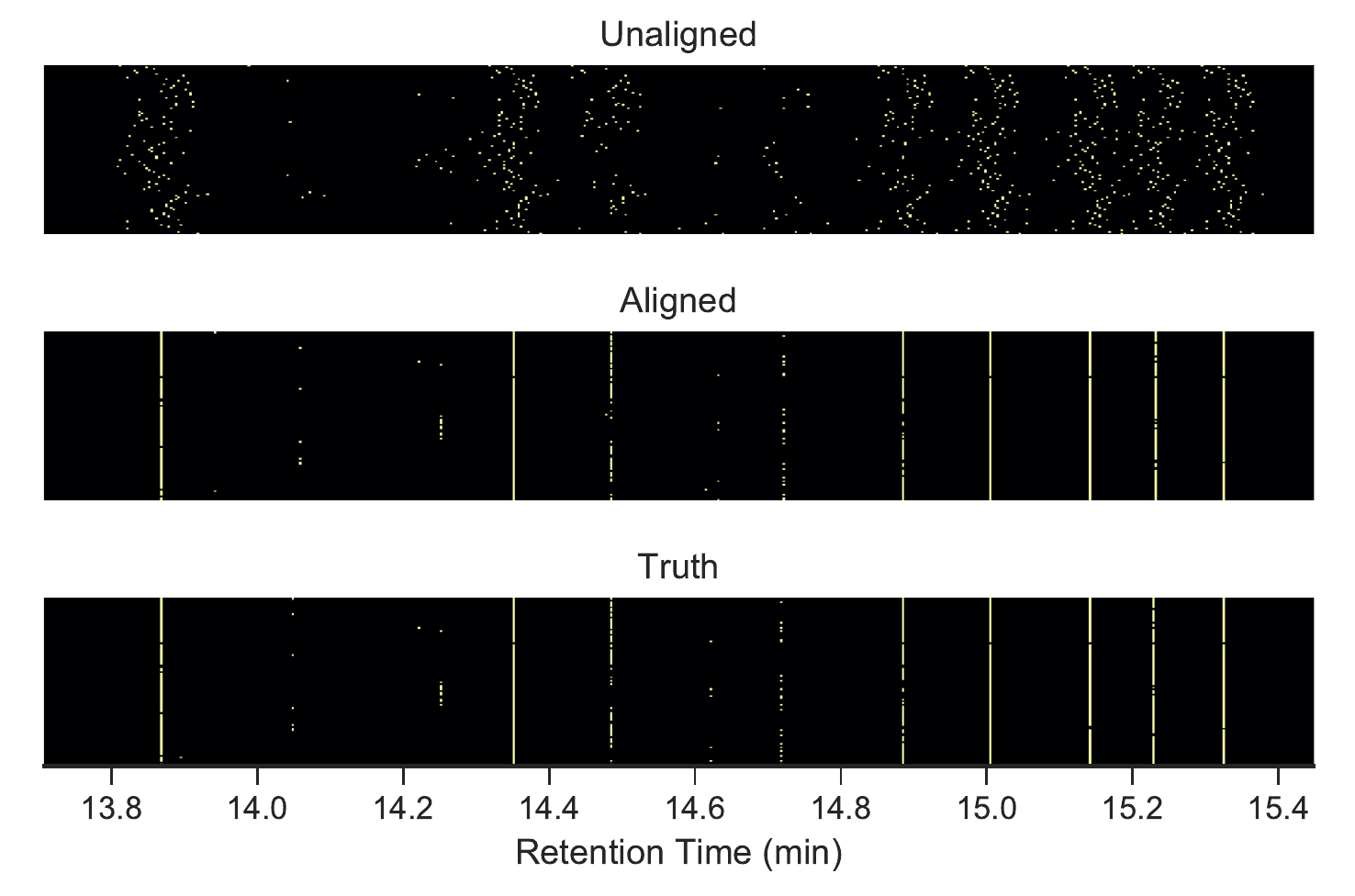}
		\end{subfigure}
		
		\begin{subfigure}[t]{\figwidthfull}
			\centering
			\caption{} \vspace{-0.3cm}
			\includegraphics[width=\figwidthfull]{alignment_plot_breath115_C01}
		\end{subfigure}
	\end{center}
	\caption{Alignment outcome for the data set Breath115 using the best performing network from model C. For this network, the true positive rate was 0.992 and the false positive rate was 0.004. (a)~Chromatographic image: each horizontal position is a single sample and the points are plotted at the peak retention times. 
		(b)~Chromatographic plot: each line is a single sample.  
	} 
	\label{fig:AlignBreath115_C01}
\end{figure}

\begin{figure}
	\begin{center}
		\begin{subfigure}[b]{\figwidthfull}
			\centering
			\caption{} \vspace{-0.3cm}
			\includegraphics[width=\figwidthfull]{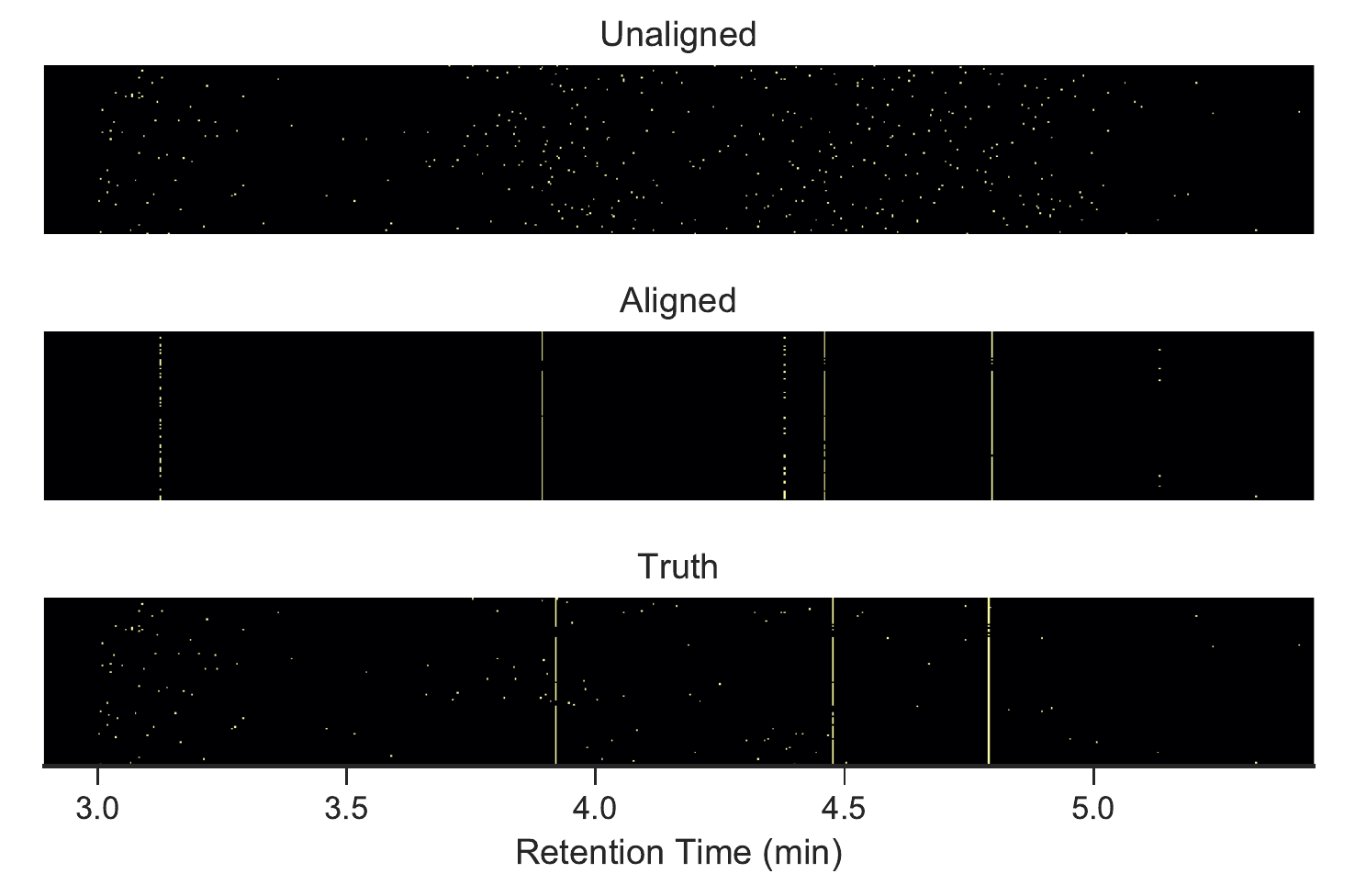}
		\end{subfigure}
		
		\begin{subfigure}[b]{\figwidthfull}
			\centering
			\caption{} \vspace{-0.3cm}
			\includegraphics[width=\figwidthfull]{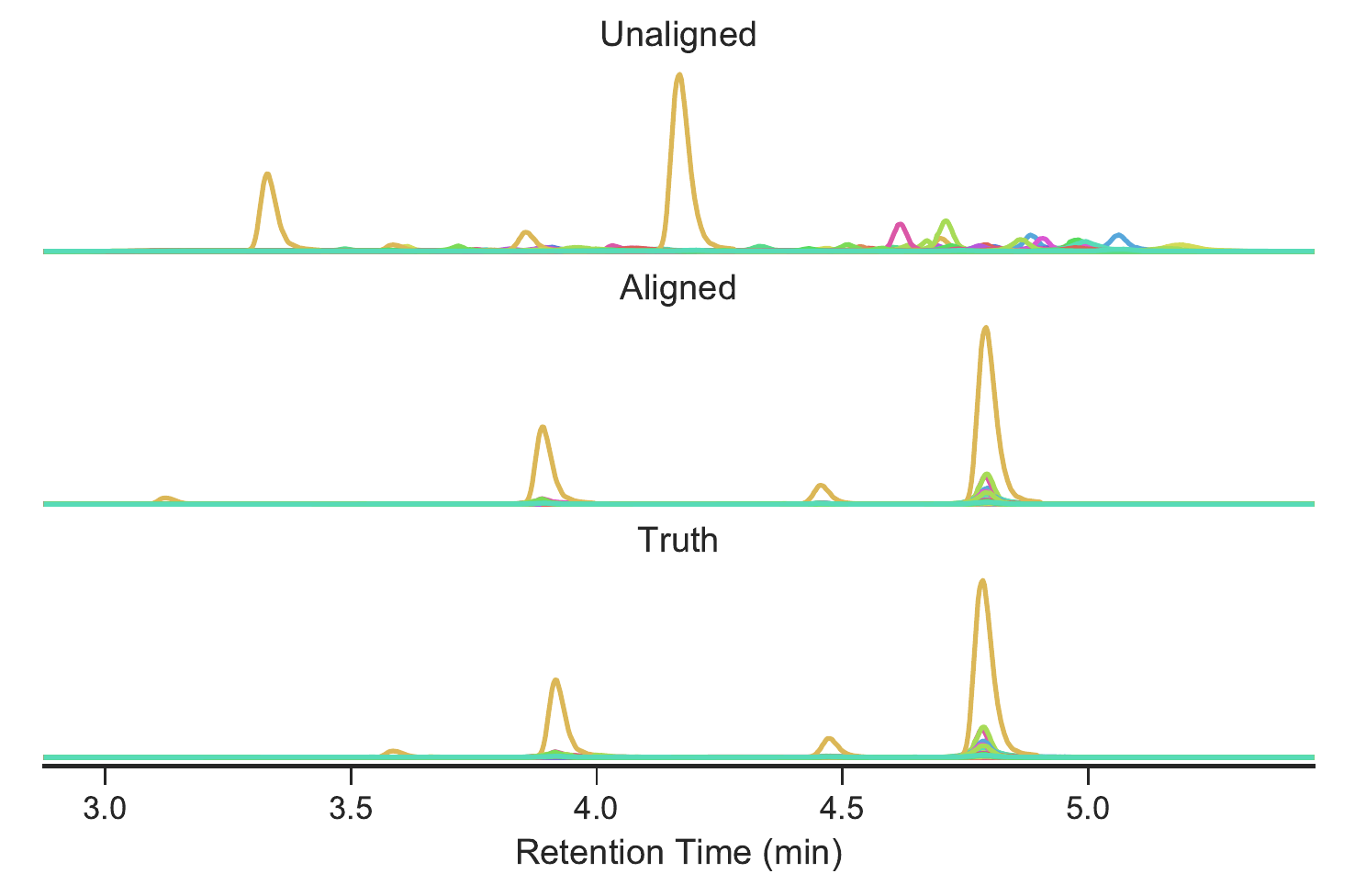}
		\end{subfigure}
	\end{center}
	\caption{Alignment outcome for the data set Breath88 using the best performing model F. For this network, the true positive rate was 0.985 and the false positive rate was 0.231. (a)~Chormatographic image. (b)~Chormatographic plot. Note, only the three thioethers peaks were expertly identified and aligned thus aligned in the ground truth image and plot.  
	} 
	\label{fig:AlignBreath88_F01}
\end{figure}

%

\clearpage
\section{Supplementary Tables}

\input{table_nomenclature}

\input{table_resultAllModels}

\input{table_resultTest}

\end{document}

%% file: table_resultsDVariants_part1.tex
\begin{table}
	\centering
	\caption{Variations made to model D. Model D is included here for reference as model D-01, and was re-run as a check of performance against those in Table~\ref{table:TrueFalsePositives}. Key: PE = peak encoder; ME = mass spectrum encoder; CE = chromatograph segment encoder; CL = convolutional layers in the chromatograph encoder. CL (L) and CL (R) refers to only the left or right stack of the convolutional layers. In the change column, 5, 20 and 30 indicates a change to the number of encoding (final layer) neurons for the modified component, while `-' indicates that no change is made. 
		`Increase' and `Decrease' refers to a change in the number of layers by one in the corresponding chromatograph encoder component. For model D-03, the simplification of the peak encoder indicates the removal of bi-directionality and changing the recurrent unit from LSTM to a GRU structure~\cite{ChungNIPS2014}.
		If dropout is unspecified it kept at its original value (0.2 for the mass and peak encoders and after the convolutions of the chromatogram encoder, 0 between the convolutions of the chromatogram encoder). The values report the average across the 10 repetitions.}
	\label{table:VariationModels}
	\resizebox{\textwidth}{!}{%
    \begin{tabular}{@{}cccccccccccccccc@{}}
    \toprule
    \multicolumn{1}{l}{} &  &  &  &  &  \multicolumn{3}{c}{Breath 103} &       & \multicolumn{3}{c}{Breath 73} &       & \multicolumn{3}{c}{Breath 88} \\
    \cmidrule(lr){6-8} \cmidrule(lr){10-12} \cmidrule(l){14-16} 
    \textbf{Models} & \begin{tabular}[c]{@{}c@{}}Modified\\ Component\end{tabular} & Dropout & Change &   & TP & FP & AUC &       & TP & FP & AUC &       & TP & FP & AUC  \\ 
    \midrule
    D-01  & -     & -   & -     &       & 0.997 & 0.004 & 0.9996 &       & 0.943 & 0.294 & 0.9220 &       & 0.945 & 0.327 & 0.9072 \\
    D-02  & PE    & -   & Remove &       & 0.999 & 0.005 & 0.9996 &       & 0.940 & 0.313 & 0.9198 &       & 0.942 & 0.324 & 0.9124 \\
    D-03  & PE    & -   & Simplify &       & 0.998 & 0.004 & 0.9996 &       & 0.944 & 0.312 & 0.9197 &       & 0.946 & 0.325 & 0.9112 \\
    \midrule
    D-04  & PE    & 0.5 & -     &       & 0.999 & 0.004 & 0.9997 &       & 0.947 & 0.302 & 0.9258 &       & 0.947 & 0.324 & 0.9154 \\
    D-05  & PE    & -   & 5  &       & 0.998 & 0.004 & 0.9995 &       & 0.922 & 0.260 & 0.9258 &       & 0.928 & 0.280 & 0.9177 \\
    D-06  & PE    & 0.5 & 20 &       & 0.997 & 0.003 & 0.9998 &       & 0.936 & 0.269 & 0.9263 &       & 0.942 & 0.290 & 0.9172 \\
    D-07  & PE    & 0.5 & 30 &       & 0.998 & 0.004 & 0.9998 &       & 0.921 & 0.237 & 0.9298 &       & 0.925 & 0.254 & 0.9214 \\
    D-08  & ME    & -   & 5  &       & 0.999 & 0.005 & 0.9996 &       & 0.938 & 0.275 & 0.9288 &       & 0.936 & 0.278 & 0.9226 \\
    D-09  & ME    & 0.5 & 20 &       & 0.997 & 0.003 & 0.9998 &       & 0.934 & 0.283 & 0.9248 &       & 0.934 & 0.297 & 0.9158 \\
    D-10  & ME    & 0.5 & 30 &       & 0.998 & 0.003 & 0.9996 &       & 0.921 & 0.266 & 0.9171 &       & 0.922 & 0.289 & 0.9076 \\
    D-11  & CE    & -   & 5  &       & 0.999 & 0.010 & 0.9993 &       & 0.924 & 0.288 & 0.9116 &       & 0.921 & 0.297 & 0.9024 \\
    D-12  & CE    & 0.5 & 20 &       & 0.996 & 0.005 & 0.9995 &       & 0.902 & 0.249 & 0.9178 &       & 0.903 & 0.278 & 0.9021 \\
    D-13  & CE    & 0.5 & 30 &       & 0.998 & 0.005 & 0.9997 &       & 0.934 & 0.307 & 0.9228 &       & 0.939 & 0.341 & 0.9113 \\
    \midrule
    D-14  & CL    & 0.2 & -  &       & 0.997 & 0.006 & 0.9995 &       & 0.904 & 0.279 & 0.9113 &       & 0.911 & 0.318 & 0.8938 \\
    D-15  & CL    & 0.5 & -  &       & 0.996 & 0.052 & 0.9983 &       & 0.886 & 0.261 & 0.8901 &       & 0.884 & 0.302 & 0.8664 \\
    D-16  & CL (R)  & -     & Increase &       & 0.999 & 0.004 & 0.9996 &       & 0.938 & 0.285 & 0.9262 &       & 0.938 & 0.300 & 0.9159 \\
    D-17  & CL (R)  & -     & Decrease &       & 0.997 & 0.006 & 0.9996 &       & 0.910 & 0.243 & 0.9171 &       & 0.912 & 0.255 & 0.9098 \\
    D-18  & CL (L)  & -     & Increase &       & 0.998 & 0.005 & 0.9995 &       & 0.920 & 0.288 & 0.9178 &       & 0.921 & 0.298 & 0.9110 \\
    D-19  & CL (L)  & -     & Decrease &       & 0.998 & 0.004 & 0.9996 &       & 0.927 & 0.268 & 0.9209 &       & 0.927 & 0.275 & 0.9134 \\
    \bottomrule
    \end{tabular}%
	}
\end{table}

%% file: table_resultsDVariants_part2.tex
\begin{table}
	\centering
	\caption{Further variations made to model D. Each variant here is a combination of variant 2 or 3 and six other variants. Variant 2 has no peak encoder, resulting in reducing the training and prediction time by at least an order of magnitude. Variant 3 has a simplified peak encoder, resulting in halving the training and prediction time. The six other variants have variations to mass or chromatogram encoders, and have slightly better performance than the original model. Model D is included here for reference as model D-01.  The values report the average across 10 repetitions. Note we did not use variant 7 here even though it had low FP rate because variant 7 uses a modification on the peak encoder, which is omitted in variant 2. }
	\label{table:VariationModels2}
	\resizebox{\textwidth}{!}{%
    \begin{tabular}{@{}ccccccccccccc@{}}
	\toprule
	&       & \multicolumn{3}{c}{Breath 103} &       & \multicolumn{3}{c}{Breath 73} &      & \multicolumn{3}{c}{Breath 88} \\
	\cmidrule(lr){3-5} \cmidrule(lr){7-9} \cmidrule(l){11-13} 
	Models & combination & TP    & FP    & AUC &       & TP    & FP    & AUC &       & TP    & FP    & AUC  \\
	D-01  &       & 0.997 & 0.004 & 0.9996 &       & 0.943 & 0.294 & 0.9220 &       & 0.945 & 0.327 & 0.9072 \\
	\midrule
	D-20  & 2, 8  & 0.999 & 0.004 & 0.9996 &       & 0.932 & 0.290 & 0.9223 &       & 0.935 & 0.299 & 0.9163 \\
	D-21  & 2, 9  & 0.999 & 0.005 & 0.9996 &       & 0.957 & 0.343 & 0.9237 &       & 0.960 & 0.375 & 0.9109 \\
	D-22  & 2, 11 & 0.998 & 0.004 & 0.9996 &       & 0.920 & 0.281 & 0.9141 &       & 0.920 & 0.297 & 0.9053 \\
	D-23  & 2, 12 & 0.998 & 0.004 & 0.9997 &       & 0.934 & 0.272 & 0.9262 &       & 0.937 & 0.307 & 0.9106 \\
	D-24  & 2, 17 & 0.998 & 0.005 & 0.9997 &       & 0.914 & 0.254 & 0.9197 &       & 0.913 & 0.277 & 0.9069 \\
	D-25  & 2, 19 & 0.997 & 0.004 & 0.9995 &       & 0.925 & 0.291 & 0.9107 &       & 0.925 & 0.308 & 0.8992 \\
	\midrule
	D-26  & 3, 8  & 0.998 & 0.005 & 0.9997 &       & 0.925 & 0.301 & 0.9144 &       & 0.926 & 0.318 & 0.9077 \\
	D-27  & 3, 9  & 0.999 & 0.005 & 0.9996 &       & 0.950 & 0.319 & 0.9190 &       & 0.946 & 0.341 & 0.9066 \\
	D-28  & 3, 11 & 0.997 & 0.005 & 0.9993 &       & 0.893 & 0.254 & 0.9056 &       & 0.898 & 0.282 & 0.8936 \\
	D-29  & 3, 12 & 0.998 & 0.004 & 0.9995 &       & 0.928 & 0.276 & 0.9163 &       & 0.930 & 0.316 & 0.9010 \\
	D-30  & 3, 17 & 0.999 & 0.005 & 0.9996 &       & 0.915 & 0.258 & 0.9171 &       & 0.921 & 0.286 & 0.9048 \\
	D-31  & 3, 19 & 0.998 & 0.004 & 0.9997 &       & 0.904 & 0.257 & 0.9180 &       & 0.904 & 0.268 & 0.9101 \\
	\bottomrule
	\end{tabular}%
	}
\end{table}

%% file: table_nomenclature.tex
\begin{longtable}{>{\raggedright}p{0.20\linewidth}>{\raggedright\arraybackslash}p{0.80\linewidth}}
	\caption{Nomenclature used in this paper.}
	\label{tab:nomenclature}\\
	\hline
	\textbf{Terms} & \textbf{Definition} \\
	\hline
	\endfirsthead
	\multicolumn{2}{c}%
	{\tablename\ \thetable\ -- \textit{Continued from previous page}} \\
	\hline
	\textbf{Terms} & \textbf{Definition} \\
	\hline
	\endhead
	\hline \multicolumn{2}{r}{\textit{Continued on next page}} \\
	\endfoot
	\endlastfoot
		accuracy & 
			the proportion of the total data that a classification system got right \\
		dropout & 
			a regularisation technique that randomly removes a number of neurons in the network during each training iteration \\
		encoder & 
			a network that maps raw inputs into feature representations\\
		false positive (FP) & 
			the proportion of negative cases that are incorrectly predicted as positive \\
		feature & 
			abstract representation extracted from the data, features can have physical meanings such as edges in an image \\
		group & 
			a set of peaks from multiple samples that are from the same compound \\
		loss & 
			a function of how much the predictions differs from the actual labels \\
		neurons & 
			a module in neural network that transforms multiple input data into one output value \\
		overfitting & 
			when a model fits training data so perfectly that it fails when predicting new data \\
		peaks & 
			local maxima on a chromatogram \\
		peaks detected & 
			peaks found by automatic peak detection algorithm \\
		peaks identified & 
			peaks that have been manually verified by domain experts such that identity of the peaks are known \\
		prediction & 
			model's output when provided with an input\\
		sample & 
			a breath or air specimen collected from an individual or a location at a specific time \\
		test data & 
			a set of data used to test the performance of the trained models\\
		training & 
			the process used to determine the best parameters for the models\\
		training data & 
			data used to train the models \\
		true positive (TP) & 
			the proportion of positive cases that are correctly determined as such \\
		validation data & 
			a subset of data set (dis-joint from training data) that is used to verify the increase in accuracy over training iterations \\
	\hline
\end{longtable}

%% file: table_resultAllModels.tex
\begin{table}[!h]
	\centering
	\caption{True positives and false positive of predictions for each model against each data set. The values report the average over ten repetitions. Cells where the values are italic and in bracket indicate models that were trained with the data set.}
	\label{table:TrueFalsePositives}
	\resizebox{\textwidth}{!}{%
	\begin{tabular}{@{}lccccccclccccccc@{}}
		\toprule
		& \multicolumn{7}{c}{\textbf{True Positives}}   &  & \multicolumn{7}{c}{\textbf{False Positive}}   \\ \cmidrule(lr){2-8} \cmidrule(l){10-16} 
		& \multicolumn{7}{c}{Models}                   &  & \multicolumn{7}{c}{Models}                   \\ \cmidrule(lr){2-8} \cmidrule(l){10-16} 
Data Set   & A     & B     & C     & D     & E     & F     & G     &  & A     & B     & C     & D     & E     & F     & G     \\ \midrule
Air103    & (\textit{1.000}) & (\textit{1.000}) & (\textit{1.000}) & (\textit{1.000}) & (\textit{1.000}) & (\textit{1.000}) & (\textit{1.000}) &  & (\textit{0.000}) & (\textit{0.000}) & (\textit{0.000}) & (\textit{0.000}) & (\textit{0.000})& (\textit{0.000}) & (\textit{0.001}) \\
Air115    & (\textit{1.000}) & (\textit{1.000}) & (\textit{1.000}) & (\textit{1.000}) & (\textit{1.000}) & (\textit{1.000}) & (\textit{1.000}) &  & (\textit{0.000}) & (\textit{0.000}) & (\textit{0.000}) & (\textit{0.000}) & (\textit{0.000}) & (\textit{0.001}) & (\textit{0.001}) \\
Air143    & 1.000          & (\textit{1.000}) & 1.000          & 1.000          & 1.000          & 0.999          & 0.997          &  & 0.009          & (\textit{0.000}) & 0.013          & 0.017          & 0.014          & 0.045          & 0.028          \\
Breath103 & 0.982          & 0.964          & (\textit{1.000}) & 0.998          & (\textit{1.000}) & (\textit{1.000}) & (\textit{1.000})         &  & 0.012          & 0.007          & (\textit{0.000}) & 0.005          & (\textit{0.000}) & (\textit{0.001}) & (\textit{0.000})          \\
Breath115 & 0.965          & 0.956          & 0.986          & (\textit{1.000}) & (\textit{1.000}) & 0.986          & 0.983 &  & 0.018          & 0.017          & 0.004          & (\textit{0.000}) & (\textit{0.000}) & 0.023          & 0.017 \\
Breath73  & 0.888          & 0.835          & 0.938          & 0.934          & 0.916          & (\textit{1.000}) & 0.957          &  & 0.238 & 0.187 & 0.264 & 0.294 & 0.298 & (\textit{0.250}) & 0.277          \\
Breath88  & 0.885          & 0.826          & 0.938          & 0.934          & 0.922          & 0.982          & (\textit{1.000}) &  & 0.254 & 0.216 & 0.290 & 0.320 & 0.294 & 0.192 & (\textit{0.159}) \\
\bottomrule
	\end{tabular}
	}
\end{table}

%% file: table_resultTest.tex
\begin{table}[!h]
  \centering
  \caption{True positives and false positive of predictions for six variant of model H (trained using all training data sets) against each test data set. The values report the average over ten repetitions. The values in bold are those with the best performance for the data set.}
  	\label{table:resultsTest}%
  	\resizebox{\textwidth}{!}{%
    \begin{tabular}{@{}lcccccclcccccc@{}}
		\toprule
		& \multicolumn{6}{c}{\textbf{True Positives}}   &  & \multicolumn{6}{c}{\textbf{False Positive}}   \\ \cmidrule(lr){2-7} \cmidrule(l){9-14} 
		& \multicolumn{6}{c}{Models}                   &  & \multicolumn{6}{c}{Models}                   \\ \cmidrule(lr){2-7} \cmidrule(l){9-14} 
		Data Set& H-01 & H-02 & H-20 & H-21 & H-23 & H-24 &  & H-01 & H-02 & H-20 & H-21 & H-23 & H-24 \\
		\midrule
    Air92 & 0.983 & 0.981 & \textbf{0.987} & \textbf{0.987} & 0.977 & 0.983 &  & 0.342 & 0.321 & 0.332 & 0.304 & \textbf{0.243} & 0.278 \\
    Air134 & 0.986 & 0.983 & 0.987 & 0.975 & \textbf{0.990} & 0.989 &  & 0.083 & 0.103 & 0.088 & \textbf{0.072} & 0.083 & 0.074 \\
    Field73 & 0.867 & 0.837 & 0.842 & 0.874 & 0.841 & \textbf{0.889} &  & 0.336 & 0.309 & 0.317 & 0.407 & \textbf{0.292} & 0.355 \\
    Field88 & 0.889 & 0.876 & 0.881 & \textbf{0.923} & 0.873 & 0.903 &  & 0.087 & \textbf{0.065} & 0.082 & 0.113 & 0.073 & 0.088 \\
    Field134 & 0.952 & 0.949 & 0.950 & 0.945 & 0.947 & \textbf{0.960} &  & 0.050 & 0.060 & 0.058 & \textbf{0.042} & 0.065 & 0.058 \\
    \midrule
    overall & 0.935 & 0.925 & 0.929 & 0.941 & 0.926 & \textbf{0.945} &       & 0.180 & 0.172 & 0.176 & 0.188 & \textbf{0.151} & 0.171 \\    
	\bottomrule
    \end{tabular}%
	}
\end{table}

%% file: manuscript.bbl
\begin{thebibliography}{10}
\expandafter\ifx\csname url\endcsname\relax
  \def\url#1{\texttt{#1}}\fi
\expandafter\ifx\csname urlprefix\endcsname\relax\def\urlprefix{URL }\fi
\expandafter\ifx\csname href\endcsname\relax
  \def\href#1#2{#2} \def\path#1{#1}\fi

\bibitem{LNS_ER2013}
B.~P. Lankadurai, E.~G. Nagato, M.~J. Simpson,
  \href{http://dx.doi.org/10.1139/er-2013-0011}{Environmental metabolomics: an
  emerging approach to study organism responses to environmental stressors},
  Environmental Reviews 21~(3) (2013) 180--205.
\newblock \href {http://dx.doi.org/10.1139/er-2013-0011}
  {\path{doi:10.1139/er-2013-0011}}.
\newline\urlprefix\url{http://dx.doi.org/10.1139/er-2013-0011}

\bibitem{Ellis_BM2012}
J.~Ellis, T.~Athersuch, L.~Thomas, F.~Teichert, M.~Perez-Trujillo, C.~Svendsen,
  D.~Spurgeon, R.~Singh, L.~Jarup, J.~Bundy, H.~Keun,
  \href{http://www.biomedcentral.com/1741-7015/10/61}{Metabolic profiling
  detects early effects of environmental and lifestyle exposure to cadmium in a
  human population}, BMC Medicine 10~(1) (2012) 61.
\newblock \href {http://dx.doi.org/10.1186/1741-7015-10-61}
  {\path{doi:10.1186/1741-7015-10-61}}.
\newline\urlprefix\url{http://www.biomedcentral.com/1741-7015/10/61}

\bibitem{BernaJID2015}
A.~Z. Berna, J.~S. McCarthy, X.~R. Wang, K.~J. Saliba, F.~G. Bravo,
  J.~Cassells, B.~Padovan, S.~C. Trowell,
  \href{http://jid.oxfordjournals.org/content/212/7/1120.abstract}{Analysis of
  breath specimens for biomarkers of plasmodium falciparum infection}, Journal
  of Infectious Diseases 212~(7) (2015) 1120--1128.
\newblock \href {http://dx.doi.org/10.1093/infdis/jiv176}
  {\path{doi:10.1093/infdis/jiv176}}.
\newline\urlprefix\url{http://jid.oxfordjournals.org/content/212/7/1120.abstract}

\bibitem{DragonieriJACI2007}
S.~Dragonieri, R.~Schot, B.~J. Mertens, S.~Le~Cessie, S.~A. Gauw,
  A.~Spanevello, O.~Resta, N.~P. Willard, T.~J. Vink, K.~F. Rabe, et~al., An
  electronic nose in the discrimination of patients with asthma and controls,
  Journal of Allergy and Clinical Immunology 120~(4) (2007) 856--862.

\bibitem{DragonieriLung2017}
S.~Dragonieri, G.~Pennazza, P.~Carratu, O.~Resta, Electronic nose technology in
  respiratory diseases, Lung 195~(2) (2017) 157--165.

\bibitem{BernaJBR2018}
A.~Berna, J.~McCarthy, X.~R. Wang, M.~Michie, F.~G. Bravo, J.~Cassells,
  S.~Trowell, \href{http://stacks.iop.org/1752-7163/12/i=4/a=046014}{Diurnal
  variation in expired breath volatiles in malaria-infected and healthy
  volunteers}, Journal of Breath Research 12~(4) (2018) 046014.
\newline\urlprefix\url{http://stacks.iop.org/1752-7163/12/i=4/a=046014}

\bibitem{GrissaFMB2016}
D.~Grissa, M.~P{\'e}t{\'e}ra, M.~Brandolini, A.~Napoli, B.~Comte,
  E.~Pujos-Guillot,
  \href{https://www.frontiersin.org/article/10.3389/fmolb.2016.00030}{Feature
  selection methods for early predictive biomarker discovery using untargeted
  metabolomic data}, Frontiers in Molecular Biosciences 3 (2016) 30.
\newblock \href {http://dx.doi.org/10.3389/fmolb.2016.00030}
  {\path{doi:10.3389/fmolb.2016.00030}}.
\newline\urlprefix\url{https://www.frontiersin.org/article/10.3389/fmolb.2016.00030}

\bibitem{VuMetabolites2013}
T.~N. Vu, K.~Laukens, \href{http://www.mdpi.com/2218-1989/3/2/259}{Getting your
  peaks in line: A review of alignment methods for nmr spectral data},
  Metabolites 3~(2) (2013) 259--276.
\newblock \href {http://dx.doi.org/10.3390/metabo3020259}
  {\path{doi:10.3390/metabo3020259}}.
\newline\urlprefix\url{http://www.mdpi.com/2218-1989/3/2/259}

\bibitem{KohJCA2010}
Y.~Koh, K.~K. Pasikanti, C.~W. Yap, E.~C.~Y. Chan,
  \href{http://www.sciencedirect.com/science/article/pii/S0021967310014962}{Comparative
  evaluation of software for retention time alignment of gas
  chromatography/time-of-flight mass spectrometry-based metabonomic data},
  Journal of Chromatography A 1217~(52) (2010) 8308 -- 8316.
\newblock \href {http://dx.doi.org/10.1016/j.chroma.2010.10.101}
  {\path{doi:10.1016/j.chroma.2010.10.101}}.
\newline\urlprefix\url{http://www.sciencedirect.com/science/article/pii/S0021967310014962}

\bibitem{ZhengJCA2013}
Y.-B. Zheng, Z.-M. Zhang, Y.-Z. Liang, D.-J. Zhan, J.-H. Huang, Y.-H. Yun,
  H.-L. Xie,
  \href{http://www.sciencedirect.com/science/article/pii/S0021967313003701}{Application
  of fast fourier transform cross-correlation and mass spectrometry data for
  accurate alignment of chromatograms}, Journal of Chromatography A 1286 (2013)
  175 -- 182.
\newblock \href
  {http://dx.doi.org/https://doi.org/10.1016/j.chroma.2013.02.063}
  {\path{doi:https://doi.org/10.1016/j.chroma.2013.02.063}}.
\newline\urlprefix\url{http://www.sciencedirect.com/science/article/pii/S0021967313003701}

\bibitem{DomingoAC2016}
X.~Domingo-Almenara, J.~Brezmes, M.~Vinaixa, S.~Samino, N.~Ramirez,
  M.~Ramon-Krauel, C.~Lerin, M.~D{\'\i}az, L.~Ib{\'a}{\~n}ez, X.~Correig,
  A.~Perera-Lluna, O.~Yanes,
  \href{https://doi.org/10.1021/acs.analchem.6b02927}{erah: A computational
  tool integrating spectral deconvolution and alignment with quantification and
  identification of metabolites in gc/ms-based metabolomics}, Analytical
  Chemistry 88~(19) (2016) 9821--9829, pMID: 27584001.
\newblock \href {http://dx.doi.org/10.1021/acs.analchem.6b02927}
  {\path{doi:10.1021/acs.analchem.6b02927}}.
\newline\urlprefix\url{https://doi.org/10.1021/acs.analchem.6b02927}

\bibitem{FuJCA2017}
H.-Y. Fu, O.~Hu, Y.-M. Zhang, L.~Zhang, J.-J. Song, P.~Lu, Q.-X. Zheng, P.-P.
  Liu, Q.-S. Chen, B.~Wang, X.-Y. Wang, L.~Han, Y.-J. Yu,
  \href{http://www.sciencedirect.com/science/article/pii/S0021967317310452}{Mass-spectra-based
  peak alignment for automatic nontargeted metabolic profiling analysis for
  biomarker screening in plant samples}, Journal of Chromatography A 1513
  (2017) 201 -- 209.
\newblock \href
  {http://dx.doi.org/https://doi.org/10.1016/j.chroma.2017.07.044}
  {\path{doi:https://doi.org/10.1016/j.chroma.2017.07.044}}.
\newline\urlprefix\url{http://www.sciencedirect.com/science/article/pii/S0021967317310452}

\bibitem{CouprieJCA2017}
C.~Couprie, L.~Duval, M.~Moreaud, S.~H{\'e}non, M.~Tebib, V.~Souchon,
  \href{http://www.sciencedirect.com/science/article/pii/S0021967317300171}{Barchan:
  Blob alignment for robust chromatographic analysis}, Journal of
  Chromatography A 1484 (2017) 65 -- 72.
\newblock \href
  {http://dx.doi.org/https://doi.org/10.1016/j.chroma.2017.01.003}
  {\path{doi:https://doi.org/10.1016/j.chroma.2017.01.003}}.
\newline\urlprefix\url{http://www.sciencedirect.com/science/article/pii/S0021967317300171}

\bibitem{YangJCA2018}
T.-B. Yang, P.~Yan, M.~He, L.~Hong, R.~Pei, Z.-M. Zhang, L.-Z. Yi, X.-Y. Yuan,
  \href{http://www.sciencedirect.com/science/article/pii/S0021967318307209}{Application
  of subwindow factor analysis and mass spectral information for accurate
  alignment of non-targeted metabolic profiling}, Journal of Chromatography A
  1563 (2018) 162 -- 170.
\newblock \href
  {http://dx.doi.org/https://doi.org/10.1016/j.chroma.2018.05.071}
  {\path{doi:https://doi.org/10.1016/j.chroma.2018.05.071}}.
\newline\urlprefix\url{http://www.sciencedirect.com/science/article/pii/S0021967318307209}

\bibitem{Ottensmann2018}
M.~Ottensmann, M.~A. Stoffel, H.~J. Nichols, J.~I. Hoffman,
  \href{http://europepmc.org/articles/PMC5991698}{Gcalignr: An r package for
  aligning gas-chromatography data for ecological and evolutionary studies},
  PloS one 13~(6) (2018) e0198311.
\newblock \href {http://dx.doi.org/10.1371/journal.pone.0198311}
  {\path{doi:10.1371/journal.pone.0198311}}.
\newline\urlprefix\url{http://europepmc.org/articles/PMC5991698}

\bibitem{Chollet2017}
F.~Chollet, Deep Learning with Python, Manning Publications, 2017.

\bibitem{Goodfellow2016}
A.~C. Ian~Goodfellow, Yoshua~Bengio,
  \href{http://www.deeplearningbook.org}{Deep Learning}, MIT Press, 2016.
\newline\urlprefix\url{http://www.deeplearningbook.org}

\bibitem{fei2006}
L.~Fei-Fei, R.~Fergus, P.~Perona, One-shot learning of object categories, IEEE
  transactions on pattern analysis and machine intelligence 28~(4) (2006)
  594--611.

\bibitem{Bromley1994}
J.~Bromley, I.~Guyon, Y.~LeCun, E.~S{\"a}ckinger, R.~Shah, Signature
  verification using a" siamese" time delay neural network, in: Advances in
  neural information processing systems, 1994, pp. 737--744.

\bibitem{Chopra2005}
S.~Chopra, R.~Hadsell, Y.~LeCun, Learning a similarity metric discriminatively,
  with application to face verification, in: Computer Vision and Pattern
  Recognition, 2005. CVPR 2005. IEEE Computer Society Conference on, Vol.~1,
  IEEE, 2005, pp. 539--546.

\bibitem{Mueller2016}
J.~Mueller, A.~Thyagarajan, Siamese recurrent architectures for learning
  sentence similarity., in: AAAI, Vol.~16, 2016, pp. 2786--2792.

\bibitem{Szegedy2015}
C.~Szegedy, W.~Liu, Y.~Jia, P.~Sermanet, S.~Reed, D.~Anguelov, D.~Erhan,
  V.~Vanhoucke, A.~Rabinovich, Going deeper with convolutions, in: Proceedings
  of the IEEE conference on computer vision and pattern recognition, 2015, pp.
  1--9.

\bibitem{Simonyan2014}
K.~Simonyan, A.~Zisserman, {Very Deep Convolutional Networks for Large-Scale
  Image Recognition} (Sep. 2014).
\newblock \href {http://arxiv.org/abs/1409.1556} {\path{arXiv:1409.1556}}.

\bibitem{WangJCB2018}
X.~R. Wang, J.~Cassells, A.~Z. Berna,
  \href{http://www.sciencedirect.com/science/article/pii/S1570023218306470}{Stability
  control for breath analysis using gc-ms}, Journal of Chromatography B
  1097-1098 (2018) 27 -- 34.
\newblock \href {http://dx.doi.org/10.1016/j.jchromb.2018.08.024}
  {\path{doi:10.1016/j.jchromb.2018.08.024}}.
\newline\urlprefix\url{http://www.sciencedirect.com/science/article/pii/S1570023218306470}

\bibitem{Eilers2005}
P.~H. Eilers, H.~F. Boelens, Baseline correction with asymmetric least squares
  smoothing, Leiden University Medical Centre Report 1~(1) (2005) 5.

\bibitem{VivoTruyols2005133}
G.~Viv\'{o}-Truyols, J.~Torres-Lapasi\'{o}, A.~van Nederkassel, Y.~V. Heyden,
  D.~Massart,
  \href{http://www.sciencedirect.com/science/article/pii/S0021967305006424}{Automatic
  program for peak detection and deconvolution of multi-overlapped
  chromatographic signals: Part i: Peak detection}, Journal of Chromatography A
  1096~(1--2) (2005) 133 -- 145, chemical Separations and Chemometrics.
\newblock \href {http://dx.doi.org/10.1016/j.chroma.2005.03.092}
  {\path{doi:10.1016/j.chroma.2005.03.092}}.
\newline\urlprefix\url{http://www.sciencedirect.com/science/article/pii/S0021967305006424}

\bibitem{Nielsen1998}
N.-P.~V. Nielsen, J.~M. Carstensen, J.~Smedsgaard,
  \href{http://www.sciencedirect.com/science/article/pii/S0021967398000211}{Aligning
  of single and multiple wavelength chromatographic profiles for chemometric
  data analysis using correlation optimised warping}, Journal of Chromatography
  A 805~(1) (1998) 17 -- 35.
\newblock \href {http://dx.doi.org/10.1016/S0021-9673(98)00021-1}
  {\path{doi:10.1016/S0021-9673(98)00021-1}}.
\newline\urlprefix\url{http://www.sciencedirect.com/science/article/pii/S0021967398000211}

\bibitem{TomasiJChemom2004}
G.~Tomasi, F.~van~den Berg, C.~Andersson,
  \href{https://onlinelibrary.wiley.com/doi/abs/10.1002/cem.859}{Correlation
  optimized warping and dynamic time warping as preprocessing methods for
  chromatographic data}, Journal of Chemometrics 18~(5) (2004) 231--241.
\newblock \href {http://dx.doi.org/10.1002/cem.859}
  {\path{doi:10.1002/cem.859}}.
\newline\urlprefix\url{https://onlinelibrary.wiley.com/doi/abs/10.1002/cem.859}

\bibitem{SkovJChemom2007}
T.~Skov, F.~van~den Berg, G.~Tomasi, R.~Bro,
  \href{https://onlinelibrary.wiley.com/doi/abs/10.1002/cem.1031}{Automated
  alignment of chromatographic data}, Journal of Chemometrics 20~(11‐12)
  (2007) 484--497.
\newblock \href {http://dx.doi.org/10.1002/cem.1031}
  {\path{doi:10.1002/cem.1031}}.
\newline\urlprefix\url{https://onlinelibrary.wiley.com/doi/abs/10.1002/cem.1031}

\bibitem{SmithAC2006}
C.~A. Smith, E.~J. Want, G.~O'Maille, R.~Abagyan, G.~Siuzdak,
  \href{https://doi.org/10.1021/ac051437y}{Xcms: Processing mass spectrometry
  data for metabolite profiling using nonlinear peak alignment, matching, and
  identification}, Analytical Chemistry 78~(3) (2006) 779--787.
\newblock \href {http://dx.doi.org/10.1021/ac051437y}
  {\path{doi:10.1021/ac051437y}}.
\newline\urlprefix\url{https://doi.org/10.1021/ac051437y}

\bibitem{DaszykowskiJCA2007}
M.~Daszykowski, B.~Walczak,
  \href{http://www.sciencedirect.com/science/article/pii/S0021967307019036}{Target
  selection for alignment of chromatographic signals obtained using monochannel
  detectors}, Journal of Chromatography A 1176~(1--2) (2007) 1 -- 11.
\newblock \href {http://dx.doi.org/10.1016/j.chroma.2007.10.099}
  {\path{doi:10.1016/j.chroma.2007.10.099}}.
\newline\urlprefix\url{http://www.sciencedirect.com/science/article/pii/S0021967307019036}

\bibitem{Wang2019}
X.~R. Wang, M.~Li,
  \href{https://data.csiro.au/dap/landingpage?pid=csiro:38958}{{GCMS} peak
  alignment: Data and results}.
\newblock \href {http://dx.doi.org/10.25919/5ca16f2db73a9}
  {\path{doi:10.25919/5ca16f2db73a9}}.
\newline\urlprefix\url{https://data.csiro.au/dap/landingpage?pid=csiro:38958}

\bibitem{Lecun2015}
Y.~LeCun, Y.~Bengio, G.~Hinton, Deep learning, Nature 521~(7553) (2015) 436.

\bibitem{Lecun2010}
Y.~LeCun, K.~Kavukcuoglu, C.~Farabet, Convolutional networks and applications
  in vision, in: Proceedings of 2010 IEEE International Symposium on Circuits
  and Systems, IEEE, 2010, pp. 253--256.

\bibitem{Osadchy2007}
M.~Osadchy, Y.~L. Cun, M.~L. Miller, Synergistic face detection and pose
  estimation with energy-based models, Journal of Machine Learning Research
  8~(May) (2007) 1197--1215.

\bibitem{Krizhevsky2012}
A.~Krizhevsky, I.~Sutskever, G.~E. Hinton, Imagenet classification with deep
  convolutional neural networks, in: Advances in neural information processing
  systems, 2012, pp. 1097--1105.

\bibitem{Raina2009}
R.~Raina, A.~Madhavan, A.~Y. Ng, Large-scale deep unsupervised learning using
  graphics processors, in: Proceedings of the 26th annual international
  conference on machine learning, ACM, 2009, pp. 873--880.

\bibitem{Hinton2012}
G.~Hinton, L.~Deng, D.~Yu, G.~E. Dahl, A.-r. Mohamed, N.~Jaitly, A.~Senior,
  V.~Vanhoucke, P.~Nguyen, T.~N. Sainath, et~al., Deep neural networks for
  acoustic modeling in speech recognition: The shared views of four research
  groups, IEEE Signal processing magazine 29~(6) (2012) 82--97.

\bibitem{Sutskever2014}
I.~Sutskever, O.~Vinyals, Q.~V. Le, Sequence to sequence learning with neural
  networks, in: Advances in neural information processing systems, 2014, pp.
  3104--3112.

\bibitem{Chellapilla2006a}
K.~Chellapilla, M.~Shilman, P.~Simard, Optimally combining a cascade of
  classifiers, in: Document Recognition and Retrieval XIII, Vol. 6067,
  International Society for Optics and Photonics, 2006, p. 60670Q.

\bibitem{Chellapilla2006b}
K.~Chellapilla, P.~Simard, A new radical based approach to offline handwritten
  east-asian character recognition, in: Tenth International Workshop on
  Frontiers in Handwriting Recognition, Suvisoft, 2006.

\bibitem{Hadsell2009}
R.~Hadsell, P.~Sermanet, J.~Ben, A.~Erkan, M.~Scoffier, K.~Kavukcuoglu,
  U.~Muller, Y.~LeCun, Learning long-range vision for autonomous off-road
  driving, Journal of Field Robotics 26~(2) (2009) 120--144.

\bibitem{Lecun1989}
Y.~LeCun, Generalization and network design strategies, Tech. Rep. CRG-TR-89-4,
  University of Toronto (Jun. 1989).

\bibitem{Zhou1988}
Y.-T. Zhou, R.~Chellappa, Computation of optical flow using a neural network,
  in: IEEE International Conference on Neural Networks, Vol. 1998, 1988, pp.
  71--78.

\bibitem{RumelhartNature1986}
D.~E. Rumelhart, G.~E. Hinton, R.~J. Williams, Learning representations by
  back-propagating errors, Nature 323 (1986) 533--536.
\newblock \href {http://dx.doi.org/10.1038/323533a0}
  {\path{doi:10.1038/323533a0}}.

\bibitem{Schmidhuber1997}
J.~Schmidhuber, S.~Hochreiter, Long short-term memory, Neural Comput 9~(8)
  (1997) 1735--1780.

\bibitem{Glorot2011}
X.~Glorot, A.~Bordes, Y.~Bengio, Deep sparse rectifier neural networks, in:
  Proceedings of the fourteenth international conference on artificial
  intelligence and statistics, 2011, pp. 315--323.

\bibitem{Srivastava2014}
N.~Srivastava, G.~Hinton, A.~Krizhevsky, I.~Sutskever, R.~Salakhutdinov,
  Dropout: a simple way to prevent neural networks from overfitting, The
  Journal of Machine Learning Research 15~(1) (2014) 1929--1958.

\bibitem{ChungNIPS2014}
J.~Chung, C.~Gulcehre, K.~Cho, Y.~Bengio, Empirical evaluation of gated
  recurrent neural networks on sequence modeling, in: NIPS 2014 Workshop on
  Deep Learning, December 2014, 2014.

\bibitem{Bai2018}
S.~Bai, J.~Z. Kolter, V.~Koltun, An empirical evaluation of generic
  convolutional and recurrent networks for sequence modeling, arXiv preprint
  arXiv:1803.01271.

\bibitem{Yin2017}
W.~Yin, K.~Kann, M.~Yu, H.~Schütze, Comparative study of cnn and rnn for
  natural language processing (2017).
\newblock \href {http://arxiv.org/abs/1702.01923} {\path{arXiv:1702.01923}}.

\bibitem{Gross2017}
W.~Gro{\ss}, S.~Lange, J.~B{\"o}decker, M.~Blum, Predicting time series with
  space-time convolutional and recurrent neural networks, in: Proceeding of
  European Symposium on Artificial Neural Networks, Computational Intelligence
  and Machine Learning, 2017, pp. 71--76.

\bibitem{Lin2017}
T.~Lin, T.~Guo, K.~Aberer, Hybrid neural networks for learning the trend in
  time series, Tech. rep. (2017).

\end{thebibliography}
